\documentclass{article}

\PassOptionsToPackage{numbers,compress}{natbib}
\usepackage[preprint]{neurips_2026}
\usepackage[utf8]{inputenc}
\usepackage[T1]{fontenc}
\usepackage[hidelinks]{hyperref}
\usepackage{url}
\usepackage{booktabs}
\usepackage{amsfonts}
\usepackage{amsmath}
\usepackage{amssymb}
\usepackage{microtype}
\usepackage{wrapfig}
\usepackage{xcolor}
\usepackage[most]{tcolorbox}
\usepackage{cleveref}
\usepackage{graphicx}
\usepackage{tabularx}
\usepackage{array}
\usepackage{capt-of}
\usepackage{enumitem}
\usepackage{authblk}

\title{VAMPS: Visual-Assisted Mathematical Problem Solving Benchmark}

\author[1]{Amirhossein Dabiriaghdam}
\author[2]{Shayan Vassef\textsuperscript{*}}
\author[3]{Mohammadreza Bakhtiari\textsuperscript{*}}
\author[1]{Yasamin Medghalchi\textsuperscript{*}}
\author[1]{Ilker Hacihaliloglu}
\author[2]{Mesrob Ohannessian}
\author[1]{Lele Wang}
\author[1]{Giuseppe Carenini}

\affil[ ]{
\textsuperscript{1}University of British Columbia \quad
\textsuperscript{2}University of Illinois Chicago \quad
\textsuperscript{3}Stony Brook University
}

\affil[*]{Equal Contribution}
\affil[ ]{
{GitHub (Code \& Data)}: \href{https://github.com/vampsbenchmark/VAMPS}{vampsbenchmark/VAMPS}
}

\newcounter{prompt}[section]
\renewcommand{\theprompt}{\thesection.\arabic{prompt}}
\newenvironment{promptbox}[2][]{%
    \refstepcounter{prompt}%
    \begin{tcolorbox}[
        colback=yellow!10!white,      
        colframe=yellow!60!orange,    
        fonttitle=\bfseries,       
        title={Prompt \theprompt: #2},
        enhanced,                  
        attach boxed title to top left={yshift=-2mm, xshift=5mm},
        boxed title style={colback=yellow!60!orange},
        sharp corners=south,       
        drop shadow,
        breakable,                 
        #1                         
    ]%
    \ttfamily\small
    \setlength{\baselineskip}{1.6\baselineskip}%
    \obeylines\obeyspaces\ignorespaces
}{%
    \end{tcolorbox}
}

\newcommand{\yesmark}{\textcolor{green!60!black}{\checkmark}}
\newcommand{\nomark}{\textcolor{red!70!black}{\(\times\)}}

\newcommand{\promptfont}{\ttfamily\fontsize{8}{9}\selectfont}

\begin{document}

\maketitle

\begin{abstract}
Multimodal large language models are increasingly capable of complex reasoning, yet their performance often degrades when they must externalize a problem through a tool and then reason over the tool’s output, specifically when they rely on visual aids. This gap is especially important because real engineering and scientific workflows often rely on visualization tools for analysis, validation, and decision-making. To study this discrepancy, we introduce \textbf{VAMPS} (\emph{Visual-Assisted Mathematical Problem Solving}), a benchmark for graph-assisted mathematics. VAMPS contains 1,168 multimodal, bilingual multiple-choice question-answer pairs drawn from Iranian University Entrance Exam algebra and calculus problems and expanded with human-reviewed LLM-generated synthetic variants, all selected so that plotting provides a natural solution strategy by revealing intersections, extrema, asymptotes, etc. Designed for both benchmarking and diagnosis, VAMPS goes beyond prior multimodal benchmarks that primarily evaluate reasoning over fixed visual inputs by testing whether a model can benefit from \emph{constructing} a useful graph and \emph{grounding} its answer in the resulting visualization.
Overall, we found that across a diverse set of models, direct analytical solving surprisingly outperforms tool-enabled visual solving, even on problems where plotting is a natural strategy.
\end{abstract}

\section{Introduction}
\label{sec:intro}


Recent large language models (LLMs) have demonstrated strong capabilities in solving mathematical problems directly from text, often by generating symbolic derivations, decomposing intermediate steps, or delegating exact computation to code-like representations~\citep{gao2022pal, chen2022pot, gou2023tora}. These advances extend to highly challenging, olympiad-level problems~\citep{luong2025towards}. However, real-world scientific and engineering workflows require more than text-based analysis alone. In practice, scientific workflows are inherently multimodal and iterative, involving the integration of computation, simulation, visualization, and interpretation. Experts rely on this interplay to validate hypotheses, refine models, and make informed decisions~\citep{callahan2006vistrails, keim2008visualanalytics, vanliere1997computational, barker2008workflow}.

Motivated by this gap, recent work has explored extending reasoning capabilities from LLMs to vision–language models (VLMs), and multimodal LLMs (MLLMs). In LLMs, significant gains in reasoning have been driven by reinforcement learning–based fine-tuning, which encourages structured, multi-step problem solving. However, attempts to replicate this success in the multimodal setting have largely fallen short~\citep{zhou2025r1, liu2025seg}. 
Many existing approaches remain fundamentally text-driven: images are processed only during an initial stage, while subsequent reasoning is carried out entirely in text, without intermediate visual reasoning steps.
To address this limitation, recent inference frameworks such as Visual Sketchpad~\citep{hu2024visual} and Refocus~\citep{fu2025refocus} introduce intermediate visual reasoning during inference and demonstrate improved performance on multimodal tasks. These approaches suggest that reasoning need not remain entirely textual; instead, intermediate visual representations can serve as evidence to guide and validate downstream decisions. This perspective is especially natural in mathematics, where coordinating algebraic and graphical representations is widely regarded as foundational for learning functions and equations~\citep{leinhardt1990functions, donnelly2020impact}.

In algebraic problem solving, this coordination is often supported by external visual tools. Graphing calculators allow a solver to transform symbolic expressions into plots, inspect intersections, monotonicity, extrema, asymptotes, inverse relationships, and relative ordering, and then use these visual cues to guide the final answer. Thus, graph-assisted mathematical reasoning requires more than solving equations symbolically or interpreting a given image. A model must decide what should be plotted, communicate that intent to a tool, inspect the resulting visualization, and integrate the visual evidence into its final decision.
Despite this, existing math benchmarks still largely evaluate models in text-only settings or with static visual inputs. As a result, they do not fully test the complete \emph{reasoning-to-perception handoff} required by graph-assisted problem solving: the model must transform the problem into an informative plot, and the generated plot must then be read visually to find the answer. This raises a central question: can modern models actually benefit from the equation-to-graph translation that makes plotting useful for human problem solvers?

Following this motivation, we introduce \textbf{VAMPS}, a benchmark built around the Iranian University Entrance Exam\footnote{\href{https://en.wikipedia.org/wiki/Iranian\_University\_Entrance\_Exam}{Konkour}} mathematics problems (algebra and calculus). VAMPS contains 1,168 multimodal multiple-choice question-answer (QA) pairs. The benchmark’s core consists of 218 real problems, each provided in Persian (Farsi) alongside a manually-checked English translation, yielding 436 original question instances in total. We further extend this core with synthetic multimodal variants generated from the real question seeds with LLM assistance, and then reviewed by humans. To our knowledge, VAMPS is the first bilingual Persian--English benchmark designed to evaluate visual reasoning with an external graphing tool. Crucially, the questions are curated such that plotting is often a natural---and in many cases preferable---solution strategy, making the benchmark well suited for evaluating whether models can construct and use visual evidence during mathematical reasoning.
In this study, we use Desmos \citep{desmos} as the graphing tool for generating visual representations. Desmos naturally supports function plotting and visual analysis, while also producing auditable artifacts---including expressions, plots, and screenshots---that can be inspected post hoc. This enables a fine-grained evaluation of both final-answer accuracy and intermediate tool-use behavior: Whether models request appropriate plots, whether they correctly interpret the generated graphs, and whether they use the visual evidence to select the correct answer.

This framing leads to a simple but important empirical question: Should access to a plotting tool reliably improve model performance? For human problem solvers, visual inspection often clarifies relationships that are difficult to track symbolically. For current models, however, the graphing interface may also expose new failure modes, including incorrect plot construction, incomplete tool use, misinterpretation of visual cues, or failure to integrate graphical evidence into the final answer. VAMPS is designed to test whether external visual tools help models in practice, or whether the reasoning-to-perception handoff remains a bottleneck for current systems.
In summary, our key \textbf{contributions} are as follows: \textbf{(1)} We introduce VAMPS, to our knowledge the first Persian--English mathematics benchmark for agentic model evaluation, with 1{,}168 multimodal QA pairs. \textbf{(2)} We formalize the \emph{reasoning-to-perception handoff} as a core bottleneck in tool-assisted algebraic reasoning, and provide an analysis framework for understanding why visual tool use can hurt performance even when visualization appears fair, meaningful, and helpful. \textbf{(3)} We report bilingual benchmark results across complementary solving regimes, directly comparing analytical text-only baselines, tool-enabled visual solving, and provided-visualization probes.

\section{Related Work}
\label{sec:related}

\textbf{Agentic and tool-augmented approaches to mathematical reasoning.}
Work on mathematical reasoning with LLMs has moved well beyond treating them as a standalone text generator. A line of research improves mathematical performance by delegating exact execution to external runtimes. PAL translates a word problem into executable code and leaves exact computation to the interpreter \citep{gao2022pal}. Program of Thoughts Prompting sharpens this decomposition by explicitly separating reasoning from calculation \citep{chen2022pot}. ToRA extends the idea to a more agentic format, where the model alternates between natural-language reasoning and tool use over multi-step trajectories \citep{gou2023tora}. In this line of work, tools are valuable because they help preserve rigor: the external interface returns symbolic, numeric, or executable structure rather than visual output that must be interpreted. A second line of work uses tightly structured neuro-symbolic systems to attack harder mathematical domains. AlphaGeometry and AlphaGeometry~2 combine learned proposal mechanisms with symbolic deduction to solve advanced geometry problems \citep{trinh2024alphageometry, deepmind2024alphageometry2}. Likewise, Inter-GPS, GeoQA, and UniGeo show that multimodal geometry can benefit from explicit symbolic programs and unified sequence-generation views of geometric calculation and proof \citep{lu2021intergps, chen2021geoqa, chen2022unigeo}. These systems are important contrast cases for our VAMPS. Their strongest gains come from maintaining or recovering formal structure. VAMPS instead studies a setting where the external tool does \emph{not} return a proof object, a program, or a symbolic state; it returns a plot that must be interpreted visually. This distinction matters for how we think about tool calling more broadly. Toolformer and ReAct established the now-standard view that tool use expands model capability by enabling external action, execution, and retrieval \citep{schick2023toolformer, yao2023react}. However, a substantial benchmark literature has since shown that tool calling also creates new failure surfaces: models must choose the right tool, form valid arguments, track state, recover from poor intermediate outputs, and remain stable over multi-turn interactions \citep{li2023apibank, patil2025bfcl, lu2024toolsandbox, yao2024taubench, liu2024agentbench, mialon2023gaia}. VAMPS inherits all of those difficulties and adds one more: the tool output is an image whose decisive content may be subtle.
\begin{figure}[t]
    \centering
    \includegraphics[width=\linewidth]{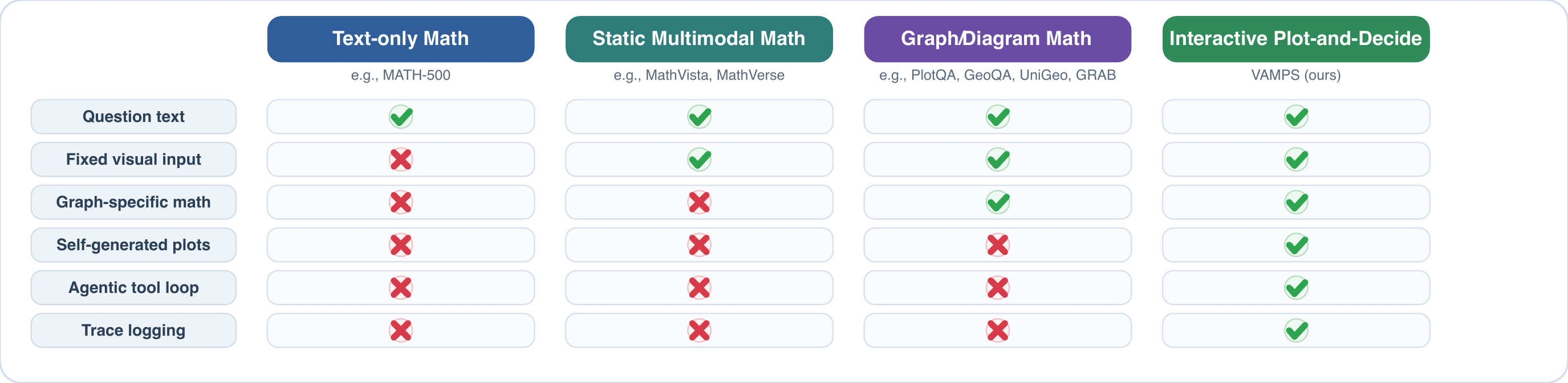}
    \caption{Comparison of existing mathematical benchmarks by whether they provide visual input, target graph-specific mathematics, require self-generated plots, have an agentic tool loop, etc.}
    \label{fig:benchmark_comparison}
\end{figure}
VAMPS' setting also connects to recent work on visual scratchpads and diagrammatic reasoning. Hsu et al. study whether LLMs benefit from generating and reading diagrammatic abstractions \citep{hsu2023scratchpad}. VAMPS enable a similar investigation, but more systematically and centered on plotting.

\textbf{Benchmark datasets for visual and multimodal mathematics.}
The benchmark landscape around visual mathematics is now broad, but highly heterogeneous. Some datasets focus on synthetic chart or plot reading, such as FigureQA, PlotQA, and ChartQA \citep{kahou2017figureqa, methani2020plotqa, masry2022chartqa}. These resources are valuable for probing perception over graphs and charts, but they usually evaluate models on fixed images rather than on self-generated visual evidence. A separate family of benchmarks focuses on geometry diagrams and multimodal mathematical programs, including Geometry3K, GeoQA, and UniGeo \citep{lu2021intergps, chen2021geoqa, chen2022unigeo}. These benchmarks emphasize geometry and formal reasoning pipelines more than visual-aided decision-making. Recent multimodal math benchmarks broadened the scope further. MathVista consolidates 28 prior multimodal datasets and three new datasets---IQTest, FunctionQA, and PaperQA---into a large benchmark for mathematical reasoning in visual contexts \citep{lu2024mathvista}. VCBench focuses on explicit visual dependency, with multi-image mathematics problems designed so that the decisive information is distributed across supplied visuals rather than recoverable from text alone \citep{wang2025vcbench}. MV-MATH pushes further toward interleaved multi-visual settings, where mathematical evidence is spread across several coordinated images \citep{wang2025mvmath}. MathVerse pushes harder on modality control by rewriting each problem into multiple versions that vary the balance of textual and visual information \citep{zhang2024mathverse}. GRAB targets graph analysis directly, with questions about chart properties, transforms, and realistic graph variants \citep{roberts2025grab}. Collectively, these datasets show that visual mathematical reasoning remains challenging even for strong multimodal LLMs. However, they still mostly treat the image as a \emph{given} input. VAMPS differs in both source material and evaluation protocol. It is anchored in real Konkour questions rather than only synthetic plots or heavily reauthored textbook-style collections, and it is released bilingually. Most importantly, it evaluates a multi-turn agentic regime where the model must produce its own plots, inspect them, and answer visually from that artifact. This makes VAMPS both a visual math reasoning benchmark and an agentic tool-use benchmark. Figure.~\ref{fig:benchmark_comparison} gives a high-level positioning of VAMPS relative to prior math benchmark families; Table~\ref{tab:benchmark-landscape-appendix} in the Appendix provides a more comprehensive comparison.

\section{The VAMPS Benchmark}
\label{sec:benchmark}
\subsection{Data Creation}
\label{sec:data-creation}

\begin{figure*}[h]
    \centering
    \includegraphics[width=\textwidth]{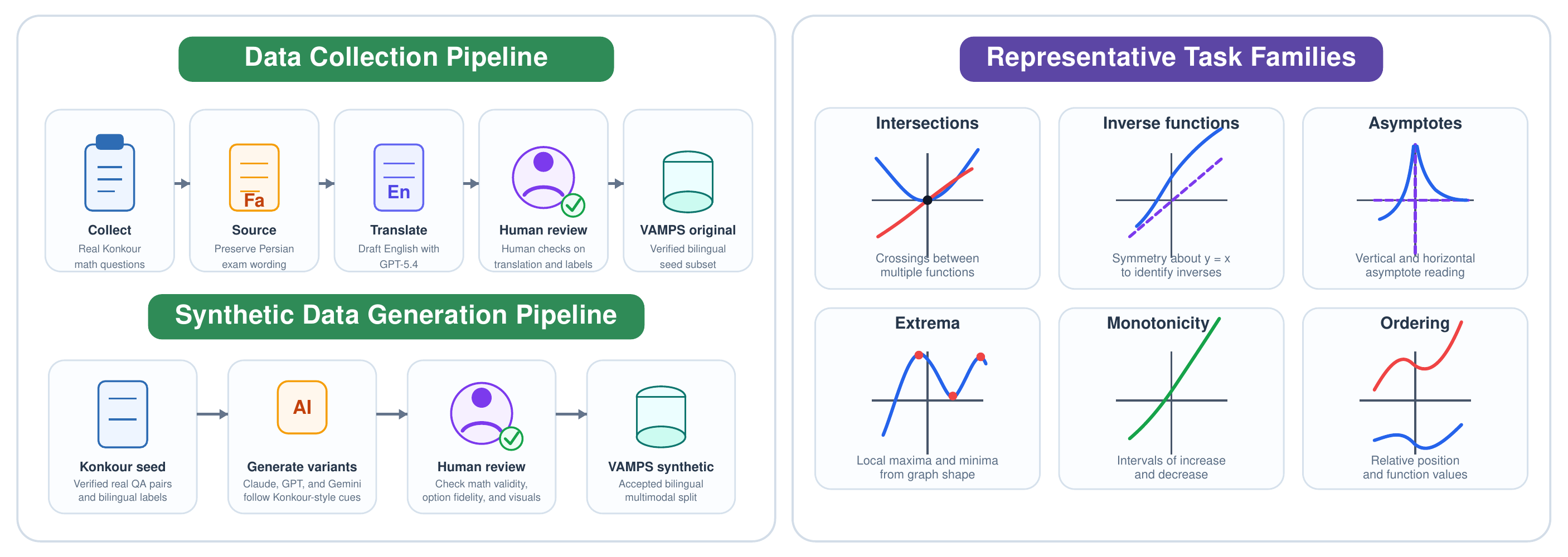}
    \caption{VAMPS dataset construction. \textit{Left:} The real seed is built by collecting Konkour math questions, preserving original Persian wording, translating to English with GPT-5.4, and manually verifying it. Synthetic variants are then generated from this seed using LLM assistants, accepted only after human review for validity and consistency. \textit{Right:} Representative graph-mediated task families in VAMPS questions: intersections, inverse functions, asymptotes, extrema, monotonicity, and ordering; illustrating the visual cues that motivate the benchmark design.}
    \vspace{-12pt}
    \label{fig:seed-overview}
\end{figure*}

VAMPS is anchored in 218 Konkour questions drawn from eight consecutive exam years (2016-2023). Each year of the Konkour mathematics section contains 140 questions, yielding an initial pool of 1{,}120 candidate questions. The questions pool spans multiple difficulty levels rather than only highly stylized or purely synthetic problems. We manually inspected every question in this pool and retained only those that fit the benchmark's scope, namely, questions like Figure \ref{fig:seed-overview} (right) for which graphing is a natural and informative solution strategy, resulting in 218 core questions. For each retained question, we keep the original Persian wording, generate an English translation with GPT-5.4~\citep{openai2026gpt54}, and then manually review the translation, option fidelity, and answer correctness in a second pass. This verified real seed contributes 436 original QA instances. To expand coverage, we generated synthetic multimodal variants from the seed QAs using Claude Opus 4.7~\citep{anthropic2026opus47}, GPT-5.4~\citep{openai2026gpt54}, and Gemini 3.1 Pro \citep{google2026gemini31pro}. As with the real questions, every synthetic sample is produced in both Persian and English so that the bilingual character of the benchmark is preserved. These synthetic instances are not accepted automatically. Human supervisors review them for mathematical validity and answer consistency. The final dataset, therefore, contains 1{,}168 multimodal multiple-choice mathematics QA instances in total. We defer detailed dataset statistics, including per-split text and option lengths and correct-label distributions, to Appendix~\ref{app:statistics}.

Figure \ref{fig:seed-overview} (left) shows an overview of dataset construction. Beyond the bilingual QA items themselves, VAMPS also includes a diagnostic regime; more specifically, for the English subset of questions, four fixed visualization solution layers per question (for visual-aided reasoning), ordered from coarse to highly informative, were generated using Claude Opus 4.7 (see R3 in the next subsection). Figure~\ref{fig:pipeline} shows two QA pairs with their corresponding model's solution trajectories. Sample VAMPS Questions and Agentic Interaction with Desmos tool are available in appendices \ref{app:sample_question} and \ref{app:catalog}. Within the dataset, some QA pairs include plots in the question body and/or among the answer options, while some pairs are text-only in exam form but were selected because plotting is still a natural solution strategy. Overall, VAMPS is designed around these principles:

\textbf{Graph-mediated solvability.} Questions are selected such that plotting is a natural and informative solution strategy. The answer should be recoverable from visual mathematical structure, such as intersections, extrema, monotonicity, asymptotes, relative ordering, roots, or inverse relationships.\\
\textbf{Reasoning-to-perception isolation.} Each problem is designed to evaluate the transition from symbolic intent to visual evidence: the model must determine what mathematical objects should be plotted, inspect the resulting graph, and use the visual evidence to select the answer.\\
\textbf{Comparable solving regimes.} Each question can be attempted under three regimes: analytical no-tool solving, complementary visual-only solving, and tool-enabled solving, enabling us to distinguish failures of analytical reasoning, visual interpretation, and tool-mediated graph construction.

Together, these principles make VAMPS both an evaluation benchmark and a diagnostic resource. Rather than only asking which model answers more questions correctly, VAMPS is designed to reveal whether models can complete the full graph-assisted reasoning pipeline from algebraic formulation to visual interpretation to final answer selection.

\subsection{Tasks Definition}

The benchmark evaluates VLMs under three complementary solving regimes: 
\textbf{Direct Analytical Solving}, \textbf{Tool-enabled Visual Solving}, and \textbf{Provided-Visualization Solving}. 
These regimes are designed to separate general mathematical problem-solving ability from the ability to generate and comprehend visual evidence.

\textbf{R1: Direct Analytical Solving.}
This regime evaluates whether a model can solve the multiple-choice mathematics problem directly and analytically from the question statement and options. It follows the standard evaluation protocol used in prior mathematical reasoning benchmarks, but applies it to our newly collected set of graph-relevant math problems. In this setting, the model is expected to rely on its internal mathematical intuition and, when needed, analytical solution steps to arrive at an answer.\\
\textbf{Setup.} The model receives the question text and answer choices and is asked to select the correct option. No additional visual input is provided, unless a figure is already part of the original question stem or answer choices. The model does not have access to any external tools or visualizations; therefore, its answer must be produced through analytical reasoning based on the given problem statement. This setting serves as the baseline for our experiments (refer to Prompt~\ref{prompt:analytical}).

\textbf{R2: Tool-enabled Visual Solving.}
This regime evaluates whether a model can use an external graphing tool, Desmos, in our experiments, to solve VAMPS problems. The focus here is on whether the model can decide what to plot, request useful visualizations, and interpret the resulting visualization. Importantly, the model is instructed to base its reasoning only on visible evidence from the plots and not to solve the problem analytically through algebraic derivations, symbolic manipulation, or other non-visual solution steps.\\
\textbf{Setup.} The model is given a structured prompt (Prompt~\ref{prompt:tooL_use}) describing how to call the \texttt{desmos\_plot} tool and how to use available features, including expression plotting, window selection, zooming, and optional automatic labeling of points of interest (i.e., intersections, extrema, intercepts, and zeros). The model is asked explicitly to obtain at least one and at most four successful Desmos screenshots before producing a final answer. Its reasoning is constrained to visible evidence from the original problem images, answer options, and Desmos screenshots. In particular, the model is instructed not to solve the problem analytically, use external calculators, run code, or rely on non-visual derivations. Appendix Figure~\ref{fig:r2_end_to_end_trajectory} gives an end-to-end view of this R2 trajectory, from prompt intake and iterative Desmos calls to strict and soft option extraction.

\textbf{R3: Provided-Visualization Solving.}
The tool-enabled setting in R2 evaluates both whether the model can request an appropriate plot and whether it can interpret the resulting visualization. However, these two abilities are entangled: an incorrect answer may result either from requesting an unhelpful plot or from failing to interpret a useful plot. To separate these factors, R3 evaluates the model's ability to solve the problem when relevant graph-based visual evidence is already provided. In this regime, the model receives the question together with prepared visual aids and is instructed to answer using only the supplied visual evidence, rather than analytical or algebraic solution steps.\\
\textbf{Setup.}
For each problem, we prepared four visualization levels that make the graph-based evidence increasingly explicit. The evaluation is conducted as a progressive chat: the model first receives the question with the least informative visualization and must either answer from the visible evidence or request the next, more detailed visualization level. When more evidence is requested, the next visualization is added to the same conversation, preserving the accumulated context.
These layered visualizations are created by starting from a sparse global plot and then progressively adding only the next visual cue needed for disambiguation, such as a tighter crop, highlighted intersections, labeled extrema, visible asymptotes, intercepts, or ordering relationships. The process continues until the model selects an answer or all visualization levels have been shown 
(see Prompt~\ref{prompt:visualization_only}). \vspace{-10pt}

\section{Experiments and Results}
\label{sec:results}
\begin{figure}[t]
    \centering

    \begin{minipage}[t]{0.49\linewidth}

        \begin{tcolorbox}[
            enhanced, colback=blue!4, colframe=blue!60!black,
            boxrule=0.4pt, boxsep=3pt, left=4pt, right=4pt, top=3pt, bottom=3pt,
            title={\bfseries\scriptsize Question 186 (EN)},
            fonttitle=\scriptsize, coltitle=white,
            attach boxed title to top left={yshift=-1mm, xshift=3mm},
            boxed title style={colback=blue!60!black, sharp corners},
            sharp corners=south,
            width=\linewidth
        ]
        \scriptsize
        What is the absolute minimum of $f(x)=2x\sqrt{4-x^{2}}$ on $[-2,2]$?\\[1pt]
        \textbf{Options:} (1)~$0$\quad (2)~$-\sqrt{2}$\quad (3)~$-4$\quad (4)~$\sqrt{2}$
        \end{tcolorbox}

        \begin{tcolorbox}[
            enhanced, colback=gray!7, colframe=gray!55!black,
            boxrule=0.4pt, boxsep=2pt, left=4pt, right=4pt, top=2pt, bottom=2pt,
            title={\bfseries\scriptsize 1 -- Tool call},
            fonttitle=\scriptsize, coltitle=white,
            attach boxed title to top left={yshift=-1mm, xshift=3mm},
            boxed title style={colback=gray!55!black, sharp corners},
            sharp corners=south, width=\linewidth
        ]
        \scriptsize
        \textit{rationale:} ``Plot $f$ to see minimum on $[-2,2]$''\\[1pt]
        \textbullet~expr.: $2x\sqrt{4-x^{2}}$\quad
        \textbullet~bounds: $[-3,3]\!\times\![-6,6]$\\
        \textbullet~\texttt{label\_extrema = true}
        \end{tcolorbox}

        \begin{tcolorbox}[
            enhanced, colback=white, colframe=gray!55!black,
            boxrule=0.4pt, boxsep=1pt, left=1pt, right=1pt, top=1pt, bottom=1pt,
            title={\bfseries\scriptsize 2 -- Generated screenshot},
            fonttitle=\scriptsize, coltitle=white,
            attach boxed title to top left={yshift=-1mm, xshift=3mm},
            boxed title style={colback=gray!55!black, sharp corners},
            sharp corners=south, width=\linewidth
        ]
        \centering
        \includegraphics[width=\linewidth, height=3.0cm, keepaspectratio]{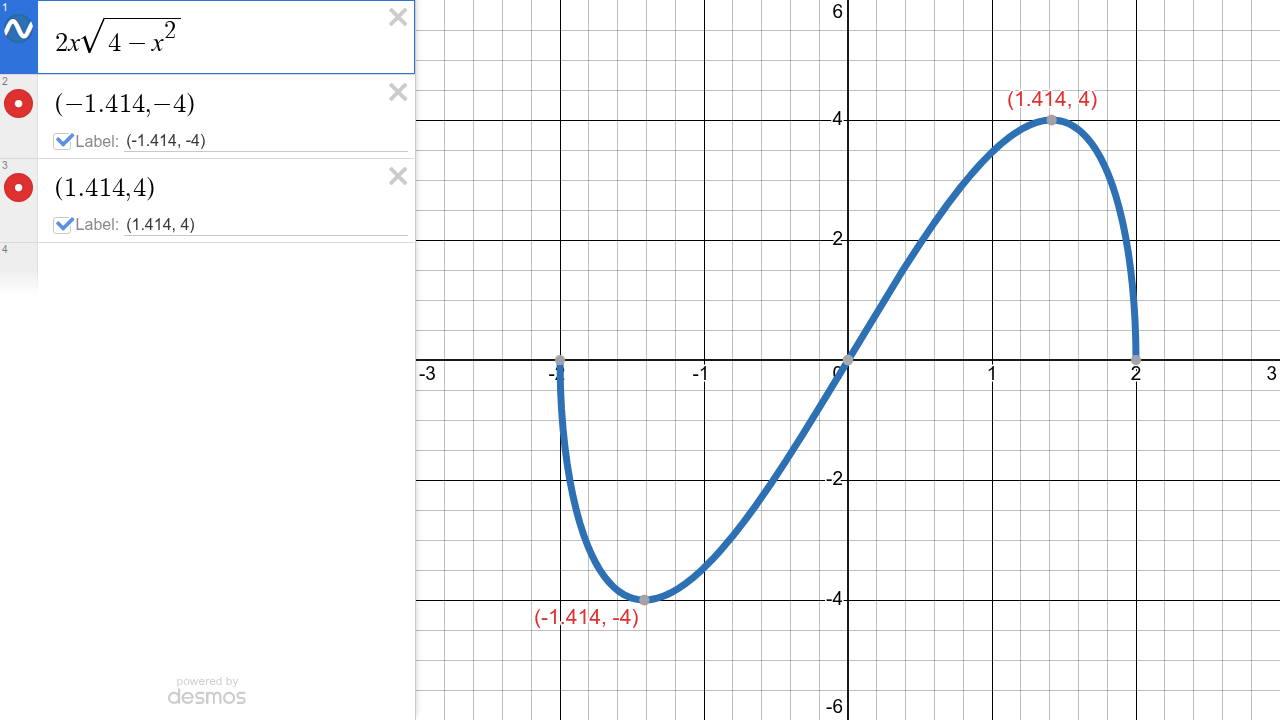}
        \end{tcolorbox}

        \begin{tcolorbox}[
            enhanced, colback=green!4, colframe=green!45!black,
            boxrule=0.4pt, boxsep=2pt, left=4pt, right=4pt, top=2pt, bottom=2pt,
            title={\bfseries\scriptsize 3 -- Final answer (correct)},
            fonttitle=\scriptsize, coltitle=white,
            attach boxed title to top left={yshift=-1mm, xshift=3mm},
            boxed title style={colback=green!45!black, sharp corners},
            sharp corners=south, width=\linewidth
        ]
        \scriptsize
        \textit{summary:} ``Plotted $f(x)=2x\sqrt{4-x^{2}}$. Visible extremum label shows minimum at $(-1.414,-4)$.''\\
        \textbf{selected:} \texttt{"3"}\quad\textbf{gt:} \texttt{"3"}\quad{\color{green!50!black}\checkmark}
        \end{tcolorbox}
    \end{minipage}\hfill
    \begin{minipage}[t]{0.49\linewidth}

        \begin{tcolorbox}[
            enhanced, colback=blue!4, colframe=blue!60!black,
            boxrule=0.4pt, boxsep=3pt, left=4pt, right=4pt, top=3pt, bottom=3pt,
            title={\bfseries\scriptsize Question 199 (EN)},
            fonttitle=\scriptsize, coltitle=white,
            attach boxed title to top left={yshift=-1mm, xshift=3mm},
            boxed title style={colback=blue!60!black, sharp corners},
            sharp corners=south,
            width=\linewidth
        ]
        \scriptsize
        If $\tfrac{\pi}{2}<x<\pi$, evaluate $\dfrac{\tan x}{\sqrt{1+\tan^{2}x}}\!\left(\dfrac{1}{\sin x}-\sin x\right)$.\\[1pt]
        \textbf{Options:} (1)~$-\cos^{2}x$\,\, (2)~$-\cos x$\,\, (3)~$\cos^{2}x$\,\, (4)~$\cos x$
        \end{tcolorbox}

        \begin{tcolorbox}[
            enhanced, colback=gray!7, colframe=gray!55!black,
            boxrule=0.4pt, boxsep=2pt, left=4pt, right=4pt, top=2pt, bottom=2pt,
            title={\bfseries\scriptsize 1 -- Tool call},
            fonttitle=\scriptsize, coltitle=white,
            attach boxed title to top left={yshift=-1mm, xshift=3mm},
            boxed title style={colback=gray!55!black, sharp corners},
            sharp corners=south, width=\linewidth
        ]
        \scriptsize
        \textit{rationale:} ``Compare expression with $\cos^{2}x$, $-\cos^{2}x$, $\cos x$''\\[1pt]
        \textbullet~exprs: target \& candidates\quad
        \textbullet~bounds: $[1.5,3.2]\!\times\![-1.2,1.2]$\\
        \textit{(Two plots issued; second shown.)}
        \end{tcolorbox}

        \begin{tcolorbox}[
            enhanced, colback=white, colframe=gray!55!black,
            boxrule=0.4pt, boxsep=1pt, left=1pt, right=1pt, top=1pt, bottom=1pt,
            title={\bfseries\scriptsize 2 -- Generated screenshot},
            fonttitle=\scriptsize, coltitle=white,
            attach boxed title to top left={yshift=-1mm, xshift=3mm},
            boxed title style={colback=gray!55!black, sharp corners},
            sharp corners=south, width=\linewidth
        ]
        \centering
        \includegraphics[width=\linewidth, height=3.0cm, keepaspectratio]{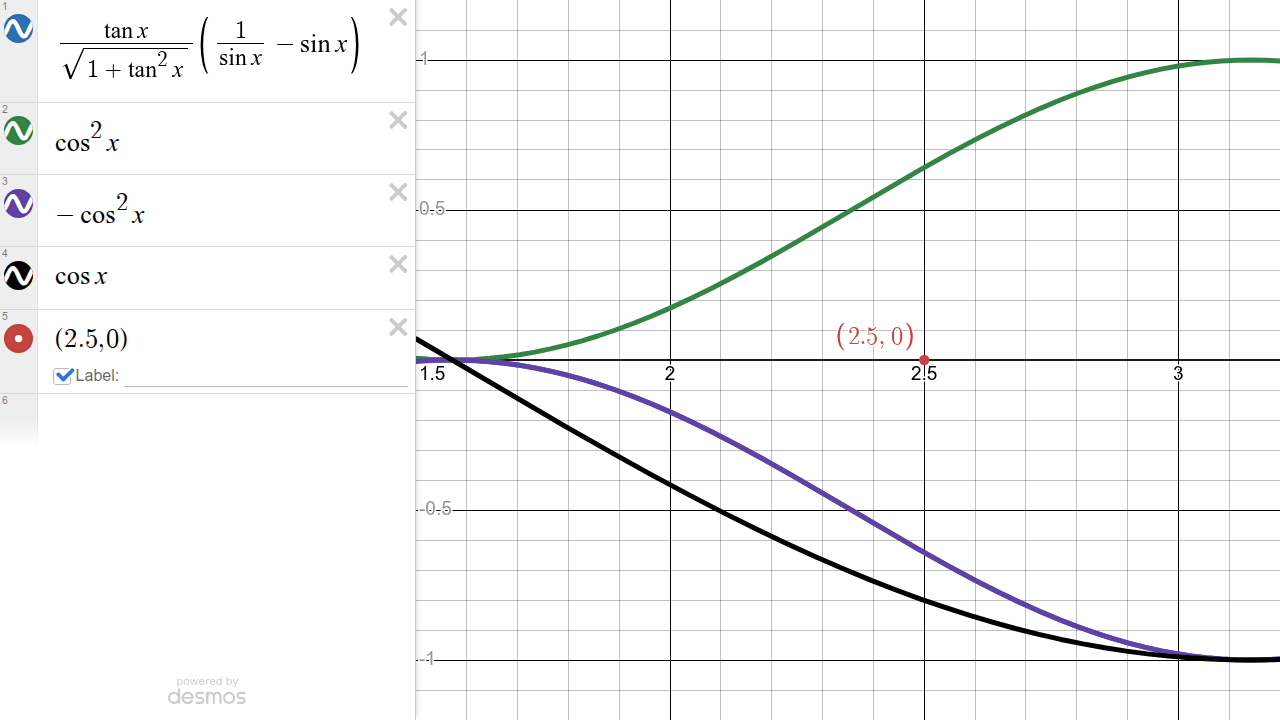}
        \end{tcolorbox}

        \begin{tcolorbox}[
            enhanced, colback=red!3, colframe=red!55!black,
            boxrule=0.4pt, boxsep=2pt, left=4pt, right=4pt, top=2pt, bottom=2pt,
            title={\bfseries\scriptsize 3 -- Final answer (incorrect: sign / quadrant misread)},
            fonttitle=\scriptsize, coltitle=white,
            attach boxed title to top left={yshift=-1mm, xshift=3mm},
            boxed title style={colback=red!55!black, sharp corners},
            sharp corners=south, width=\linewidth
        ]
        \scriptsize
        \textit{summary:} ``Blue expression curve appears to coincide with $\cos x$, both reaching $-1$ near $x=\pi$.''\\
        \textbf{selected:} \texttt{"4"}\quad\textbf{gt:} \texttt{"1"}\quad{\color{red!70!black}$\times$}
        \end{tcolorbox}
    \end{minipage}

    \caption{Illustrative VAMPS trajectories for tool-enabled runs on two questions, both solved by Claude Opus~4.7. \emph{Left:} on Question~186, the model issues a single \texttt{desmos\_plot} call with \texttt{label\_extrema} enabled, reads the labeled extremum coordinate from the returned screenshot, and selects the correct option. \emph{Right:} on Question~199, the model plots the target expression alongside several answer candidates, but misreads the visual overlap on $(\pi/2,\pi)$ and selects incorrect option.}
    \vspace{-10pt}
    \label{fig:pipeline}
\end{figure}

We evaluate a variety of models with different sizes and architectures, spanning the Qwen, Gemma, Ministral, Nemotron open-weight family of models, as well as Gemini, Claude, and GPT proprietary models; namely: Qwen2.5-VL 7B, Qwen3-VL 8B, Nemotron Nano 12B 2 VL, Ministral3 8B, Ministral3 14B, Gemma3 12B, Gemma3 27B, Gemma4 26B, Gemma4 31B, Qwen3-VL 32B, Qwen3.5 27B, Qwen3.5 35B, Qwen3.5 397B, Gemini 2.5 Flash, Claude Sonnet 4.6, Claude Opus 4.7, GPT-4o, and GPT-5.4~\citep{bai2025qwen25vl, bai2025qwen3vl, gemmateam2025gemma3, google2026gemma4, liu2026ministral3, nvidia2025nemotronnano, google2025gemini25flash, anthropic2026sonnet46, anthropic2026opus47, openai2024gpt4o, openai2026gpt54}. Throughout the paper, we use \emph{reasoning} to describe the symbolic-to-visual mathematical problem-solving process. Several models support an extended internal deliberation mode, referred to as \emph{thinking mode}. We disable it across all models for fairness and auditability: in early experiments, models with hidden deliberation showed a tendency to fall back to analytical solving even when instructed otherwise, with no way to detect or penalize this. Instead, models are prompted to externalize their step-by-step reasoning in the visible output, ensuring a consistent and verifiable evaluation protocol across all models.  Across all evaluations, answers are extracted from a structured JSON block that models are required to include in their final output. We report the accuracy of the model's final selected option. For tool-enabled runs, we additionally report \emph{filtered accuracy} obtained via a VLM-as-a-judge protocol. The judge ({Qwen3-VL-30B-A3B}) inspects each model response alongside its associated visual evidence and flags answers if the model appears to have solved the problem analytically rather than grounding its answer in the produced visualization (see Prompt \ref{prompt:vlm_as_judge}). Filtered accuracy is, therefore, typically lower than raw accuracy and serves as a stricter measure of whether a correct answer was genuinely reached through visual reasoning. Please refer to Appendix \ref{app:experimental-settings} regarding the other experimental settings.

\subsection{Main Regimes Comparison}

Tables~\ref{tab:main-results}, \ref{tab:synthetic-results}, \ref{tab:main-results-soft}, and \ref{tab:synthetic-results-soft} summarize the core solving regimes comparison in VAMPS. Table~\ref{tab:main-results} reports the comparison between R1 and R2 on the original Konkour subset, Table~\ref{tab:synthetic-results} repeats the same evaluation on the synthetic subset, and Tables~\ref{tab:main-results-soft} and \ref{tab:synthetic-results-soft} repeat the same evaluations with a relaxed extractor that scans the full model response for the intended option when the required JSON output is malformed or absent, serving as a robustness check for instruction-following failures unrelated to reasoning ability. Across the tables, the main pattern is clear: direct analytical solving is typically stronger than tool-enabled solving, and judge-filtered accuracy is lower still. On the original seed, analytical no-tool accuracy exceeds raw tool-enabled accuracy for 15 of 18 models in English and 16 of 18 models in Persian. On the synthetic seed, the same trend holds for 12 of 14 and 13 of 14 matched models in each language, respectively. Claude Opus 4.7 is the strongest overall model in both regimes and both languages, reaching 98.2\% analytical accuracy on both the English and Persian Konkour-seed splits. The gap between raw tool accuracy and judge-filtered accuracy is also informative: it suggests that some apparently correct R2 answers are not grounded robustly enough in the produced visual evidence to survive stricter trace-aware evaluation. Tables~\ref{tab:main-results-soft} and \ref{tab:synthetic-results-soft} show that this overall trend is stable under a softer extractor as well. There are some exceptions, for example Qwen3.5 35B-A3B in English, whose strict analytical score is depressed by answer-format failures rather than by a genuine inability to solve the questions analytically: under softer extraction, its analytical accuracy rises from 64.7\% to 95.4\% on the original seed and from 60.9\% to 91.0\% on the synthetic seed. More detailed results are provided in Appendix~\ref{app:tables}.

\begin{table*}[t]
    \centering
    \begin{minipage}[t]{0.61\textwidth}
        \vspace{0pt}
        \centering
        \scriptsize
        \setlength{\tabcolsep}{2.8pt}
        \renewcommand{\arraystretch}{1.01}
        \captionof{table}{Konkour-subset results across direct analytical solving and tool-enabled visual solving. Values are accuracies in percent; ``Judge'' denotes the VLM-as-a-judge filtered accuracy. \textbf{Claude Opus 4.7 is the strongest model across all evaluation regimes and languages.}}
        \label{tab:main-results}
        \begin{tabular}{@{}lcccccc@{}}
            \toprule
            & \multicolumn{3}{c}{English} & \multicolumn{3}{c}{Persian} \\
            \cmidrule(lr){2-4} \cmidrule(lr){5-7}
            Model & Analyt. & Tool & Judge & Analyt. & Tool & Judge \\
            \midrule
            \multicolumn{7}{@{}l}{\textbf{Small}} \\
            Gemma3 12B & 68.3 & 39.0 & 33.9 & 64.7 & 43.6 & 37.6 \\
            Ministral3 14B & 75.7 & 59.6 & 56.0 & 67.0 & 53.2 & 48.2 \\
            Ministral3 8B & 67.4 & 58.7 & 50.9 & 56.9 & 50.5 & 46.8 \\
            Nemotron Nano 12B 2 VL & 52.8 & 36.7 & 29.8 & 39.9 & 29.8 & 23.9 \\
            Qwen2.5-VL 7B & 55.0 & 36.7 & 31.2 & 48.2 & 32.6 & 26.6 \\
            Qwen3-VL 8B & 92.2 & 50.0 & 48.6 & 83.0 & 49.1 & 48.2 \\
            \addlinespace[2pt]
            \multicolumn{7}{@{}l}{\textbf{Medium}} \\
            Gemma3 27B & 77.5 & 50.5 & 46.3 & 74.8 & 46.8 & 41.7 \\
            Gemma4 26B-A4B & 90.4 & 85.3 & 76.6 & 95.4 & 81.2 & 72.9 \\
            Gemma4 31B & 97.2 & 89.5 & 85.8 & 97.2 & 84.4 & 81.7 \\
            Qwen3-VL 32B & 92.2 & 74.3 & 71.6 & 84.4 & 71.6 & 69.7 \\
            Qwen3.5 27B & 95.9 & 83.5 & 79.8 & 95.9 & 82.6 & 78.4 \\
            Qwen3.5 35B-A3B & 64.7 & 76.2 & 72.0 & 85.8 & 76.6 & 70.6 \\
            \addlinespace[2pt]
            \multicolumn{7}{@{}l}{\textbf{Frontier / Large}} \\
            Claude Opus 4.7 & \textbf{98.2} & \textbf{93.6} & \textbf{92.2} & \textbf{98.2} & \textbf{92.7} & \textbf{91.3} \\
            Claude Sonnet 4.6 & \textbf{98.2} & 88.5 & 86.2 & 95.9 & 88.5 & 88.1 \\
            GPT-4o & 49.1 & 49.5 & 48.2 & 45.9 & 49.5 & 49.1 \\
            GPT-5.4 & 89.0 & 66.5 & 55.0 & 89.5 & 70.6 & 56.4 \\
            Gemini 2.5 Flash & 90.4 & 78.4 & 73.8 & 83.0 & 84.4 & 78.0 \\
            Qwen3.5 397B-A17B & 97.2 & 88.5 & 86.7 & 96.8 & 88.5 & 88.1 \\
            \bottomrule
        \end{tabular}
    \end{minipage}\hfill
    \begin{minipage}[t]{0.36\textwidth}
        \centering
        \vspace{0pt}
        \includegraphics[width=0.85\linewidth]{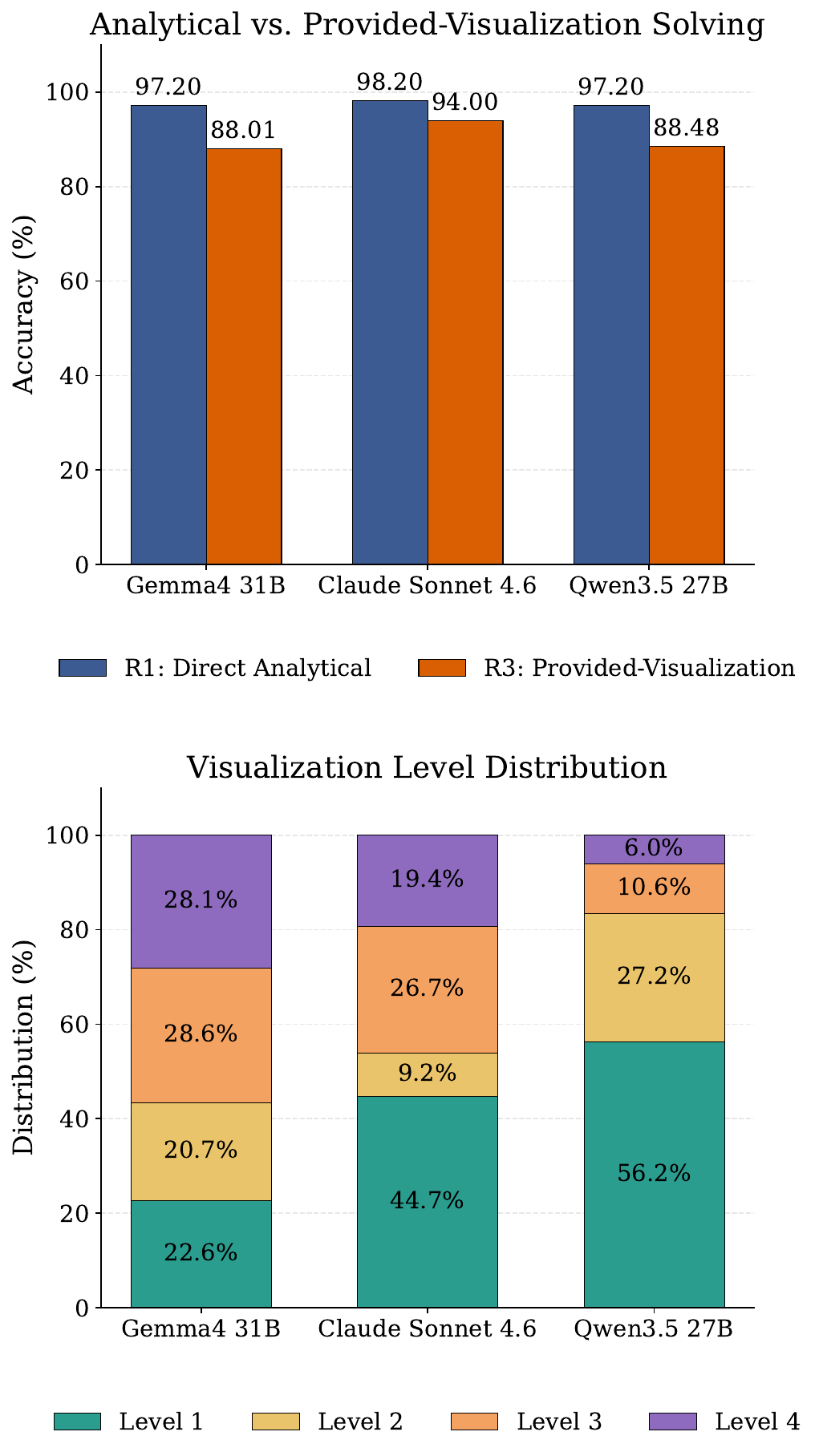}
        \captionof{figure}{Complementary probe: analytical vs.\ provided-visualization solving and the visualization-level mix used in R3.}
        \label{fig:r1_r3_visualization_distribution}
    \end{minipage}
\end{table*}

\subsection{Complementary Provided-visualization Solving}
The Provided-Visualization Solving regime, R3, is a complementary experiment that is diagnostically important. It lets us assess whether poor tool-enabled performance comes from weak visual interpretation in general or from failures that arise earlier in the tool-enabled pipeline, such as poor plot requests, weak refinement policies, or brittle handoffs between tool use and answer extraction. In this probe, we prepared four fixed visualization layers for the English subset of Konkour-seed, ranging from minimally informative plots to highly detailed ones that expose the decisive cue clearly. Figure~\ref{fig:r1_r3_visualization_distribution} shows that R3 remains below the analytical baseline for all three representative models: Gemma4 31B drops from 97.20\% in R1 to 88.01\% in R3, Claude Sonnet 4.6 drops from 98.20\% to 94.00\%, and Qwen3.5 27B drops from 97.20\% to 88.48\%. This suggests two non-exclusive explanations. 

First, these models are already very strong at analytical derivation, so even a useful fixed visualization may not outperform their symbolic baseline. Second, some models appear to fall back to analytical solving instead of requesting a more informative view. The layer-distribution histogram is consistent with this: Qwen3.5 27B stops at Level 1 in 56.2\% of runs and reaches Level 4 only 6.0\% of the time, whereas Gemma4 31B spreads its requests more evenly (22.6\%, 20.7\%, 28.6\%, 28.1\% for Levels 1--4). Notably, the fact that models do request Level 4 detail when needed --- and that R3 accuracy is non-trivially above R2 for most models --- suggests that fully-labeled visualizations are both necessary and sufficient for a meaningful fraction of questions, validating the detail level available in Desmos tool in R2. At the same time, R3 is often stronger than the corresponding tool-enabled regime, which means that visual interpretation alone is not the whole story. Claude Sonnet 4.6 rises from 88.5\% raw tool accuracy (86.2\% judge-filtered) in Table~\ref{tab:main-results} to 94.00\% in Figure~\ref{fig:r1_r3_visualization_distribution}, and Qwen3.5 27B rises from 83.5\% (79.8\% judge-filtered) to 88.48\%. Gemma4 31B is roughly tied with raw tool-enabled accuracy, 89.5\% versus 88.01\%, and still exceeds the judge-filtered tool score of 85.8\%. Taken together, these comparisons indicate that several failure modes discussed in Section~\ref{sec:discussion} arise specifically in the full tool-enabled setting and need not apply in the visual-only probe. In other words, once a useful visualization is supplied directly, the model can sometimes use it effectively, but it may still fail when it must generate and refine the visualizations on its own.
\vspace{-10pt}
\section{Discussion: Why Visual Tool Use Could Hurt}
\label{sec:discussion}
\label{sec:analysis}

VAMPS is built around a counterintuitive question: if plotting can help humans, why might a model perform worse when asked to reason through a self-generated plot? Our hypothesis is that tool-enabled visual reasoning introduces a \emph{reasoning-to-perception handoff bottleneck}. Analytical solving lets the model remain in the symbolic text domain, where recent models are heavily trained. Tool-enabled solving (R2) requires additional competencies: forming usable tool calls, obtaining visually informative renderings, and reading those renderings accurately enough to find the answer. Examining the questions on which most models converge on the wrong tool-enabled answer, namely those for which at least 10 of the evaluated models err, we identify four recurring failure patterns described below. 

\textit{Failure Mode 1 (FM1): failures in following instructions (for tool calls, etc.).}
The first loss occurs before visual perception begins. A model must translate mathematical intent into a valid tool interaction: correct expression syntax, the right collection of plotted objects, a sensible view window, and a decision when to stop plotting. Small deviations can be catastrophic: a malformed tool call, a wrong expression, an omitted branch, or a premature handoff to screenshot reading can render the whole trajectory unusable even when the underlying symbolic idea was close to correct. This bottleneck is also model-specific; some mid-sized models fail to emit a valid \texttt{desmos\_plot} request and commit to a final answer with no usable screenshot, while their textual reasoning on the same questions is otherwise competent. Interestingly, a milder but still consequential version of the same problem appears at the final answer-extraction stage. VAMPS asks models to terminate with a ``\texttt{<<FINAL>>} + JSON'' block structure and to place the predicted label in the \texttt{selected\_option} field. Qwen3.5 35B-A3B is one such example of extraction-sensitive behavior: in a nontrivial fraction of analytical runs, it outputs a malformed final answer (e.g., the \texttt{<<FINAL>>} marker is omitted). 
Under the strict extractor used for Tables~\ref{tab:main-results} and \ref{tab:synthetic-results}, such responses are scored as invalid even when the intended option is correct and is recoverable from the visible answer text. A softer extractor that scans the full response for the predicted option removes most of this artifact, raising Qwen3.5 35B-A3B's analytical English accuracy from 64.7\% to 95.4\% on the original seed and from 60.9\% to 91.0\% on the synthetic seed (see Appendix Tables~\ref{tab:main-results-soft} and \ref{tab:synthetic-results-soft}). 
See Question~52 EN in Figure~\ref{fig:catalog-1} (in Appendix~\ref{app:catalog}) for a representative example.

\textit{Failure Mode 2 (FM2): worthless tool calls and useless graphs.}
Even a syntactically valid plot request may not be helpful. Graphical questions are often sensitive to scale, zoom, and local structure: an intersection that would be decisive in a narrow window may disappear in a coarse global view; asymptotic behavior may flatten into near-straight lines; several answer choices may become visually indistinguishable. A specific recurrent pattern is \emph{Desmos auto-label over-trust}, where models relied on the count of automatically labeled points of interest (zeros, intersections, or extrema) instead of independently scanning for unlabeled features. Question~26 EN (Figure~\ref{fig:appendix-case-b}), Question~22 EN (Figure~\ref{fig:catalog-1}), are illustrative examples: in the former, Desmos's auto-zero finder silently omits a tangent root and the model trusts the count; in the latter, three intersection markers coincident at the origin are read as three distinct points. The layered visual-only probe (R3) is informative here, once a curated visualization makes the missing feature explicit, models that failed under the tool-enabled regime answer correctly, which may indicate that these models, unlike humans, have over-reliance on non-visual pretext and are not robust against noisy tool outputs.

\textit{Failure Mode 3 (FM3): correct graph, incorrect interpretation.}
Once an informative screenshot exists, the model still has to read it correctly. The decisive cue may be a relative ordering, a crossing, a branch discontinuity, the side of an asymptote, or the local shape near a boundary; current multimodal models often look strong when figures contain explicit labels or obvious salient marks, yet remain brittle when fine visual distinctions decide the answer~\citep{lu2024mathvista, roberts2025grab}. Two recurring sub-patterns dominate this mode. The first is \emph{endpoint-direction inversion}: models correctly identify the boundary values of an interval but commit to the complementary region; see Question~123 EN (Figure~\ref{fig:appendix-case-a}) for the deep-dive trajectory and Question 105~EN in Figure~\ref{fig:catalog-2}. The second is \emph{domain-of-inverse confusion}: when an inverse is requested on a restricted interval, models swap variables in the formula but reuse the original domain instead of the original function's range; Question~106 EN/FA (Figure~\ref{fig:catalog-2}) is exemplary. Question~199 EN/FA in Figure~\ref{fig:pipeline} shows a sign / quadrant variant of this kind of misread.

\textit{Failure Mode 4 (FM4): switching between algebra and plotting.}
Plotting does not eliminate symbolic reasoning; it changes when and how the model should use it. Some never commit to the visual evidence and continue generating heavy algebra even when a useful screenshot already exists. A frequent variant of this failure mode is \emph{analytic-prior hallucination}: the model invokes a mathematical fact that overrides the plot. Question~3 EN/FA (Figure~\ref{fig:catalog-4}) shows the common textbook misconception prior that ``$f$ and $f^{-1}$ always meet on $y=x$'' overriding the domain restrictions. 
Others do the reverse: they obtain a plot, abandon exact symbolic relationships too early, and answer incorrectly from an approximate coarse global view. In both cases, the issue is unstable modality switching rather than a simple lack of competence in algebra or perception alone.

\textbf{Convergent wrong answers as a signal.}
Across the high-failure questions, the wrong answers are not random. Many have a clear majority of models converging on the same wrong option, and that option is typically the one that follows from a fast analytic shortcut taken without checking the plot, or from reading a Desmos auto-label without scanning for unlabeled visual features. This convergence is itself diagnostic: it suggests systematic shortcomings shared across model families rather than independent perceptual noise. Appendix~\ref{app:catalog} catalogues every high-failure question we observe, organized by dominant failure mode.

\textbf{Why humans may show the opposite pattern.}
Previous literature on multimedia learning suggests that humans often learn more effectively when information is presented through multiple modalities rather than only through text or symbolic notation. This is commonly explained through the {dual-channel assumption}, which states that humans have separate information-processing systems for visual or pictorial material and auditory or verbal material, and the {limited-capacity assumption}, which states that only a limited amount of information can be processed in each channel at one time \citep{mayer2002multimedia}. From this perspective, graphs can support human problem solving by shifting part of the reasoning burden from purely verbal or symbolic manipulation to visual inspection. However, current AI models may not follow the same cognitive pattern: they are typically optimized more strongly for textual-symbolic continuation than for human-like multimodal perception. As a result, even when a graph makes properties such as trends, intersections, ordering, or relative position visually available, models may still struggle to use this evidence reliably.



\section{Conclusion}
\label{sec:conclusion}
This work introduces VAMPS, a benchmark for studying a specific and important skill in multimodal LLMs: the ability to use tool-enabled visual reasoning to complement analytical reasoning when solving mathematical problems --- a workflow central to real-world scientific and engineering practice, where experts routinely rely on visualization tools to validate hypotheses and guide decisions. Our experiments indicate that even on problems for which plotting should intuitively help, current models often struggle to benefit reliably from visual tool use. By comparing analytical and tool-enabled regimes, and using complementary ready-made visual probes to diagnose where failures originate, VAMPS isolates the point at which analytical reasoning must hand off to perception, a point where many current agentic LLM systems appear fragile.

VAMPS broader contribution is methodological. It treats tool use not as an automatic capability gain, but as a change in representational interface that can either help or hurt, depending on how well the model executes and interprets the tool interaction. For mathematical and scientific reasoning in particular, this distinction matters: a tool that converts symbolic structure into an image may expose weaknesses in visual grounding. VAMPS is designed to make that difference measurable and auditable, while also contributing a real-world, multimodal, and bilingual dataset grounded in authentic educational problems from the Iranian University Entrance Exam. The failure patterns in VAMPS suggest concrete improvement directions. Tighter tool-call validation and self-checking loops would address premature handoffs and bad plot requests. Training models to intelligently scan for unlabeled visual features, rather than trusting auto-generated annotations, would reduce over-trust failures. Iterative plot-refinement policies that zoom or relabel when the current view is uninformative could close much of the accuracy gap between tool-enabled (R2) and provided-visualization solving (R3). Finally, explicitly rewarding visual grounding over analytical shortcuts during training could address the tendency to override plot evidence with memorized symbolic facts.

\bibliographystyle{plainnat}
\bibliography{ref}

\newpage
\appendix
\section{Code and Data Availability and Reproducibility}
\label{app:availability}
The benchmark code accompanying this paper is publicly available at \href{https://github.com/vampsbenchmark/VAMPS}{https://github.com/vampsbenchmark/VAMPS}. The dataset accompanying this paper is publicly available at \href{https://huggingface.co/datasets/VAMPSBenchmark/VAMPS}{https://huggingface.co/datasets/VAMPSBenchmark/VAMPS}.

We release the benchmark code, prompts, metadata, and dataset artifacts for non-commercial research use under the {CC BY-NC 4.0} license. 
The repository and dataset card document the released assets, expected directory structure, and reproduction workflow; Appendix~\ref{app:experimental-settings} and Appendix~\ref{app:prompts} provide the core experimental settings, prompt templates.
VAMPS is an evaluation benchmark rather than a deployed system. The released assets contain mathematics questions, prompt templates, screenshots, and benchmark metadata; they do not contain personal or sensitive data. 

\section{Experimental Settings}
\label{app:experimental-settings}

All primary experiments were conducted via OpenRouter\footnote{\href{https://openrouter.ai/}{https://openrouter.ai}}, a cloud-based model provider. A small number of models unavailable on OpenRouter were evaluated locally, either on a workstation equipped with an Intel Core i9 CPU, 64\,GB of RAM, and NVIDIA RTX 3090 GPU, or on compute nodes with NVIDIA V100 GPU and 24\,GB of RAM; local models include: \texttt{Qwen2.5-VL 7B, Qwen3-VL 8B, Ministral3 8B, Ministral3 14B}. 

In all experiments, sampling was performed with a fixed temperature of 0.0, \texttt{top\_p=1.0}, and \texttt{seed=42}; the model provider was also held constant across all runs to improve reproducibility. We set the maximum model output length to 4096 tokens across all the solving regimes and allow the model to use up to four screenshots during the tool-enabled solving process.

Each individual experiment run took approximately 5-10 minutes wall-clock time (using OpenRouter), depending on model throughput (influenced by cloud provider load, model size, and architecture). The reported experiments required approximately \$300\,USD in total API costs. 

\section{Broader Impact and Limitation}
\label{app:broader-impact-limitations}
\textbf{Positive impact.} VAMPS enables rigorous evaluation of multimodal LLMs on graph-assisted mathematical reasoning, a capability gap that remains underrepresented in existing benchmarks. It also contributes a bilingual Persian/English mathematics benchmark of this kind, which supports fairer evaluation beyond English-only settings and increases the visibility of underrepresented educational contexts in AI research.

\textbf{Negative impact and misuse risk.} The main foreseeable risk is over-reliance: strong VAMPS performance could be misread as evidence of broad mathematical, visual, or agentic competence. The benchmark is intentionally narrow, so results should not be extrapolated to general reasoning ability, safety-critical decision making, or real-world scientific reliability. Because the dataset consists solely of mathematical multiple-choice problems, it does not create a direct safety-critical deployment pathway or expose sensitive content, but misinterpretation of benchmark scores remains a real concern.

\textbf{Mitigation and scope limitations.} The dataset is intended solely for research evaluation, and the paper explicitly documents scope limitations and out-of-scope uses. VAMPS studies easy-to-verify multiple-choice mathematics rather than open-ended proof generation, and it studies one especially interpretable plotting interface rather than every possible mathematical tool. This narrowness is a strength for diagnosis but a limitation for generalization. A model that struggles with Desmos-mediated visual decisions may still benefit from other tools, such as symbolic algebra systems or theorem provers. The benchmark also centers on image-mediated evidence, which means some mathematical tasks are naturally out of scope. Problems whose decisive content is purely algebraic, formal, or computational are not the right fit. Likewise, Desmos plots are approximate visualizations of mathematical objects, not formal certificates. VAMPS, therefore, is not designed and should not be interpreted as a universal test of mathematical intelligence.

An additional limitation of the current setup is that provider-specific thinking modes are disabled. This choice improves auditability because hidden internal deliberation traces are not consistently accessible across providers, but it may understate the best achievable performance of models that benefit substantially from those modes.

\section{Statistics}
\label{app:statistics}

Tables~\ref{tab:vamps_summary}--\ref{tab:vamps_option_stats} report descriptive statistics for the VAMPS dataset. Table~\ref{tab:vamps_summary} gives a high-level summary of the four language-source splits and the task format. Table~\ref{tab:vamps_text_stats} reports per-split character-length statistics of the question text, showing that English questions run slightly longer on average than Persian ones. Table~\ref{tab:vamps_option_stats} reports character-count statistics for the options text, summed across the four options per example. Table~\ref{tab:vamps_label_dist} reports the distribution of correct-option labels (1--4). 

\begin{table}[!t]
    \centering
    \small
    \setlength{\tabcolsep}{6pt}
    \renewcommand{\arraystretch}{1.10}
    \caption{Summary of the VAMPS dataset. VAMPS contains 1{,}168 multiple-choice math problems organized into four language-source splits.}
    \label{tab:vamps_summary}
    \begin{tabular}{@{}lp{0.62\linewidth}@{}}
        \toprule
        Property & Value \\
        \midrule
        Total instances    & 1{,}168 \\
        Languages          & English (EN), Persian (FA) \\
        Sources            & Konkour (real exam-style questions), Synthetic (LLM-generated; seeded from Konkour questions, human-reviewed) \\
        Splits and sizes   & Konkour (EN): 218; Konkour (FA): 218; Synthetic (EN): 366; Synthetic (FA): 366 \\
        Task format        & 4-option multiple-choice math question answering \\
        Answer space       & $\{1, 2, 3, 4\}$ \\
        Modalities         & Text with optional images for the question and each answer option \\
        \bottomrule
    \end{tabular}
\end{table}

\begin{table}[!t]
    \centering
    \small
    \setlength{\tabcolsep}{6pt}
    \renewcommand{\arraystretch}{1.05}
    \caption{Character-length statistics for the question text in VAMPS. English questions are slightly longer on average than Persian questions across both the Konkour and synthetic splits.}
    \label{tab:vamps_text_stats}
    \begin{tabular}{@{}lrrrrr@{}}
        \toprule
        Split            & $n$ & Min & Mean   & Median & Max \\
        \midrule
        Konkour (EN)     & 218 & 52  & 132.0  & 123    & 318 \\
        Konkour (FA)     & 218 & 40  & 102.9  &  97    & 255 \\
        \addlinespace
        Synthetic (EN)   & 366 & 51  & 125.3  & 118    & 323 \\
        Synthetic (FA)   & 366 & 40  & 102.7  &  94    & 282 \\
        \bottomrule
    \end{tabular}
\end{table}

\begin{table}[!t]
    \centering
    \small
    \setlength{\tabcolsep}{6pt}
    \renewcommand{\arraystretch}{1.05}
    \caption{Character-count statistics for the option text in VAMPS, computed by summing the four option-text fields per example. Median, mean, and max are reported per split; the bottom row aggregates across all 1{,}168 instances.}
    \label{tab:vamps_option_stats}
    \begin{tabular}{@{}lrrrr@{}}
        \toprule
        Split            & $n$    & Median & Mean   & Max \\
        \midrule
        Konkour (EN)     & 218    & 32     & 40.39  & 310 \\
        Konkour (FA)     & 218    & 30     & 38.77  & 310 \\
        \addlinespace
        Synthetic (EN)   & 366    & 29     & 39.85  & 194 \\
        Synthetic (FA)   & 366    & 29     & 39.27  & 194 \\
        \midrule
        All              & 1{,}168 & 30    & 39.57  & 310 \\
        \bottomrule
    \end{tabular}
\end{table}

\begin{table}[!t]
    \centering
    \small
    \setlength{\tabcolsep}{10pt}
    \renewcommand{\arraystretch}{1.10}
    \caption{Distribution of correct-option labels across the four answer choices in VAMPS. Each row reports the count for one source family; because the EN splits are direct translations of their FA counterparts, the answer keys are identical across languages within each family. 
    }
    \label{tab:vamps_label_dist}
    \begin{tabular}{@{}lcccc@{}}
        \toprule
        Source                & Option 1 & Option 2 & Option 3 & Option 4 \\
        \midrule
        Konkour (EN, FA)      &  65 &  54 &  53 & 46 \\
        Synthetic (EN, FA)    &  98 & 130 &  98 & 40 \\
        \bottomrule
    \end{tabular}
\end{table}

\begin{figure}[h]
    \centering
    \includegraphics[width=\linewidth]{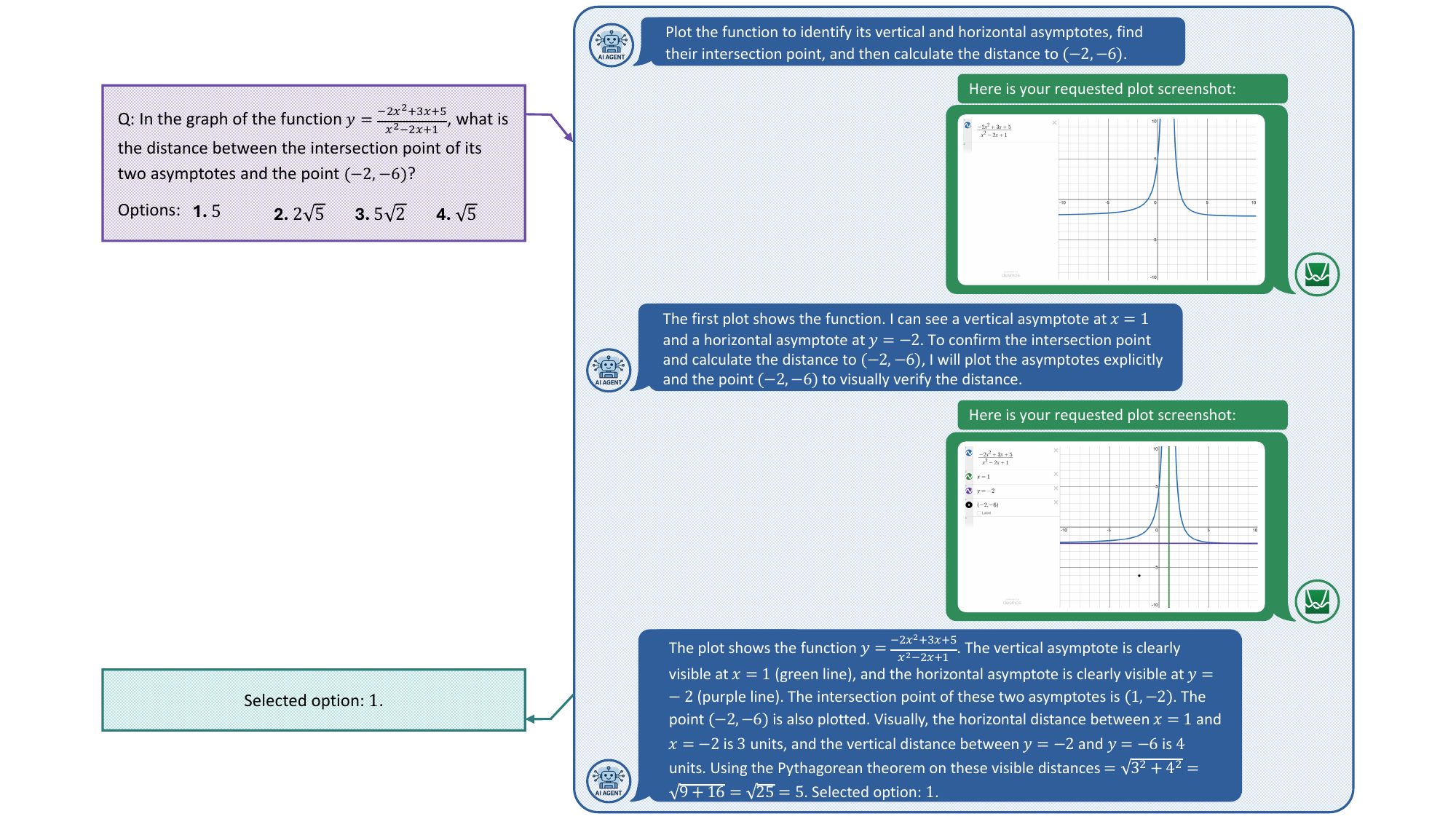}
    \caption{Example of the \textbf{R2: Tool-enabled Visual Solving} protocol. The VLM performs an agentic interaction loop with Desmos by selecting expressions to plot, inspecting the returned screenshots, and requesting additional visual evidence when needed. In this example, the model plots the rational function, identifies the vertical and horizontal asymptotes, adds the target point, and uses the resulting visual evidence to select the final answer. Note: the model output is summarized here.}
    \label{fig:R2_example}
\end{figure}

\subsection{Modality coverage}
\label{app:modality_coverage}

\begin{figure}[!t]
    \centering
    \includegraphics[width=0.92\linewidth]{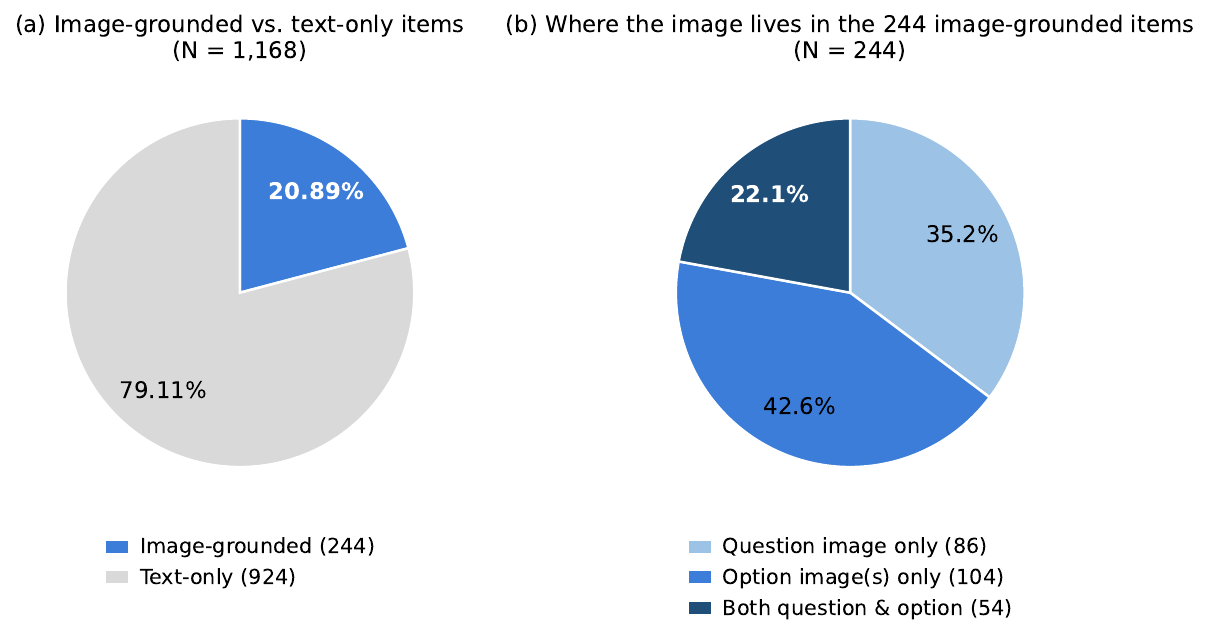}
    \caption{Modality coverage of VAMPS. \textbf{(a)}~The fraction of items that include at least one image in the input the model is given (244 of 1{,}168, or 20.89\,\%) versus the items that are text-only as posed (924 of 1{,}168, or 79.11\,\%). \textbf{(b)}~Inside the 244 image-grounded items, where the image lives: only in the question stem (86 items), only in the answer options (104), or in both (54).
    }
    \label{fig:vamps_modality_pies}
\end{figure}

Of the 1{,}168 questions in VAMPS, 244 (20.89\,\%) include at least one image as part of the input the model is given, either embedded in the question stem, attached to answer options, or both. Within this subset (Figure~\ref{fig:vamps_modality_pies}b), 86 items carry a figure only in the question, 104 items carry a figure only in the answer options, and 54 items carry figures on both sides. The remaining 924 items (79.11\,\%) are text-only as posed.

We caution that the 21\,\%\,/\,79\,\% split in Figure~\ref{fig:vamps_modality_pies}a should not be interpreted as the fraction of VAMPS that benefits from visual reasoning. The 244 image-grounded items are only a strict \emph{lower bound}: they are the cases where an image is explicitly provided to the model, making the question impossible to answer without visual parsing. The remaining 924 text-only items are not necessarily non-visual. Many describe functions, regions, inequalities, geometric objects, sequences, or parametric loci that are naturally represented as plots. Since all evaluated models can use the Desmos plotting tool, these items may also benefit from visualizing the relevant object at inference time. 

\begin{figure}[h]
\centering
\begin{minipage}{\linewidth}
\centering

\begin{tcolorbox}[
    enhanced, colback=blue!4, colframe=blue!60!black,
    boxrule=0.4pt, boxsep=3pt, left=6pt, right=6pt, top=3pt, bottom=3pt,
    title={\bfseries Question 31 (English)},
    fonttitle=\footnotesize, coltitle=white,
    attach boxed title to top left={yshift=-1mm, xshift=4mm},
    boxed title style={colback=blue!60!black, sharp corners},
    sharp corners=south, width=\linewidth
]
\footnotesize
If the graph of the function $y = f(x)$ is as shown, which of the following is the graph of $y = f'(x)$?
\end{tcolorbox}

\vspace{4pt}

\begin{tcolorbox}[
    enhanced, colback=blue!4, colframe=blue!60!black,
    boxrule=0.4pt, boxsep=3pt, left=6pt, right=6pt, top=4pt, bottom=4pt,
    title={\bfseries Question 31 (Persian)},
    fonttitle=\footnotesize, coltitle=white,
    attach boxed title to top left={yshift=-1mm, xshift=4mm},
    boxed title style={colback=blue!60!black, sharp corners},
    sharp corners=south, width=\linewidth
]
\centering
\includegraphics[width=0.8\linewidth, keepaspectratio]{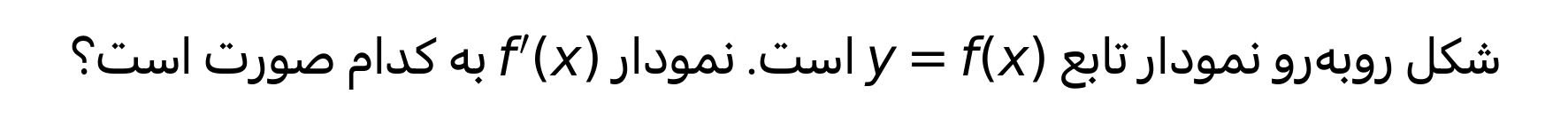}
\end{tcolorbox}

\vspace{8pt}

{\footnotesize\textbf{Question Image:} 
}
\\[3pt]
\includegraphics[width=0.42\linewidth]{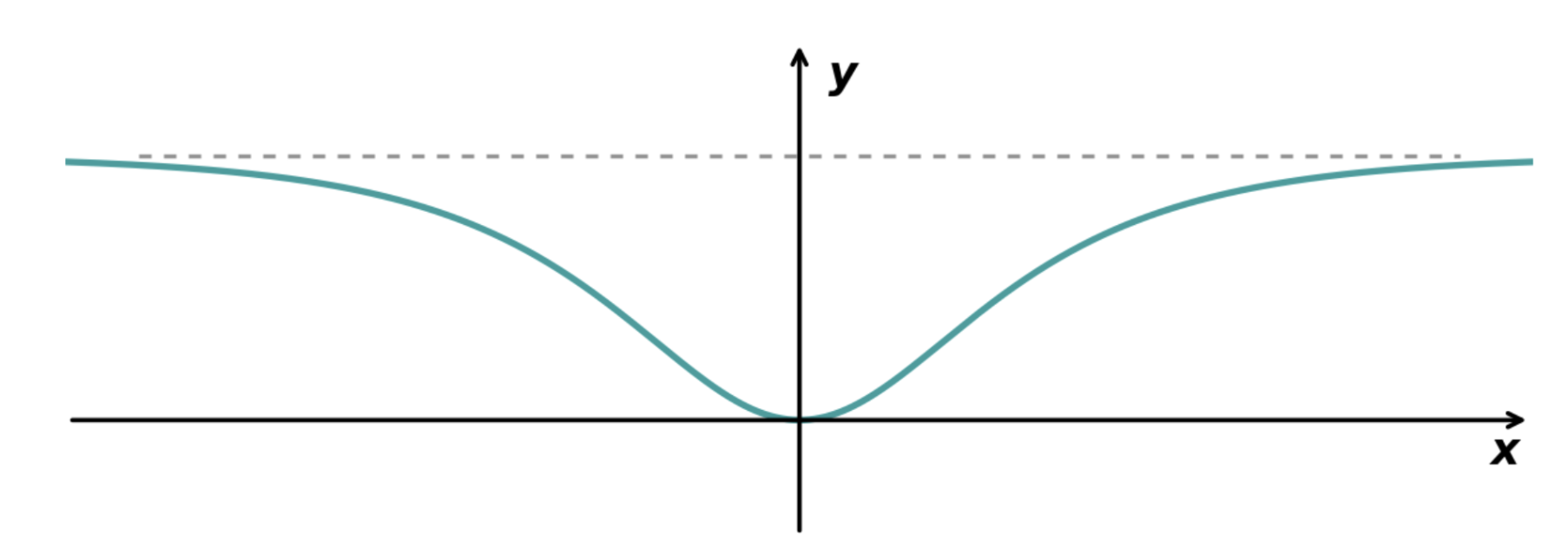}

\vspace{8pt}

{\footnotesize\textbf{Answer Options:} 
}
\\[3pt]
\setlength{\tabcolsep}{4pt}
\renewcommand{\arraystretch}{1.05}
\begin{tabular}{cccc}
\includegraphics[width=0.20\linewidth]{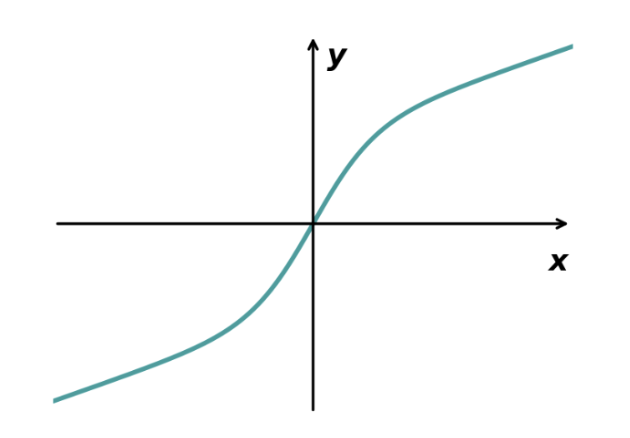} &
\includegraphics[width=0.20\linewidth]{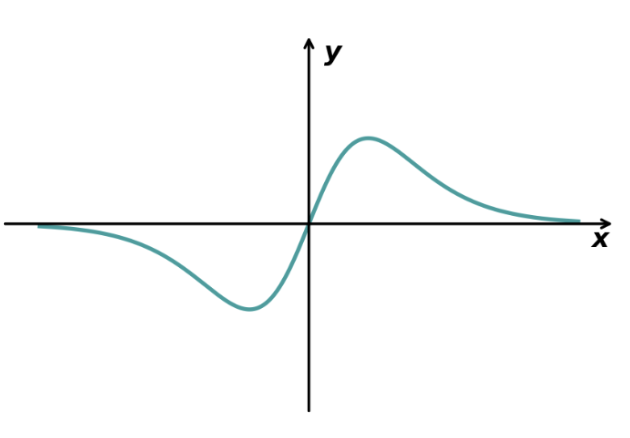} &
\includegraphics[width=0.20\linewidth]{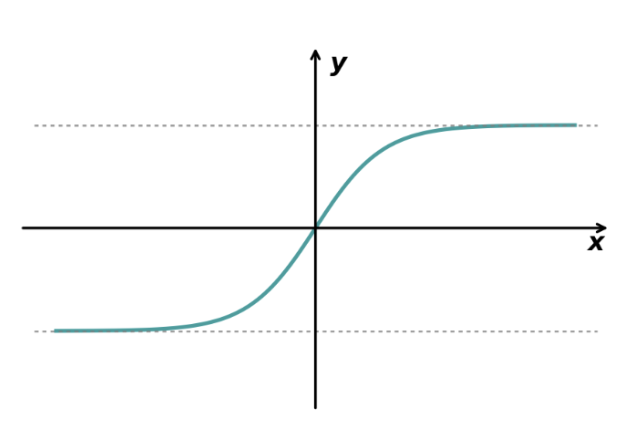} &
\includegraphics[width=0.20\linewidth]{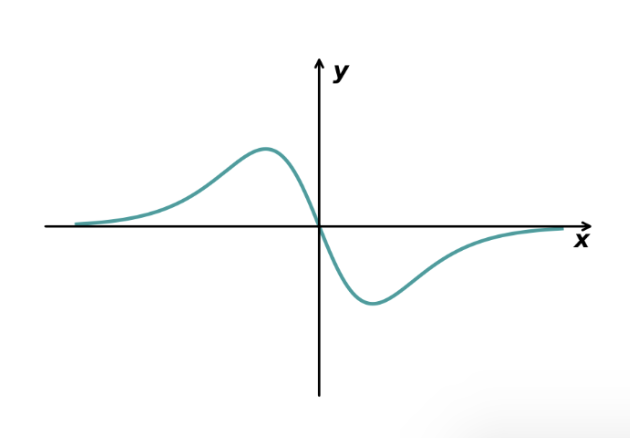} \\[2pt]
{Option 1} & {Option 2} & {Option 3} & {Option 4} \\
\end{tabular}

\end{minipage}
\caption{A representative VAMPS item (Question~31, Konkour split). The same problem is shown in English (top) and Persian (middle); the question image and the four answer-option images below are shared across the two language versions.}
\label{fig:sample_q31}
\end{figure}

\section{Sample VAMPS Questions and Agentic Interaction with Desmos tool}
\label{app:sample_question}

\begin{figure*}[!t]
    \centering
    \includegraphics[width=1\textwidth]{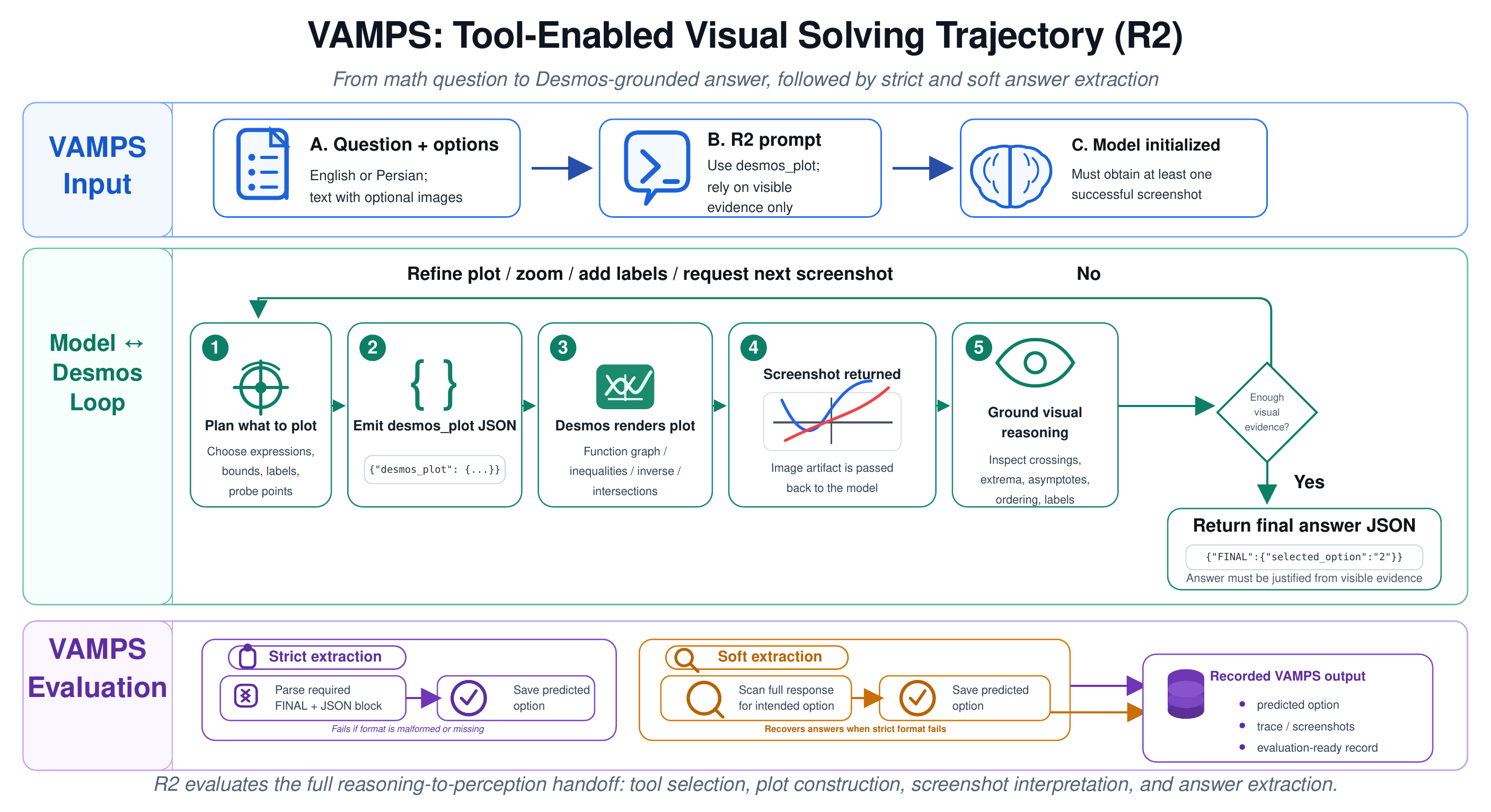}
    \caption{End-to-end VAMPS tool-enabled trajectory (R2). The pipeline begins with question intake and the R2 prompt contract, proceeds through the iterative model-Desmos loop in which the model plans plots, emits \texttt{desmos\_plot} calls, receives screenshots, and grounds its reasoning in self-generated visual evidence, and ends with a final JSON answer that is parsed by both strict and soft extraction routines before the benchmark record is saved.}
    \label{fig:r2_end_to_end_trajectory}
\end{figure*}

To give a concrete sense of the kind of multimodal multiple-choice problems in VAMPS, Figure~\ref{fig:sample_q31} shows one representative item (Question~31) from the Konkour split. We display the question in both English and Persian; the question image and the four answer-option images are shared across the two language versions, since the English split is a direct translation of its Persian counterpart, and the visual stimuli are language-agnostic. No model predictions are shown---this is the raw item as it appears to a model at evaluation time.

Moreover, Figure~\ref{fig:R2_example} demonstrates a successful example of how the model interacts with Desmos, an external graphing tool, under the R2 setting (Figure~\ref{fig:r2_end_to_end_trajectory}).

\section{Prompts}
\label{app:prompts}
\begin{promptbox}{Analytical No-tool Mode (R1)}
\label{prompt:analytical}
\promptfont
You are a helpful assistant specializing in math problem-solving for multiple-choice math questions.\\
The question and options are provided, and any attached images may contain relevant parts of the question or options. Options may be provided as text or images. Base your decision ONLY on the question, options, and any attached images. You're NOT allowed to use Python, code execution, calculators, web search, or any other external tool. Solve the problem using standard mathematical reasoning.\\
Reasoning constraints:
- Solve step by step using mathematical reasoning only.
- Each reasoning step must describe a valid mathematical observation, deduction, or calculation based on the provided question, options, and any attached images.
- Your answer MUST rely ONLY on the provided question, options, and any attached images.
- Rely on the given information to pick the option.
- If the problem is unclear or there is insufficient information, return "N/A" as the selected\_option.
- Do NOT fabricate missing information or make unsupported assumptions.\\
Output rule:
- Output ONLY: <<FINAL>>\ followed by exactly one valid JSON object. Do NOT include any markdown, code fences, explanations, or extra text.\\
Final output structure:
<<FINAL>>\ \{"steps\_summary":"\textless a concise step-by-step summary supporting your choice\textgreater","selected\_option":"\textless selected\_option\textgreater","errors":[]\}\\
Final answer notes:
- Output exactly these JSON keys and no others: steps\_summary, selected\_option, errors.
- steps\_summary must briefly describe the reasoning used to identify the answer.
- selected\_option should be "1", "2", "3", "4", or "N/A".\\
- Use "N/A" only if there's insufficient information to answer the question or the problem is unclear; explain why briefly in errors.
\end{promptbox}

\begin{promptbox}{Tool-enabled Mode (R2)}
\label{prompt:tooL_use}
\promptfont
Developer: You are a helpful assistant specializing in visual problem-solving for multiple-choice math questions.\\
Use the question text to decide what to plot, but base your justification ONLY on visible evidence from images and Desmos screenshots. Options may be provided as text or images. \\
You have exactly one allowed tool: desmos\_plot (a graphing calculator). You MUST obtain at least one successful desmos\_plot screenshot before giving the final answer. Do NOT use Python, code execution, any calculator outside desmos\_plot, web search, or any other external tool.\\
Reasoning constraints:
- Reason step by step from visual evidence only from the provided images and Desmos screenshots.
- Each reasoning step must describe a visible observation, comparison, or direct visual read-off from the supplied visualizations.
- Do NOT solve analytically (no algebra/calculus/proofs/root-finding). For instance, you may define f(x) and plot f'(x) or f'(x)=0 in Desmos, but use those only as visual evidence, not as an analytic calculus derivation.
- Light arithmetic on clearly visible values is allowed.
- Basic non-visual reasoning is allowed only to interpret visible evidence, such as matching plotted features to options, comparing positions, or doing light arithmetic on clearly visible labels. Do not use it to derive facts that are not visible in the images/screenshots.
- Your answer MUST rely ONLY on the provided images, Desmos plot screenshots, and the question/options.
- Rely on visible cues such as intersections, axis labels, grid labels, relative positions, distances, and curve behavior to pick the option.
- Do NOT verify, approximate, or complete the solution using algebra, symbolic manipulation, calculus derivations, root-finding, proofs, or heavy computation when the current image is insufficient.
- Do NOT fabricate missing information or make unsupported assumptions.
- When reading off a plot, zoom in/out until the relevant features, grid labels, and curve behavior are clearly visible before deciding.
- Do not guess when evidence is missing; Do not output "selected\_option":"N/A" while more Desmos screenshots may still be available; request another plot instead.
- If no successful Desmos screenshot is available yet, request a plot instead of finalizing.\\
Output rule:
- In each assistant message, output ONLY: the prefix followed by exactly one valid JSON object. No extra text or markdown.
(A) Tool request format (replace \textless...\textgreater\ placeholders):
<<TOOL>> \{"name":"desmos\_plot","arguments":\{"rationale":"<why this plot>","expressions":[\{"latex":"<expr1>"\}],"bounds":
\{"left":-10,"right":10,"bottom":-10,"top":10\},"probe\_points":
[\{"x":<x>,"y":<y>\}],"label\_intersections":"<bool>","label\_extrema":"<bool>",
"label\_intercepts":"<bool>","label\_zeros":"<bool>"\}\} \\
(B) Final answer format (replace \textless...\textgreater\ placeholders):
<<FINAL>> \{"steps\_summary":"<a concise step-by-step visual summary (visible observations + any tool actions) supporting your choice>","selected\_option":"<selected\_option>","errors":[]\}\\

Tool request notes:
- arguments.rationale must be a non-empty string.
- Provide a non-empty expressions list (one latex per object) and bounds; probe\_points optional.
- Use probe\_points to click/label approximate coordinates; zoom in/out by tightening/loosening bounds while maintaining the other parameters.
- Optional coordinate labels are available for POIs (gray points on the screenshots); set any of these booleans in arguments to true (only when needed to avoid clutter): label\_intersections, label\_extrema, label\_intercepts, label\_zeros.
- IMPORTANT JSON escaping: inside the JSON string \texttt{arguments.expressions[].latex}, every LaTeX backslash MUST be JSON-escaped as \texttt{\textbackslash\textbackslash} (double backslash). For example: use \texttt{\textbackslash\textbackslash frac\{1\}\{2\}x}, \texttt{\textbackslash\textbackslash cot(x)}, and \texttt{\textbackslash\textbackslash left|x\textbackslash\textbackslash right|} (not \texttt{\textbackslash frac}, not \texttt{\textbackslash cot}, not \texttt{\textbackslash left}).
- desmos\_plot expects Desmos-friendly LaTeX-ish expressions. You may find these JSON-escaped examples useful:
\quad - \texttt{\textbackslash\textbackslash left|x\textbackslash\textbackslash right|} (absolute value function)
\quad - \texttt{floor(x)} (floor function)
\quad - \texttt{\textbackslash\textbackslash sqrt[3]\{x\^{}\{5\}\}} or \texttt{x\^{}\{5/3\}} (rational exponent)
\quad - \texttt{\textbackslash\textbackslash frac\{1\}\{\textbackslash\textbackslash left(x-3 \textbackslash\textbackslash right)\^{}\{4.5\}\}} (fractional expression)
\quad - \texttt{(x+1)\^{}2\textbackslash\textbackslash left\textbackslash\textbackslash \{x>-1\textbackslash\textbackslash right\textbackslash\textbackslash \}} (domain-restricted formula)
\quad - \texttt{x=y\^{}3} (function-style expression, can be useful for inverse functions)
\quad - \texttt{x\^{}2+y\^{}2=2\^{}2 \textbackslash\textbackslash left\textbackslash\textbackslash \{0<=y<=2\textbackslash\textbackslash right\textbackslash\textbackslash \}\textbackslash\textbackslash left\textbackslash\textbackslash \{0<=x<=2\textbackslash\textbackslash right\textbackslash\textbackslash \}} (relation-style expression)
\quad - \texttt{x\^{}2=x+2} (to find solutions to equations rather than solving analytically)
\quad - \texttt{f(x)=x\^{}7-\textbackslash\textbackslash sin(x)*\textbackslash\textbackslash cos(x), then f'(x) or f'(x)=0} (define a function and plot its derivative or visually locate critical points/solutions)
\quad - \texttt{x\^{}4>x\^{}2+3} (inequality shading)
\quad - \texttt{2*\textbackslash\textbackslash log\_2 (10)} (a non-graphable numeric expression is fine, can be used to calculate and probe specific values)
- Define free parameters (a, m, k) before plotting like \texttt{a=-3}.\\
Final answer notes:
- Output exactly these JSON keys and no others: steps\_summary, selected\_option, errors.
- steps\_summary must briefly describe the visual reasoning and, if you used desmos\_plot, what you plotted; keep it visual, not algebraic.
- selected\_option should be "1", "2", "3", "4", or "N/A".
- Use "N/A" only if there is insufficient visual evidence to answer AND no further Desmos screenshot will be provided; explain why briefly in errors.
\end{promptbox}

\begin{promptbox}{Visual-only mode (R3)}
\label{prompt:visualization_only}
\promptfont
Developer: You are a helpful assistant specializing in visual problem-solving for multiple-choice math questions.\\
The question and options are provided, and any attached visualization images appearing after the options are the pre-generated visualization screenshots for the problem. Options may be provided as text or images. Base your decision only on the question, options, and supplied visualizations. You're NOT allowed to use Python, code execution, calculators, web search, or any other external tool. Do NOT solve via algebra, calculus, proofs, root-finding, or numeric computation; reason only from what is visibly shown in the images. Light arithmetic on clearly visible values is allowed.\\
Reasoning constraints:
- Reason step by step from visual evidence only.
- Each reasoning step must describe a visible observation, comparison, or direct visual read-off from the supplied images.
- Your answer MUST rely ONLY on the provided images, plot screenshots, and the question/options.
- Rely on visible cues such as intersections, axis labels, grid labels, relative positions, distances, and curve behavior to pick the option.
- Basic non-visual reasoning is allowed only to interpret visible evidence, such as matching visible features to options or doing light arithmetic on visible labels. Do not use it to derive facts missing from the images/screenshots.
- Do NOT verify, approximate, or complete the solution using algebra, symbolic manipulation, calculus derivations, root-finding, proofs, or non-visual derivations when the current image is insufficient.
- Do NOT fabricate missing information or make unsupported assumptions.
- Do NOT solve analytically (no algebra/calculus/proofs/root-finding or heavy computation).
- Do NOT use symbolic derivations, hidden formulas, or algebraic shortcuts in your answer.
- Start the final answer with <<FINAL>>\ followed by ONLY a valid JSON object defined below. Do NOT include any markdown, code fences, explanations, or extra text.
Additional interaction protocol for layered visualizations:
- Visualization images will be revealed progressively across turns.
- If the answer is not directly justified by the currently visible evidence, ask for the next visualization instead of solving it analytically.
- If the current visualization is sufficient, output the final answer using the required <<FINAL>>\ JSON format.
- If the current visualization is insufficient and you need the next higher-level visualization, reply with exactly: <<NEED\_MORE\_VISUALIZATION>>\ \{"reason":"\textless brief visual detail still missing\textgreater"\}
- Do not guess when evidence is missing.
- Do not output "selected\_option":"N/A" while more visualization stages may still be available; request the next visualization instead.
- Keep the reason concise and focused on the missing visual evidence.\\
Final output structure:
<<FINAL>>\ \{"steps\_summary":"\textless a concise step-by-step visual summary supporting your choice\textgreater","selected\_option":"\textless selected\_option\textgreater","errors":[]\}\\
Final answer notes:
- Output exactly these JSON keys and no others: steps\_summary, selected\_option, errors.
- steps\_summary must briefly describe the visual reasoning; keep it visual, not algebraic.
- selected\_option should be "1", "2", "3", "4", or "N/A".
- Use "N/A" only if there is insufficient evidence to answer the question AND no further visualization will be provided; explain why briefly in errors.
\end{promptbox}

\begin{promptbox}{VLM as a Judge}
\label{prompt:vlm_as_judge}
\promptfont
Developer: You are a helpful but strict extractor and judge for multiple-choice model outputs.\\
Inputs include the question text, answer options (and possibly plot images), and the ORIGINAL model output (and possibly a trace of messages).\\
Your tasks:
1) Extract the option label the model selected (1, 2, 3, 4).
2) Classify the ORIGINAL model output into exactly one category.

Rules (follow strictly):
1. Do NOT solve or verify the question. Use the model output to determine what the model selected.
2. Prefer an explicit option label stated by the model (use the FINAL label if multiple are mentioned).
3. If no option label is stated but the model gives an answer value/text, map it to the matching option using the provided model response only (no additional reasoning).
4. If you cannot determine a single option, use "N/A".
5. category must be exactly one of: 
    - solved\_analytically: response mainly uses algebra/calculus/symbolic or heavy computation.
    - visual\_evidence\_based: visual-cue-based solution is supported by screenshots or images with only light arithmetic on values directly visible from images or screenshots.
    - no\_definitive\_answer: response does not provide a clear final option selection.
    - other: none of the above clearly apply, explain in the 'rationale'.
6. rationale must be brief and must quote the exact substring from the model output that supports the extracted selection and category (or explain why none exists).
7. If the evidence is conflicting, incomplete, or ambiguous, do not infer; return "selected\_option":"N/A" and choose "no\_definitive\_answer" unless the category is still clearly supported by the response.\\
Output format:
Return ONLY one compact JSON object with exactly these keys:
{"rationale":"<brief>","category":"<one-category>","selected\_option":"<option-label-or-N/A>"}
No markdown, no code fences, no extra keys, no extra text beyond the required output.
\end{promptbox}

\section{Related Benchmarks Comparison}
This section summarizes how VAMPS compares with representative visual, multimodal, and mathematical reasoning benchmarks. Table~\ref{tab:benchmark-landscape-appendix} highlights the main differences in question provenance, available visual evidence, evaluation protocol, interactive tool use, and auditability.
\label{app:benchmark-comparison}
\begin{table*}[!t]
    \centering
    \scriptsize
    \setlength{\tabcolsep}{4pt}
    \renewcommand{\arraystretch}{1.08}
    \caption{Comprehensive comparison of VAMPS against representative benchmark datasets.}
    \label{tab:benchmark-landscape-appendix}
    \begin{tabularx}{\textwidth}{@{}>{\raggedright\arraybackslash}p{1.8cm}>{\raggedright\arraybackslash}p{2.3cm}>{\raggedright\arraybackslash}p{2.4cm}>{\raggedright\arraybackslash}X>{\centering\arraybackslash}p{1.2cm}>{\centering\arraybackslash}p{1.3cm}@{}}
        \toprule
        Benchmark & Question provenance & Visual evidence available to model & Typical evaluation protocol & Interactive tool loop & Auditable traces \\
        \midrule
        FigureQA \citep{kahou2017figureqa} & Synthetic scientific-style figures & Fixed plots supplied with the question & One-shot fixed-image visual reasoning & \nomark & \nomark \\
        PlotQA \citep{methani2020plotqa} & Real-world plots with synthetic QA generation & Fixed plots supplied with the question & Fixed-image plot QA, often with specialized perception modules & \nomark & \nomark \\
        ChartQA \citep{masry2022chartqa} & Web charts with human and synthetic questions & Fixed charts supplied with the question & Prompting or chart-QA models over static charts & \nomark & \nomark \\
        Geometry3K \citep{lu2021intergps} & Geometry exam-style problems & Fixed diagrams and text & Neuro-symbolic reasoning on static diagram-text pairs & \nomark & \nomark \\
        GeoQA \citep{chen2021geoqa} & Geometry exam-style problems & Fixed diagrams and text & Program-generation and multimodal numerical reasoning on static inputs & \nomark & \nomark \\
        UniGeo \citep{chen2022unigeo} & Geometry benchmark for calculation and proof & Fixed diagrams and text & Unified multimodal sequence generation on static inputs & \nomark & \nomark \\
        MathVista \citep{lu2024mathvista} & 31 source datasets & Fixed diagrams, charts, and figures & Prompting MLLMs on static visual math inputs & \nomark & \nomark \\
        VCBench \citep{wang2025vcbench} & Multi-image math problems with explicit visual dependency & Fixed sets of supplied images & Prompting LVLMs on multi-image visual-math inputs & \nomark & \nomark \\
        MV-MATH \citep{wang2025mvmath} & K--12 multi-visual math problems & Multiple fixed images interleaved with text & Prompting MLLMs on multi-visual mathematical reasoning tasks & \nomark & \nomark \\
        MathVerse \citep{zhang2024mathverse} & Public visual math sources rewritten into controlled variants & Fixed diagrams with controlled text/vision variants & Prompting on static diagrams to study modality balance & \nomark & \nomark \\
        GRAB \citep{roberts2025grab} & Synthetic and realistic graph-analysis tasks & Fixed graph visualizations & Prompting MLLMs on supplied graphs across graph-analysis tasks & \nomark & \nomark \\
        \textbf{VAMPS} & \textbf{Konkour questions plus human-reviewed synthetic multimodal variants} & \textbf{Original figures, curated visualizations, and self-generated Desmos screenshots} & \textbf{Primary analytical vs.\ tool-enabled comparison, plus complementary provided-visualization detail probes} & \yesmark & \yesmark \\
        \bottomrule
    \end{tabularx}
\end{table*}
\newpage
\section{Additional Experimental Results}
\label{app:tables}
This section provides supplementary quantitative analyses that complement the main experimental results. We include additional comparisons of model performance, token usage, and tool-use behavior across the original English and Persian questions, as well as results on the synthetic English and Persian questions.

\begin{figure*}[!t]
    \centering
    \includegraphics[width=1.0\linewidth]{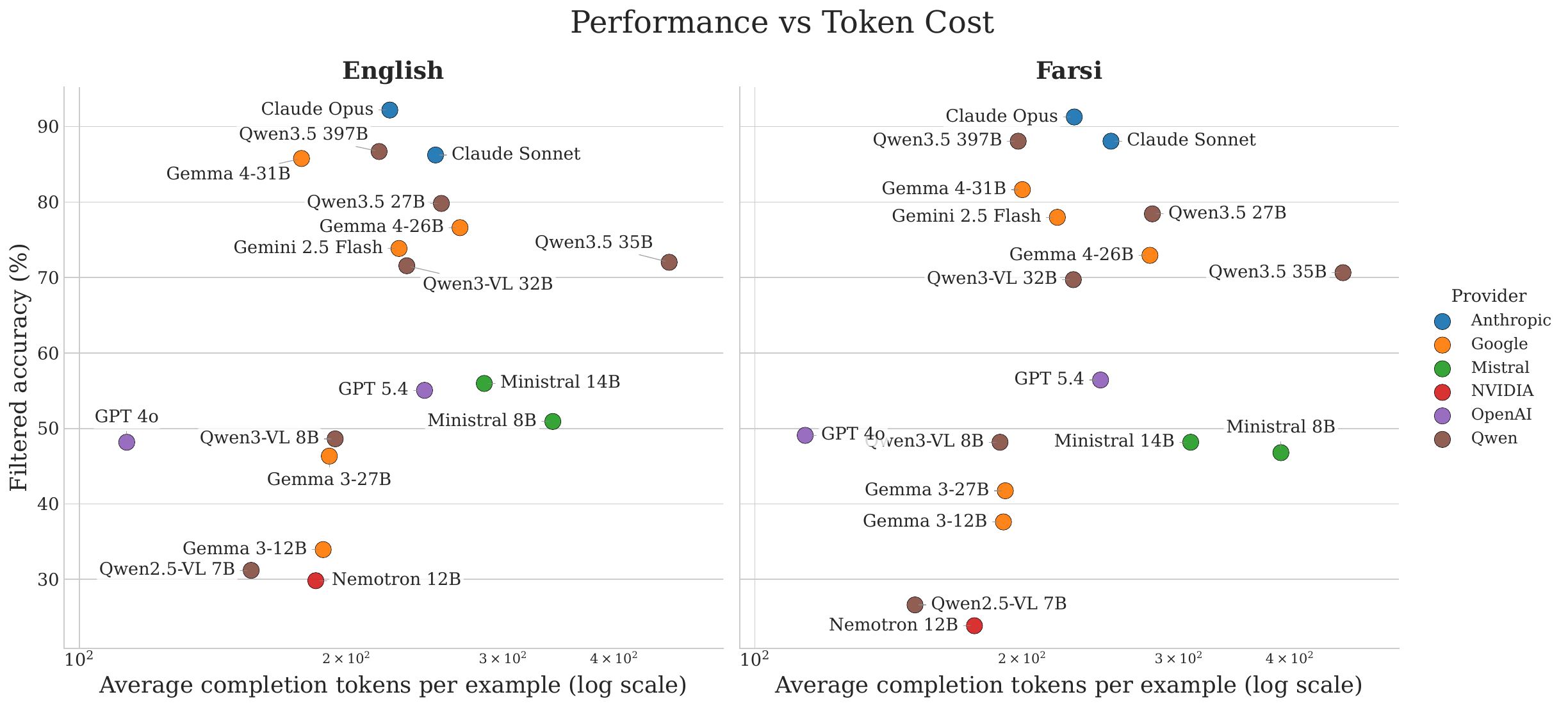}
    \caption{Performance--cost trade-off across models under \textbf{R2: Tool-enabled Visual Solving} on the original English and Persian questions. Each point represents one model, with the x-axis showing the average number of completion tokens per example on a logarithmic scale and the y-axis showing filtered accuracy. Colors indicate model providers. \textbf{The comparison highlights differences in accuracy and token efficiency across languages, showing that higher token usage does not always correspond to better performance.}}
    \label{fig:token_acc}
\end{figure*}

\begin{figure*}[!t]
    \centering
    \includegraphics[width=1.0\linewidth]{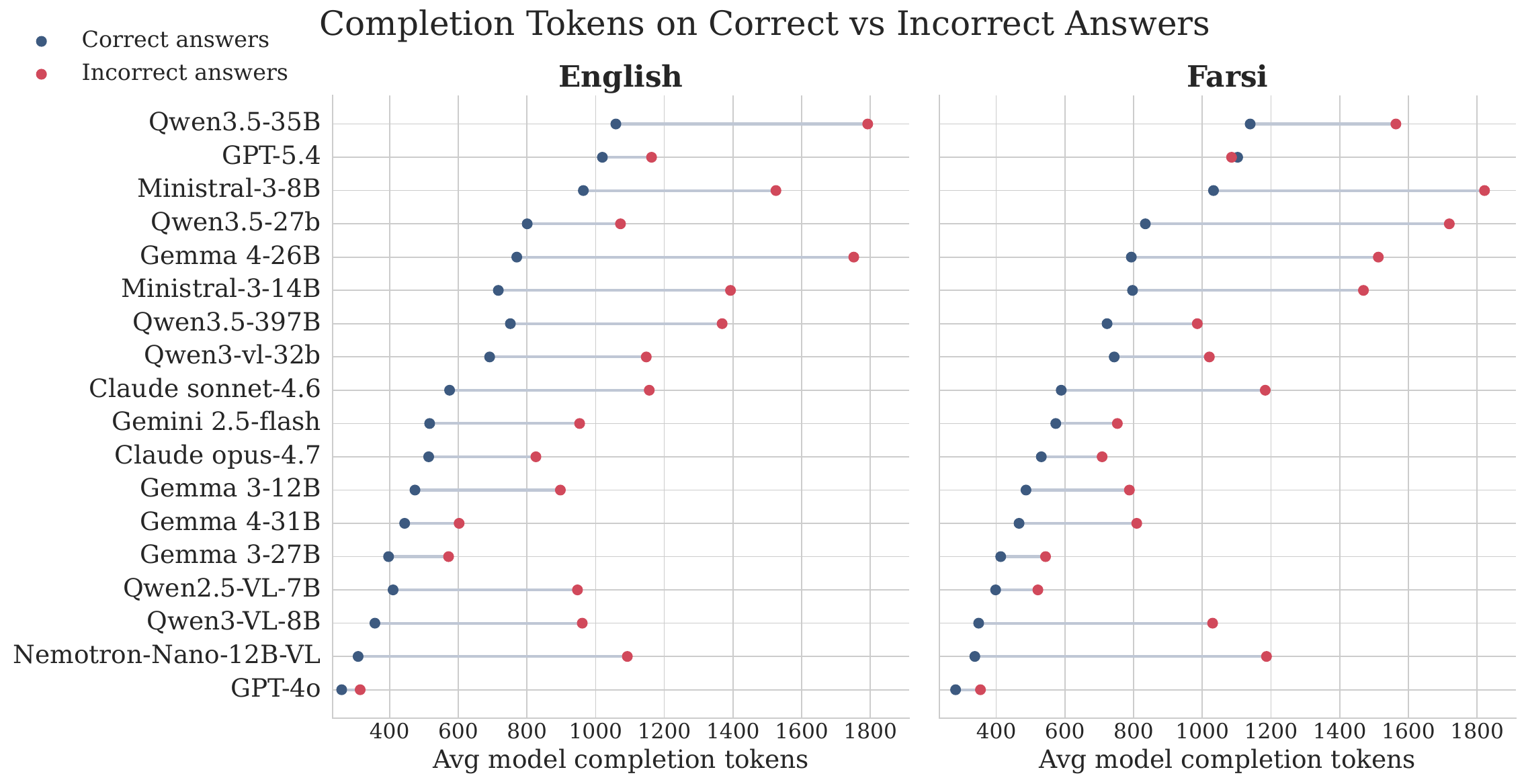}
    \caption{Average completion-token usage for correct and incorrect answers under \textbf{R2: Tool-enabled visual solving} on the original English and Persian questions. Each row corresponds to one model, with \textcolor{blue}{blue} markers indicating the average completion tokens for correctly answered examples and \textcolor{red}{red} markers indicating the average completion tokens for incorrectly answered examples. \textbf{The comparison shows that incorrect answers often require more tokens than correct answers.}}
    \label{fig:token_correct_incorrect}
\end{figure*}

\begin{figure*}[!t]
    \centering
    \includegraphics[width=1.0\linewidth]{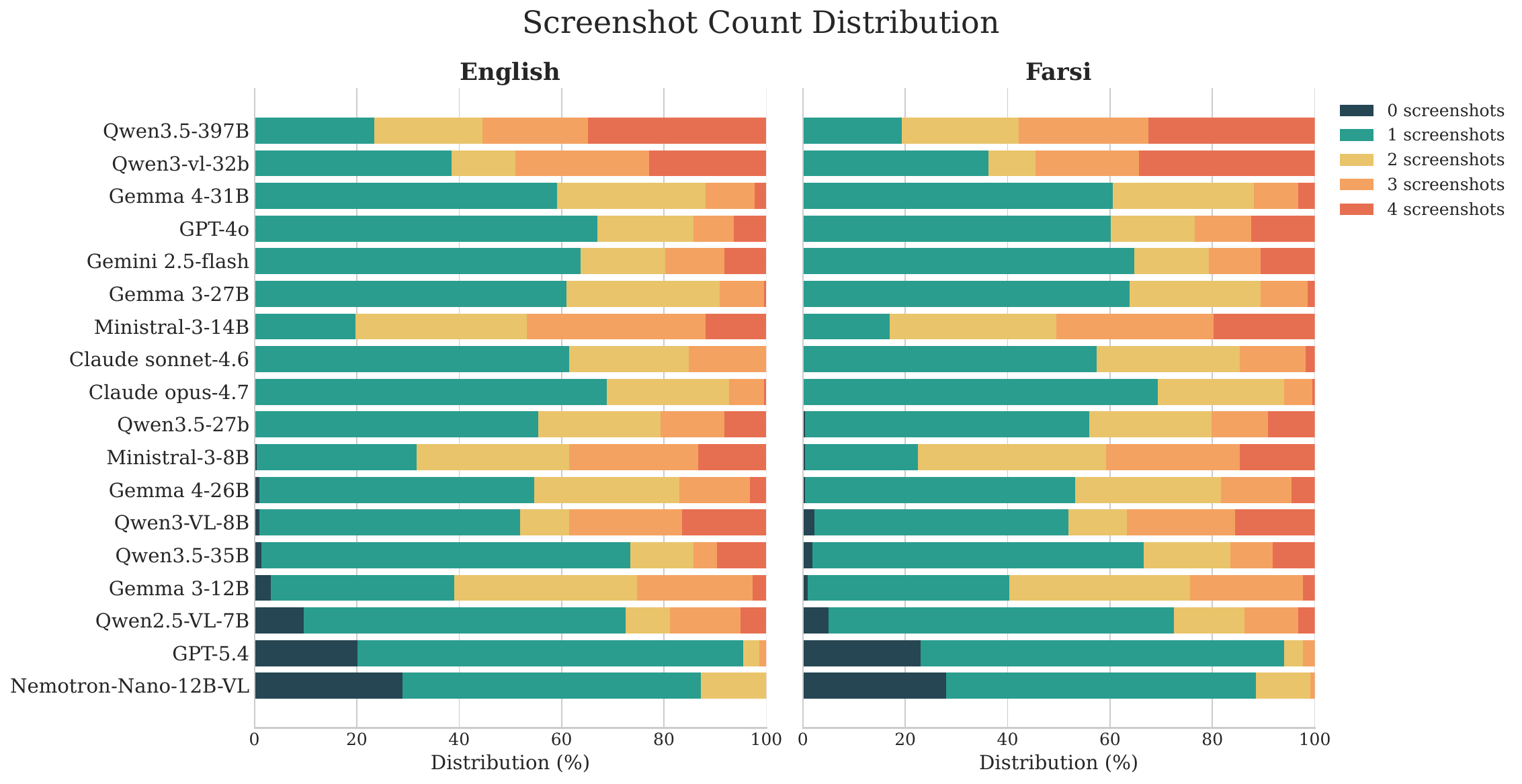}
    \caption{Distribution of screenshot counts generated under \textbf{R2: Tool-enabled visual solving} on the original English and Persian questions. Each stacked bar shows, for a given model, the percentage of examples generating 0, 1, 2, 3, or 4 screenshots during the solving process. A count of {0 screenshots} indicates that the model failed to call the tool properly, resulting in failure under the R2 protocol.}
    \label{fig:screenshot_dist}
\end{figure*}

\begin{figure*}[!t]
    \centering
    \includegraphics[width=1.0\linewidth]{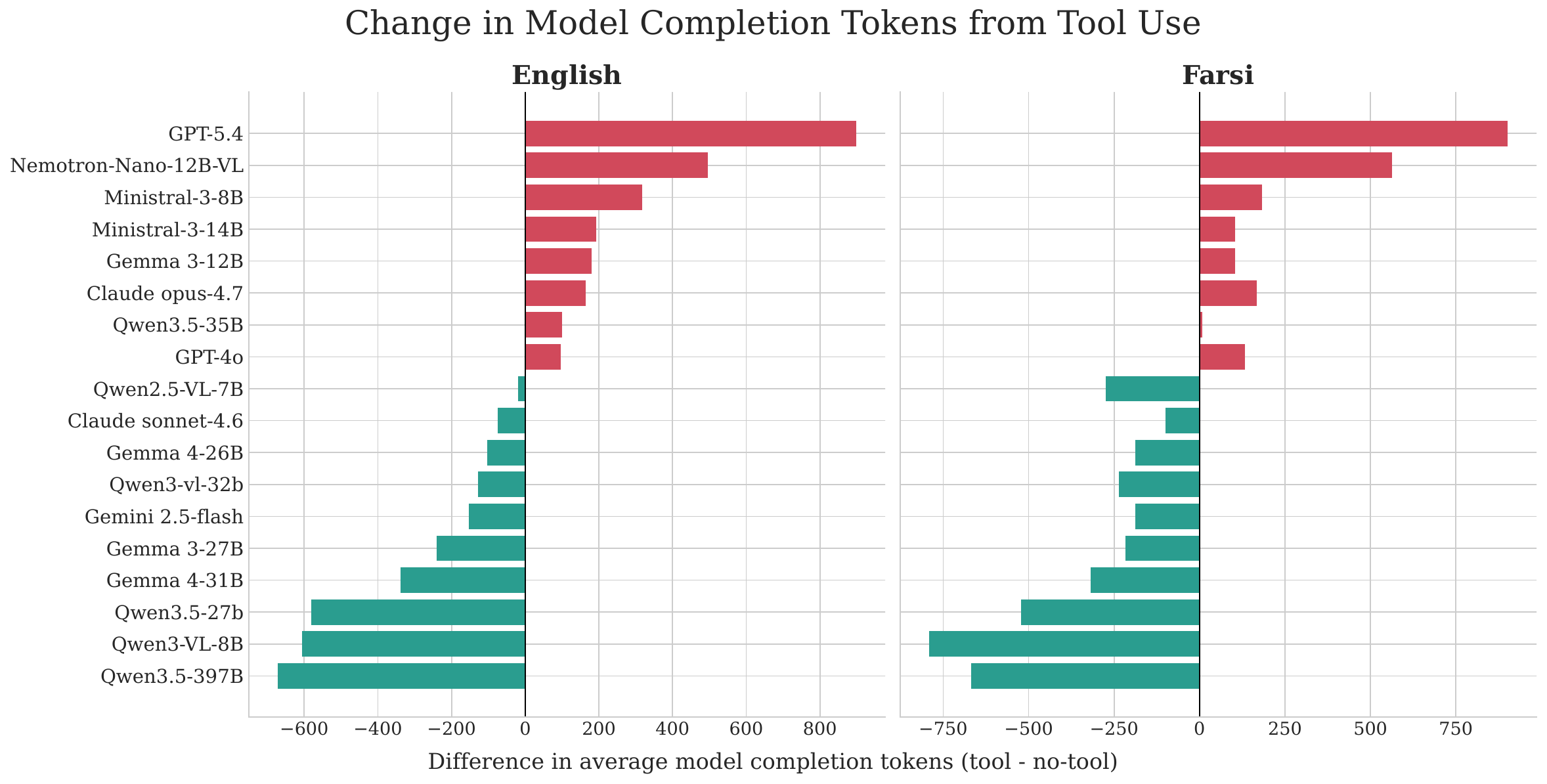}
    \caption{Change in average model completion tokens under tool use relative to {analytical solving} on the original English and Persian questions. Each bar shows the difference in average completion tokens between \textbf{R2: Tool-enabled Visual Solving} and \textbf{R1: Analytical Solving} for a given model. \textcolor{red}{Red} bars indicate increased token usage under tool use, while \textcolor{teal}{green} bars indicate reduced token usage. The increase for most models is expected: tool-enabled responses must emit tool call syntax, Desmos expressions, and screenshot inspection steps on top of the final answer. The reduction seen in stronger models such as Qwen3.5-397B and Gemma 4-31B is more telling: \textbf{sometimes a decisive plot can short-circuit lengthy symbolic derivations, making visual tool use not only accurate but more token-efficient than purely analytical solving.}}
    \label{fig:ttol_no_tool_token}
\end{figure*}

Figure~\ref{fig:token_acc} shows the performance--cost trade-off under {R2: Tool-enabled visual solving} on the original questions, comparing filtered accuracy against average completion-token usage across models. 
Figure~\ref{fig:token_correct_incorrect} compares completion-token usage for correct and incorrect answers under the R2 scenario on the original questions, showing that incorrect responses generally involve higher token generation.
Figure~\ref{fig:screenshot_dist} reports the distribution of screenshot counts generated under R2 on the original questions across different models, where zero screenshots indicate failures to properly invoke the tool. 
Additionally, Figure~\ref{fig:ttol_no_tool_token} shows the change in average completion tokens under tool use, R2, relative to analytical solving, R1, on the original questions, highlighting how the token cost of tool use varies across models and languages.
Finally, Table~\ref{tab:synthetic-results} reports accuracy results on the synthetic English and Persian questions under the R1 and R2 scenarios, along with judge-filtered accuracy. The synthetic-question results follow a broadly similar pattern to the original-question results reported in Table~\ref{tab:main-results}. Appendix Tables~\ref{tab:main-results-soft} and \ref{tab:synthetic-results-soft} additionally report softer-evaluation variants that relax the strict final-label extractor when the answer is still recoverable from the full generated response.

\begin{figure*}[!t]
    \centering
    \includegraphics[width=1.0\linewidth]{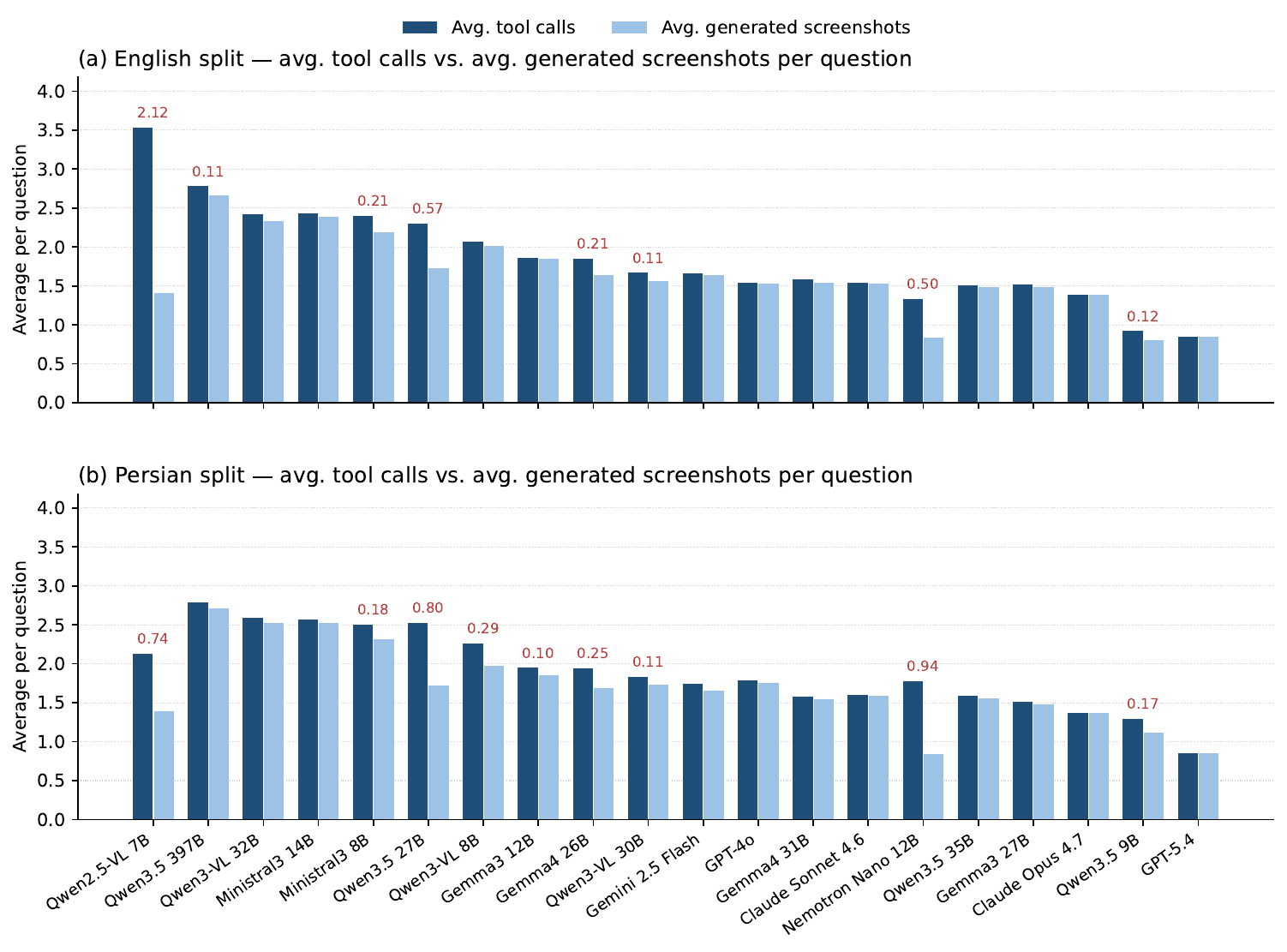}
    \caption{Tool-call reliability under \textbf{R2: Tool-enabled Visual Solving} on the original English (top) and Persian (bottom) questions. For each model we report the average number of tool calls issued per question (\textcolor{black!75}{dark blue}) alongside the average number of screenshots actually returned (\textcolor{black!40}{light blue}). The vertical gap between the two bars is the average number of \emph{failed} tool calls per question, attempts in which the model invoked the Desmos tool but no image was rendered, due to a malformed argument list, a tool-side error, or an empty plot. \textcolor{red!80!black!20}{Red} annotations show the size of the gap whenever it exceeds $0.10$ calls per question. Models are ordered (left to right) by the EN+FA total of \texttt{avg\_tool\_calls}, so heavy tool users appear first. Two patterns stand out: (i) the frontier closed-source models (Claude Opus 4.7, Claude Sonnet 4.6, GPT-5.4, GPT-4o) and several mid-size open models (Gemma\,4 31B, Qwen3.5 35B, Gemma\,3 27B) achieve near-zero gap on both languages, almost every tool call succeeds; (ii) a small set of open models leak a substantial fraction of their tool calls, most notably Qwen2.5-VL 7B on English ($+2.12$ failed calls per question, i.e.\ roughly $60\%$ of its attempts produce no screenshot), and Nemotron Nano 12B v2 VL on Persian ($+0.94$, roughly $53\%$). Tool-call reliability is therefore not a simple function of how many calls a model issues: Qwen3.5 397B issues $\sim 2.78$ calls per question yet leaks under $0.11$ on either language, while Nemotron Nano issues only $\sim 1.34$--$1.78$ calls but loses about half of them.}
    \label{fig:tool_call_reliability}
\end{figure*}

\begin{table*}[!t]
    \centering
    \scriptsize
    \setlength{\tabcolsep}{3.2pt}
    \renewcommand{\arraystretch}{1.03}
    \caption{Synthetic-subset results mirroring Table~\ref{tab:main-results}. Values are accuracies in percent; ``Judge'' denotes the filtered accuracy under the VLM-as-a-judge protocol.}
    \label{tab:synthetic-results}
    \begin{tabular}{@{}lcccccc@{}}
        \toprule
        & \multicolumn{3}{c}{English} & \multicolumn{3}{c}{Persian} \\
        \cmidrule(lr){2-4} \cmidrule(lr){5-7}
        Model & Analyt. & Tool & Judge & Analyt. & Tool & Judge \\
        \midrule
        \multicolumn{7}{@{}l}{\textbf{Small}} \\
        Gemma3 12B & 61.5 & 39.9 & 39.9 & 60.7 & 39.9 & 39.9 \\
        Ministral3 14B & 69.1 & 61.2 & 61.2 & 66.9 & 59.6 & 59.6 \\
        Ministral3 8B & 66.1 & 53.3 & 52.7 & 57.1 & 54.6 & 54.6 \\
        Nemotron Nano 12B 2 VL & 41.0 & 40.7 & 29.8 & 29.5 & 37.2 & 26.5 \\
        Qwen2.5-VL 7B & 46.2 & 33.6 & 29.5 & 38.5 & 33.6 & 29.8 \\
        Qwen3-VL 8B & 77.0 & 49.7 & 49.7 & 66.9 & 50.0 & 50.0 \\
        \addlinespace[2pt]
        \multicolumn{7}{@{}l}{\textbf{Medium}} \\
        Gemma3 27B & 64.2 & 50.3 & 50.0 & 63.9 & 51.9 & 51.6 \\
        Gemma4 26B-A4B & 89.6 & 81.2 & 80.6 & 92.3 & 81.7 & 81.2 \\
        Gemma4 31B & 70.0 & 84.7 & 84.7 & 86.3 & 85.5 & 85.5 \\
        Qwen3-VL 32B & 88.5 & 71.3 & 70.5 & 86.9 & 73.2 & 72.1 \\
        Qwen3.5 27B & 92.3 & 84.7 & 78.7 & 91.0 & 80.9 & 79.0 \\
        Qwen3.5 35B-A3B & 60.9 & 75.4 & 72.4 & 76.8 & 71.3 & 68.3 \\
        \addlinespace[2pt]
        \multicolumn{7}{@{}l}{\textbf{Frontier / Large}} \\
        Gemini 2.5 Flash & 86.1 & 73.8 & 73.8 & 86.1 & 77.6 & 77.3 \\
        Qwen3.5 397B-A17B & 91.0 & 85.0 & 85.0 & 92.1 & 88.2 & 88.2 \\
        \bottomrule
    \end{tabular}
\end{table*}

\begin{table*}[!t]
    \centering
    \scriptsize
    \setlength{\tabcolsep}{3.2pt}
    \renewcommand{\arraystretch}{1.03}
    \caption{Original Konkour-seed results under a softer final-label extraction baseline. Values are accuracies in percent; ``Judge'' denotes the filtered accuracy under the VLM-as-a-judge protocol. The softer extractor is used only when the final option is recoverable from the full model response despite strict-format violations.}
    \label{tab:main-results-soft}
    \begin{tabular}{@{}lcccccc@{}}
        \toprule
        & \multicolumn{3}{c}{English} & \multicolumn{3}{c}{Persian} \\
        \cmidrule(lr){2-4} \cmidrule(lr){5-7}
        Model & Analyt. & Tool & Judge & Analyt. & Tool & Judge \\
        \midrule
        \multicolumn{7}{@{}l}{\textbf{Small}} \\
        Gemma3 12B & 68.3 & 39.0 & 39.0 & 64.7 & 43.6 & 42.7 \\
        Ministral3 14B & 75.7 & 60.5 & 60.5 & 67.9 & 53.7 & 53.7 \\
        Ministral3 8B & 67.4 & 59.6 & 59.6 & 57.3 & 50.9 & 50.5 \\
        Nemotron Nano 12B 2 VL & 53.7 & 37.2 & 30.7 & 40.8 & 29.8 & 26.6 \\
        Qwen2.5-VL 7B & 55.0 & 36.7 & 34.4 & 48.2 & 32.6 & 30.7 \\
        Qwen3-VL 8B & 93.1 & 51.4 & 51.4 & 83.9 & 49.1 & 49.1 \\
        \addlinespace[2pt]
        \multicolumn{7}{@{}l}{\textbf{Medium}} \\
        Gemma3 27B & 77.5 & 50.5 & 50.5 & 74.8 & 46.8 & 46.8 \\
        Gemma4 26B-A4B & 96.3 & 85.8 & 85.3 & 95.4 & 81.7 & 81.2 \\
        Gemma4 31B & 97.2 & 89.5 & 89.5 & 97.2 & 84.4 & 84.4 \\
        Qwen3-VL 32B & 92.2 & 74.8 & 74.8 & 84.4 & 71.6 & 71.6 \\
        Qwen3.5 27B & 97.2 & 83.5 & 83.5 & 96.8 & 83.9 & 83.9 \\
        Qwen3.5 35B-A3B & 95.4 & 77.1 & 76.6 & 94.0 & 77.1 & 76.2 \\
        \addlinespace[2pt]
        \multicolumn{7}{@{}l}{\textbf{Frontier / Large}} \\
        Claude Opus 4.7 & 98.2 & 93.6 & 93.6 & 98.2 & 92.7 & 92.7 \\
        Claude Sonnet 4.6 & 98.2 & 88.5 & 88.5 & 95.9 & 88.5 & 88.5 \\
        GPT-4o & 49.1 & 49.5 & 49.5 & 45.9 & 49.5 & 49.5 \\
        GPT-5.4 & 89.0 & 66.5 & 55.0 & 89.5 & 70.6 & 56.4 \\
        Gemini 2.5 Flash & 90.8 & 78.9 & 78.9 & 83.9 & 84.4 & 84.4 \\
        Qwen3.5 397B-17B & 98.2 & 88.5 & 88.5 & 97.7 & 88.5 & 88.5 \\
        \bottomrule
    \end{tabular}
\end{table*}

\begin{table*}[!t]
    \centering
    \scriptsize
    \setlength{\tabcolsep}{3.2pt}
    \renewcommand{\arraystretch}{1.03}
    \caption{Synthetic-subset results under the same softer final-label extraction baseline used in Table~\ref{tab:main-results-soft}. Values are accuracies in percent; ``Judge'' denotes the filtered accuracy under the VLM-as-a-judge protocol.}
    \label{tab:synthetic-results-soft}
    \begin{tabular}{@{}lcccccc@{}}
        \toprule
        & \multicolumn{3}{c}{English} & \multicolumn{3}{c}{Persian} \\
        \cmidrule(lr){2-4} \cmidrule(lr){5-7}
        Model & Analyt. & Tool & Judge & Analyt. & Tool & Judge \\
        \midrule
        \multicolumn{7}{@{}l}{\textbf{Small}} \\
        Gemma3 12B & 62.0 & 40.2 & 40.2 & 61.5 & 40.2 & 39.9 \\
        Ministral3 14B & 69.1 & 61.8 & 61.8 & 67.5 & 59.8 & 59.8 \\
        Ministral3 8B & 66.9 & 53.5 & 53.0 & 57.4 & 54.9 & 54.9 \\
        Nemotron Nano 12B 2 VL & 45.9 & 40.7 & 29.8 & 30.6 & 37.2 & 26.5 \\
        Qwen2.5-VL 7B & 46.2 & 33.6 & 29.5 & 38.5 & 33.6 & 29.8 \\
        Qwen3-VL 8B & 77.6 & 50.0 & 50.0 & 66.9 & 50.8 & 50.8 \\
        \addlinespace[2pt]
        \multicolumn{7}{@{}l}{\textbf{Medium}} \\
        Gemma3 27B & 64.2 & 50.8 & 50.5 & 64.2 & 51.9 & 51.6 \\
        Gemma4 26B & 91.0 & 82.0 & 81.4 & 93.4 & 83.3 & 82.8 \\
        Gemma4 31B & 71.3 & 85.0 & 85.0 & 89.6 & 85.5 & 85.5 \\
        Qwen3-VL 32B & 89.3 & 71.3 & 70.5 & 86.9 & 73.8 & 72.7 \\
        Qwen3.5 27B & 93.4 & 86.1 & 79.8 & 92.6 & 82.0 & 80.1 \\
        Qwen3.5 35B & 91.0 & 76.5 & 73.5 & 87.2 & 71.9 & 68.6 \\
        \addlinespace[2pt]
        \multicolumn{7}{@{}l}{\textbf{Frontier / Large}} \\
        Gemini 2.5 Flash & 86.1 & 74.0 & 74.0 & 86.3 & 78.1 & 77.9 \\
        Qwen3.5 397B & 92.9 & 85.2 & 85.2 & 92.9 & 88.2 & 88.2 \\
        \bottomrule
    \end{tabular}
\end{table*}


\newpage
\section{Qualitative Case Studies}
\label{app:cases}

\begin{figure}[!t]
    \centering
    \begin{tcolorbox}[
        enhanced, colback=blue!4, colframe=blue!60!black,
        boxrule=0.4pt, boxsep=3pt, left=6pt, right=6pt, top=3pt, bottom=3pt,
        title={\bfseries Question 123 (English)},
        fonttitle=\footnotesize, coltitle=white,
        attach boxed title to top left={yshift=-1mm, xshift=4mm},
        boxed title style={colback=blue!60!black, sharp corners},
        sharp corners=south, width=\linewidth
    ]
    \footnotesize
    The solution set of the inequality $-1<\dfrac{3x+1}{x-3}<3$ is which of the following?\\[2pt]
    \textbf{Options:} (1)~$x<\tfrac{1}{2}$\quad (2)~$x<3$\quad (3)~$-\tfrac{1}{2}<x<3$\quad (4)~$\tfrac{1}{2}<x<3$\\[1pt]
    \textbf{Ground truth:} option~(1), $x<\tfrac{1}{2}$.
    \end{tcolorbox}

    \begin{minipage}[t]{0.46\linewidth}
        \begin{tcolorbox}[
            enhanced, colback=white, colframe=red!55!black,
            boxrule=0.4pt, boxsep=2pt, left=2pt, right=2pt, top=2pt, bottom=2pt,
            title={\bfseries Tool-enabled (R2): model-generated plot},
            fonttitle=\footnotesize, coltitle=white,
            attach boxed title to top left={yshift=-1mm, xshift=4mm},
            boxed title style={colback=red!55!black, sharp corners},
            sharp corners=south, valign=top, height=5.4cm
        ]
        \centering
        \includegraphics[width=\linewidth, height=4.4cm, keepaspectratio]{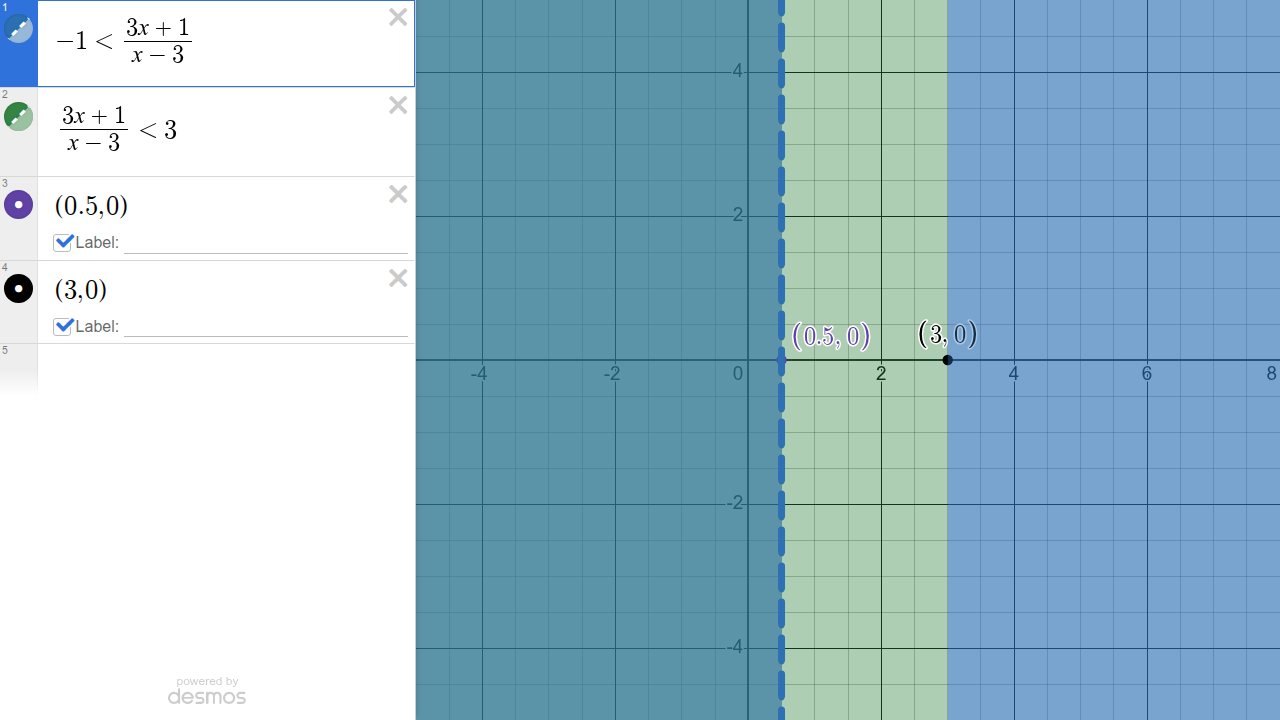}
        \end{tcolorbox}
    \end{minipage}\hfill
    \begin{minipage}[t]{0.51\linewidth}
        \begin{tcolorbox}[
            enhanced, colback=red!3, colframe=red!55!black,
            boxrule=0.4pt, boxsep=3pt, left=6pt, right=6pt, top=3pt, bottom=3pt,
            title={\bfseries Model reasoning and final answer},
            fonttitle=\footnotesize, coltitle=white,
            attach boxed title to top left={yshift=-1mm, xshift=4mm},
            boxed title style={colback=red!55!black, sharp corners},
            sharp corners=south, valign=top, height=5.4cm
        ]
        \footnotesize
        \textit{steps\_summary:} ``Plotted the two inequalities $-1<\!\!\tfrac{3x+1}{x-3}$ and $\tfrac{3x+1}{x-3}\!\!<3$ in Desmos. The overlapping (green) region appears as a vertical strip between $x=0.5$ and $x=3$, matching the labeled probe points $(0.5,0)$ and $(3,0)$. This corresponds to $\tfrac{1}{2}<x<3$.''\\[4pt]
        \textbf{selected\_option:} \texttt{"4"} \quad{\color{red!70!black}$\times$}\\
        \textbf{Failure mode:} the model commits to the visible green strip on the right of the asymptote and never inspects the left branch ($x<3$, $x<\tfrac{1}{2}$), where the inequality is also satisfied. This is an instance of \emph{endpoint-direction inversion} (\S\ref{sec:analysis}, FM3). 
        \end{tcolorbox}
    \end{minipage}

    \caption{Case A: tool-enabled-failure. Claude Opus~4.7, the strongest model in our experiments, selects the wrong option once required to ground its answer in a self-generated Desmos screenshot. The plot is informative but is read on only one side of the vertical asymptote $x=3$.}
    \label{fig:appendix-case-a}
\end{figure}

\begin{figure}[!t]
    \centering
    \begin{tcolorbox}[
        enhanced, colback=blue!4, colframe=blue!60!black,
        boxrule=0.4pt, boxsep=3pt, left=6pt, right=6pt, top=3pt, bottom=3pt,
        title={\bfseries Question 26 (English)},
        fonttitle=\footnotesize, coltitle=white,
        attach boxed title to top left={yshift=-1mm, xshift=4mm},
        boxed title style={colback=blue!60!black, sharp corners},
        sharp corners=south, width=\linewidth
    ]
    \footnotesize
    How many distinct real roots does the equation $(x^{2}-2x)^{2}-(x^{2}-2x)=2$ have?\\[2pt]
    \textbf{Options:} (1)~1\quad (2)~2\quad (3)~3\quad (4)~4 \qquad\textbf{Ground truth:} option~(3), three roots.
    \end{tcolorbox}

    \begin{minipage}[t]{0.49\linewidth}
        \begin{tcolorbox}[
            enhanced, colback=red!3, colframe=red!55!black,
            boxrule=0.4pt, boxsep=2pt, left=3pt, right=3pt, top=2pt, bottom=2pt,
            title={\bfseries R2 (tool-enabled): self-generated plot},
            fonttitle=\footnotesize, coltitle=white,
            attach boxed title to top left={yshift=-1mm, xshift=4mm},
            boxed title style={colback=red!55!black, sharp corners},
            sharp corners=south, valign=top
        ]
        \centering
        \includegraphics[width=\linewidth, height=3.6cm, keepaspectratio]{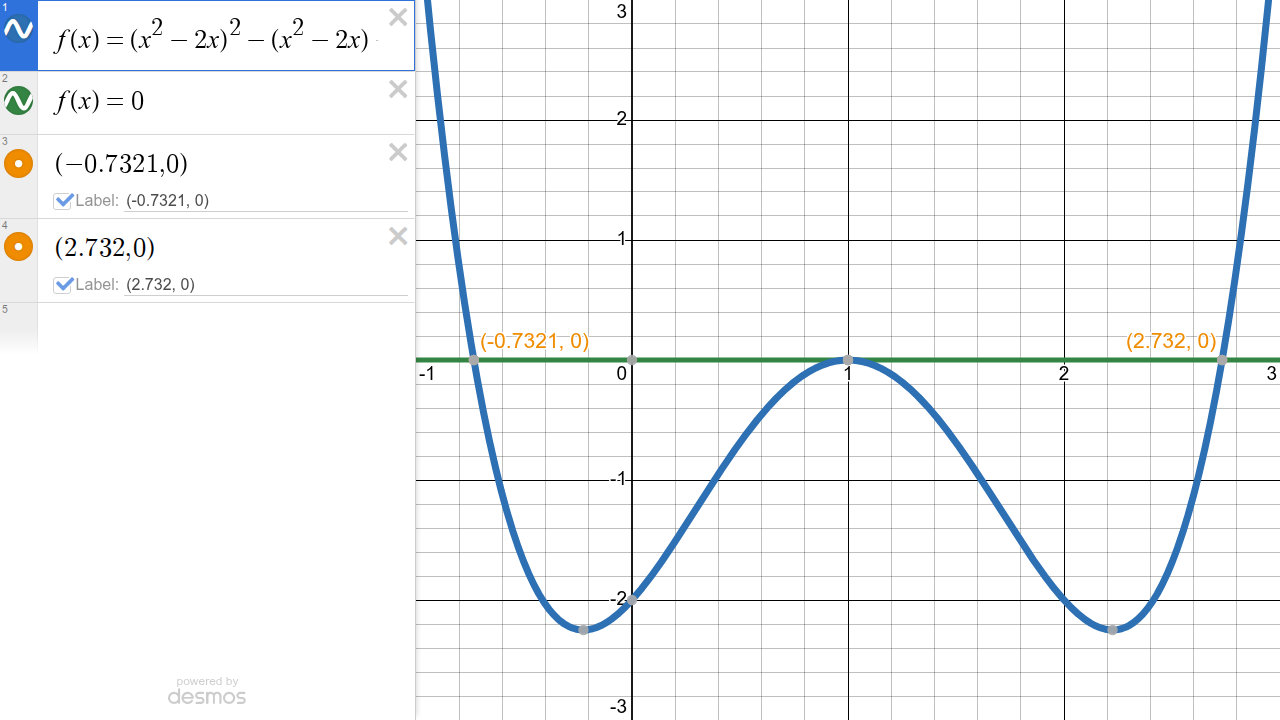}\\[2pt]
        \begin{flushleft}
        \footnotesize
        \textit{steps\_summary:} ``The Desmos screenshots show exactly 2 zero crossings labeled at $(-0.7321,0)$ and $(2.732,0)$.''\\[2pt]
        \textbf{selected\_option:} \texttt{"2"} \quad{\color{red!70!black}$\times$}\\[1pt]
        \textbf{Why it fails:} Desmos's automatic \texttt{label\_zeros} marks two crossings and silently omits the tangent root at $x=1$. The model trusts the label count.
        \end{flushleft}
        \end{tcolorbox}
    \end{minipage}\hfill
    \begin{minipage}[t]{0.49\linewidth}
        \begin{tcolorbox}[
            enhanced, colback=green!3, colframe=green!45!black,
            boxrule=0.4pt, boxsep=2pt, left=3pt, right=3pt, top=2pt, bottom=2pt,
            title={\bfseries R3 (visual-only): curated visualization},
            fonttitle=\footnotesize, coltitle=white,
            attach boxed title to top left={yshift=-1mm, xshift=4mm},
            boxed title style={colback=green!45!black, sharp corners},
            sharp corners=south, valign=top
        ]
        \centering
        \includegraphics[width=\linewidth, height=3.6cm, keepaspectratio]{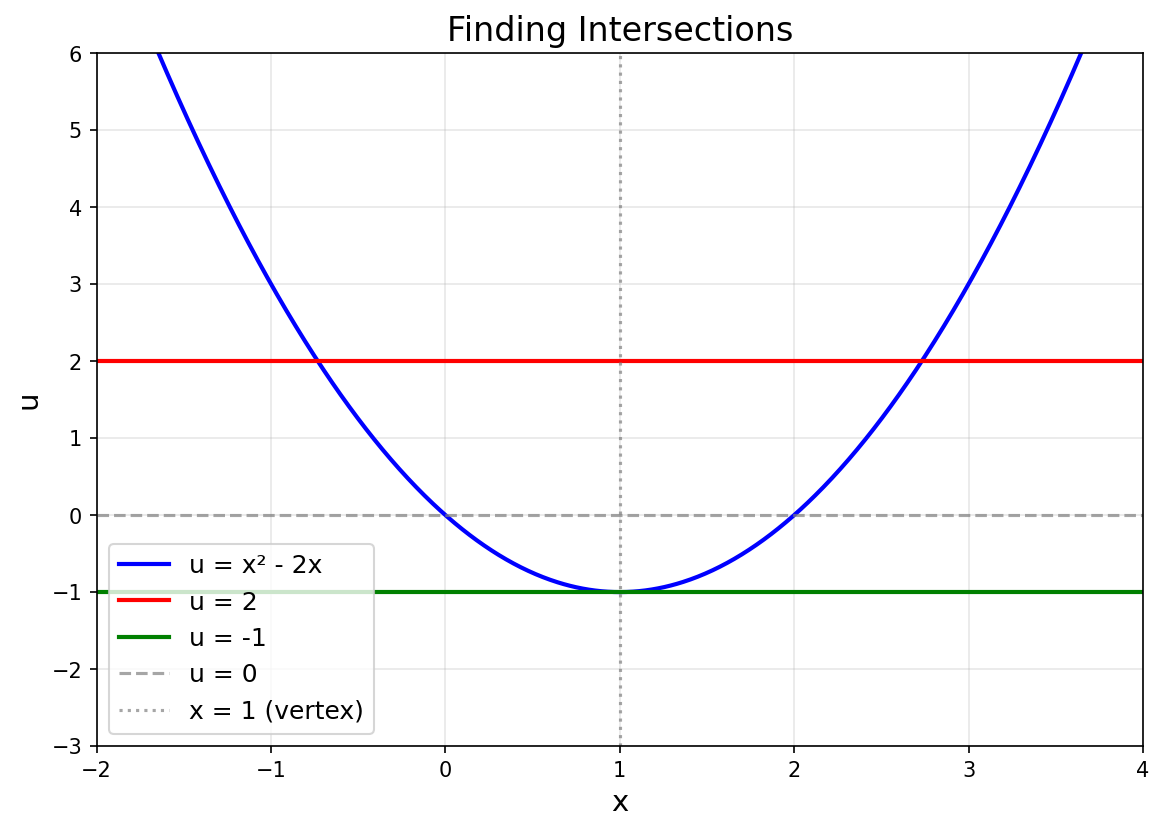}\\[2pt]
        \begin{flushleft}
        \footnotesize
        \textit{steps\_summary:} ``The curve crosses $u=2$ at two points and touches $u=-1$ tangentially at $x=1$. Total $=2+1=3$ distinct real roots.''\\[2pt]
        \textbf{selected\_option:} \texttt{"3"} \quad{\color{green!50!black}\checkmark}\\[1pt]
        \textbf{Why it succeeds:} the substitution $u=x^{2}-2x$ makes the tangent contact at $x=1$ explicit, so the model no longer depends on Desmos's POI labels.
        \end{flushleft}
        \end{tcolorbox}
    \end{minipage}

    \caption{Case B: visual-only-success / tool-enabled-failure. \emph{Same model} (Claude Sonnet~4.6), \emph{same question}. In R2, Sonnet's self-generated Desmos plot returns two auto-labeled zeros and the tangent root at $x=1$ is silently omitted; Sonnet trusts the count and answers ``2''. In R3, the curated layered visualization reframes the problem in $u=x^{2}-2x$ and exposes the tangent contact at $u=-1$; Sonnet now reads the screenshot correctly and answers ``3''.}
    \label{fig:appendix-case-b}
\end{figure}
\newpage

\section{Catalog of Questions}
\label{app:catalog}

\begin{figure}[p]
    \centering
\begin{minipage}[t]{0.485\linewidth}
\begin{tcolorbox}[
    enhanced, colback=white, colframe=purple!85!black,
    boxrule=0.4pt, boxsep=2pt, left=4pt, right=4pt, top=2pt, bottom=2pt,
    title={\bfseries\scriptsize Q52 \,\textbar\, EN \quad \itshape\textnormal{FM1 \textbullet{} Composition-plot syntax failure}},
    fonttitle=\scriptsize, coltitle=white,
    boxed title style={colback=purple!60!black, sharp corners},
    sharp corners=south
]
\scriptsize
Let $f(x)=\log(2x-5)$ and $g(x)=x+\sqrt{2x-4}$. If $y=g^{-1}(f^{-1}(x))$, then at which point does the graph intersect the $y$-axis?\\[2pt]
\textbf{Options:}\\[-2pt]
{\setlength{\tabcolsep}{2pt}%
\begin{tabular}[t]{@{}p{0.04\linewidth}p{0.40\linewidth}p{0.04\linewidth}p{0.40\linewidth}@{}}
(1) & $4-\sqrt{2}$ & (2) & $4-\sqrt{3}$ \\
(3) & $4+\sqrt{2}$ & (4) & $4+\sqrt{3}$ \\
\end{tabular}}\\[2pt]
\centering
\fbox{\parbox[c][2.3cm][c]{0.92\linewidth}{\centering\itshape\scriptsize no usable plot generated}}

\begin{flushleft}
\textit{Qwen3.5}; ``The model failed to emit a valid desmos\_plot tool call after multiple attempts; final answer N/A. \textbf{Reasoning trace shows the model attempting an analytical solution instead.}''\\[1pt]
\textbf{pred:}~N/A\quad\textbf{gt:}~2\quad{\color{red!70!black}$\times$}
\end{flushleft}
\end{tcolorbox}
\end{minipage}
\hfill
\begin{minipage}[t]{0.485\linewidth}
\begin{tcolorbox}[
    enhanced, colback=white, colframe=orange!75!black,
    boxrule=0.4pt, boxsep=2pt, left=4pt, right=4pt, top=2pt, bottom=2pt,
    title={\bfseries\scriptsize Q22 \,\textbar\, EN \quad \itshape\textnormal{FM2 \textbullet{} Coincident-label miscount}},
    fonttitle=\scriptsize, coltitle=white,
    boxed title style={colback=orange!75!black, sharp corners},
    sharp corners=south
]
\scriptsize
How many points of intersection does the graph of \(f(x)=\sqrt{\,1-\sqrt{1+x}\,}\) have with its inverse function?\\[2pt]
\textbf{Options:}\\[-2pt]
{\setlength{\tabcolsep}{2pt}%
\begin{tabular}[t]{@{}p{0.04\linewidth}p{0.40\linewidth}p{0.04\linewidth}p{0.40\linewidth}@{}}
(1) & 4 & (2) & 3 \\
(3) & 2 & (4) & 1 \\
\end{tabular}}\\[2pt]
\centering
\includegraphics[width=\linewidth, height=2.3cm, keepaspectratio]{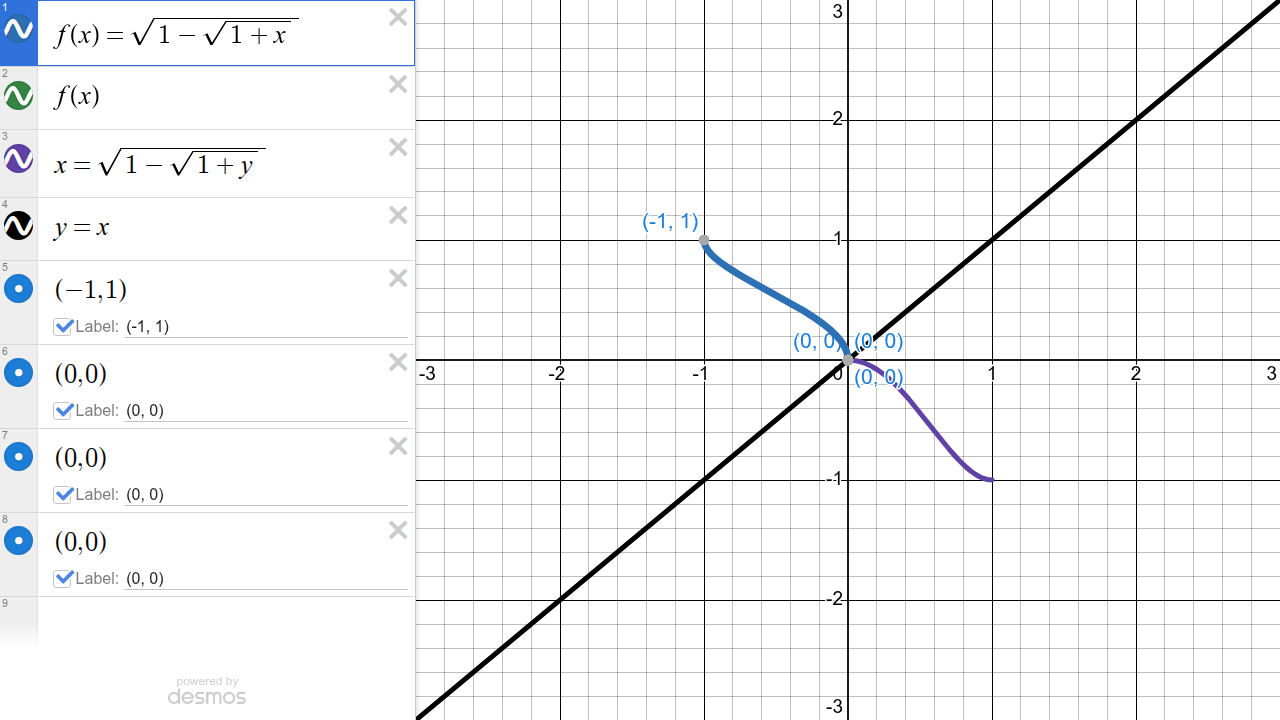}

\begin{flushleft}
\textit{Claude Sonnet 4.6}: ``Let me check: -1 $\neq$ 1, so (-1,1) is an intersection of f with its inverse but NOT on y=x.''\\[1pt]
\textbf{pred:}~3\quad\textbf{gt:}~4\quad{\color{red!70!black}$\times$}
\end{flushleft}
\end{tcolorbox}
\end{minipage}
\\[6pt]
\begin{minipage}[t]{0.485\linewidth}
\begin{tcolorbox}[
    enhanced, colback=white, colframe=orange!75!black,
    boxrule=0.4pt, boxsep=2pt, left=4pt, right=4pt, top=2pt, bottom=2pt,
    title={\bfseries\scriptsize Q45 \,\textbar\, EN \quad \itshape\textnormal{FM2 \textbullet{} Auto-label over-trust (wrong-curve label)}},
    fonttitle=\scriptsize, coltitle=white,
    boxed title style={colback=orange!75!black, sharp corners},
    sharp corners=south
]
\scriptsize
At the point of intersection of the graphs $f(x)=\sin x+\frac{1}{2}\cos x$ and $g(x)=\frac{3}{2}\sin x$ on the interval $[0,\pi]$, a tangent line to the graph of \$f…\\[2pt]
\textbf{Options:}\\[-2pt]
{\setlength{\tabcolsep}{2pt}%
\begin{tabular}[t]{@{}p{0.04\linewidth}p{0.40\linewidth}p{0.04\linewidth}p{0.40\linewidth}@{}}
(1) & $\frac{\pi}{4}-1$ & (2) & $\frac{\pi}{4}-3$ \\
(3) & $\frac{\pi}{4}+\frac{1}{8}$ & (4) & $\frac{\pi}{4}+\frac{3}{8}$ \\
\end{tabular}}\\[2pt]
\centering
\includegraphics[width=\linewidth, height=2.3cm, keepaspectratio]{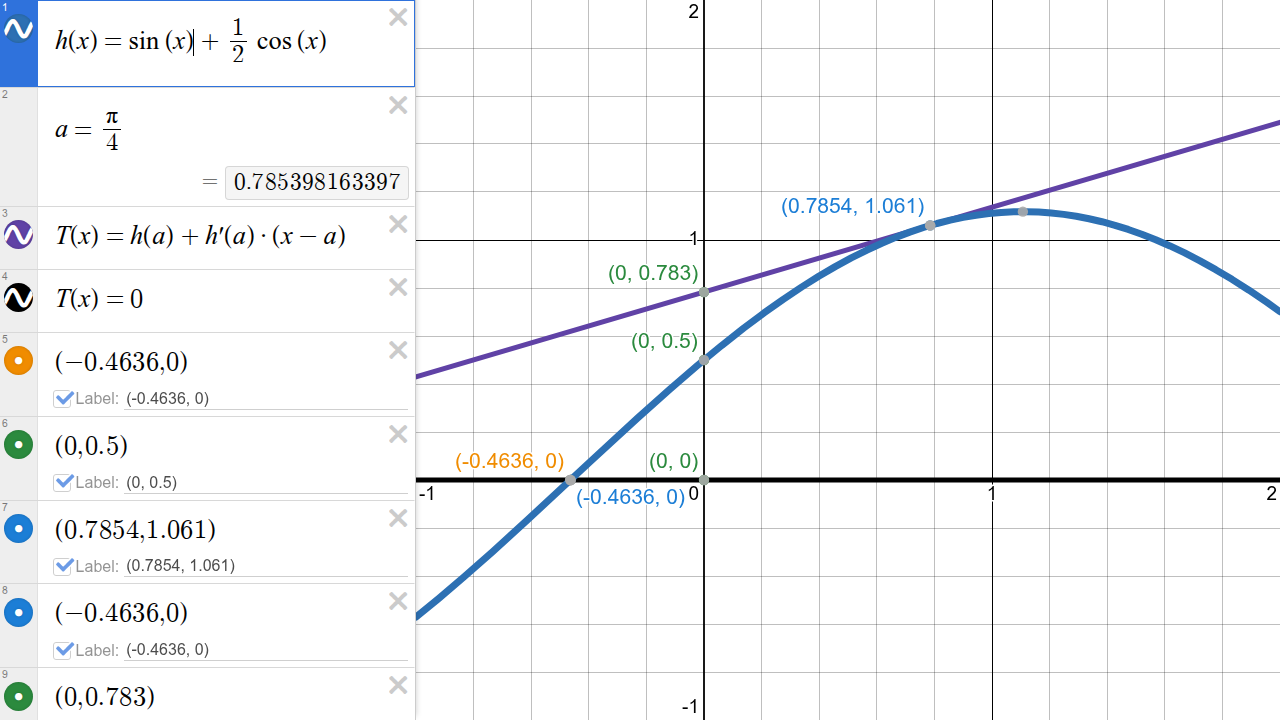}

\begin{flushleft}
\textit{Claude Sonnet 4.6}: ``Now I need to check which option this matches. $\pi$/4 $\approx$ 0.7854 Option 1: $\pi$/4 - 1 $\approx$ 0.7854 - 1 = -0.2146 Option 2: $\pi$/4 - 3 $\approx$ 0.7854 - 3 =…''\\[1pt]
\textbf{pred:}~1\quad\textbf{gt:}~2\quad{\color{red!70!black}$\times$}
\end{flushleft}
\end{tcolorbox}
\end{minipage}
\hfill
\begin{minipage}[t]{0.485\linewidth}
\begin{tcolorbox}[
    enhanced, colback=white, colframe=orange!75!black,
    boxrule=0.4pt, boxsep=2pt, left=4pt, right=4pt, top=2pt, bottom=2pt,
    title={\bfseries\scriptsize Q49 \,\textbar\, FA \quad \itshape\textnormal{FM2 \textbullet{} Auto-label over-trust (spurious intersection)}},
    fonttitle=\scriptsize, coltitle=white,
    boxed title style={colback=orange!75!black, sharp corners},
    sharp corners=south
]
\scriptsize
At the intersection point of the graphs $f(x)=-\sqrt[3]{x}$ and $g(x)=\sqrt[3]{x^2}-2$ in the second quadrant of the coordinate plane, a tangent line to $g(x)$ is d…\\[2pt]
\textbf{Options:}\\[-2pt]
{\setlength{\tabcolsep}{2pt}%
\begin{tabular}[t]{@{}p{0.04\linewidth}p{0.40\linewidth}p{0.04\linewidth}p{0.40\linewidth}@{}}
(1) & $\frac{5}{3}$ & (2) & $\frac{4}{3}$ \\
(3) & $\frac{2}{3}$ & (4) & $\frac{1}{3}$ \\
\end{tabular}}\\[2pt]
\centering
\includegraphics[width=\linewidth, height=2.3cm, keepaspectratio]{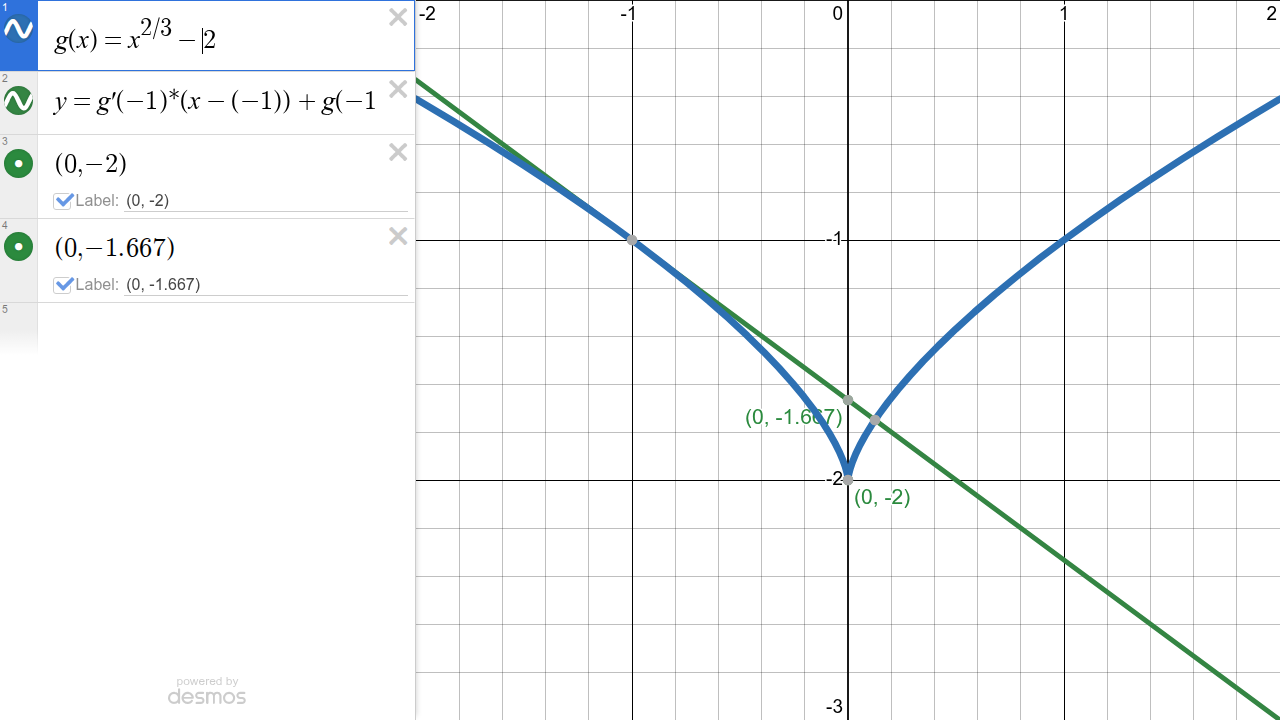}

\begin{flushleft}
\textit{Claude Sonnet 4.6}: ``The screenshot clearly shows the tangent line to g(x) at x=-1 intersects the y-axis at (0, -1.667), which is approximately -5/3.''\\[1pt]
\textbf{pred:}~1\quad\textbf{gt:}~3\quad{\color{red!70!black}$\times$}
\end{flushleft}
\end{tcolorbox}
\end{minipage}
\\[6pt]
\begin{minipage}[t]{0.485\linewidth}
\begin{tcolorbox}[
    enhanced, colback=white, colframe=orange!75!black,
    boxrule=0.4pt, boxsep=2pt, left=4pt, right=4pt, top=2pt, bottom=2pt,
    title={\bfseries\scriptsize Q217 \,\textbar\, FA \quad \itshape\textnormal{FM2 \textbullet{} Auto-label over-trust (boundary singularity)}},
    fonttitle=\scriptsize, coltitle=white,
    boxed title style={colback=orange!75!black, sharp corners},
    sharp corners=south
]
\scriptsize
How many positive roots does the equation $\frac{1}{\sqrt{2-x}+2}-\frac{1}{2-\sqrt{2-x}}=\frac{2-x}{5\sqrt{2-x}}$ have?\\[2pt]
\textbf{Options:}\\[-2pt]
{\setlength{\tabcolsep}{2pt}%
\begin{tabular}[t]{@{}p{0.04\linewidth}p{0.40\linewidth}p{0.04\linewidth}p{0.40\linewidth}@{}}
(1) & 0 & (2) & 1 \\
(3) & 2 & (4) & 3 \\
\end{tabular}}\\[2pt]
\centering
\includegraphics[width=\linewidth, height=2.3cm, keepaspectratio]{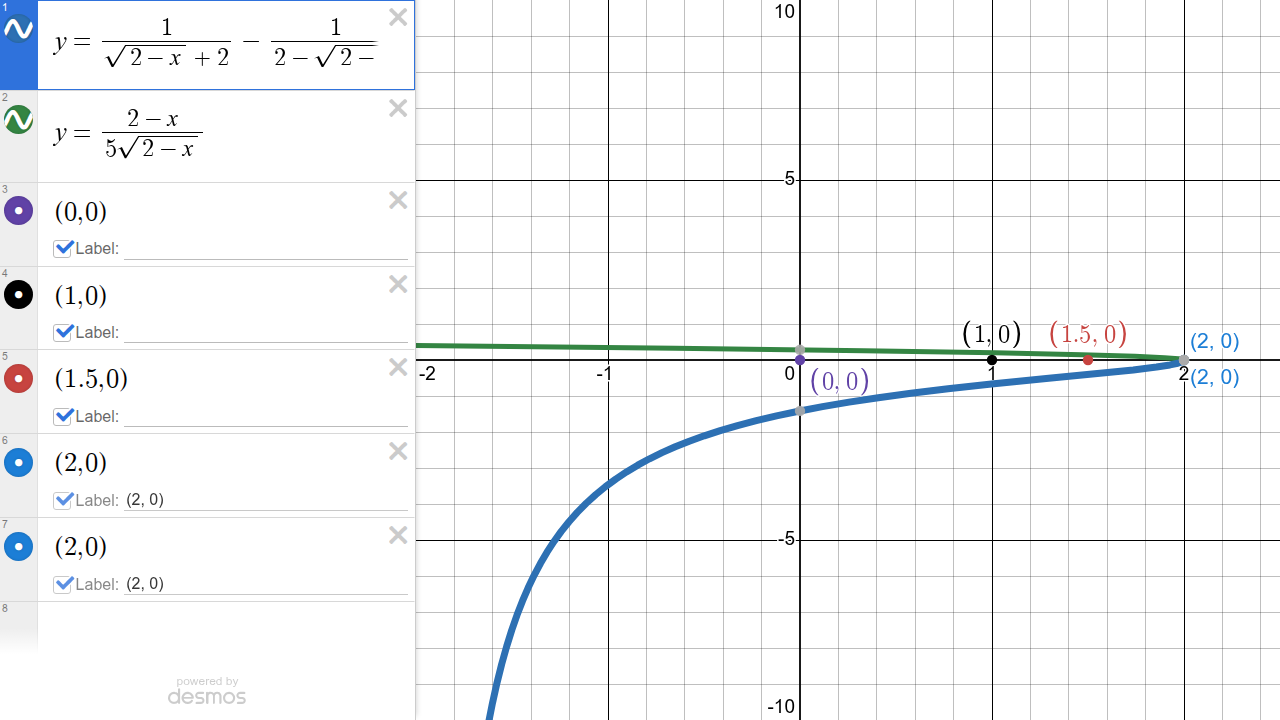}

\begin{flushleft}
\textit{GPT-5.4}: ``From the Desmos screenshot, the green and blue graphs are the two sides of the equation. The only visible intersection marker between the two graphs is at (2,0). For positive…''\\[1pt]
\textbf{pred:}~2\quad\textbf{gt:}~1\quad{\color{red!70!black}$\times$}
\end{flushleft}
\end{tcolorbox}
\end{minipage}
\hfill
\begin{minipage}[t]{0.485\linewidth}
\begin{tcolorbox}[
    enhanced, colback=white, colframe=orange!75!black,
    boxrule=0.4pt, boxsep=2pt, left=4pt, right=4pt, top=2pt, bottom=2pt,
    title={\bfseries\scriptsize Q10 \,\textbar\, EN \quad \itshape\textnormal{FM2 \textbullet{} Floor-function zoom-precision failure}},
    fonttitle=\scriptsize, coltitle=white,
    boxed title style={colback=orange!75!black, sharp corners},
    sharp corners=south
]
\scriptsize
What is the limit of the expression \(\lfloor \sin\left(x-\frac{\pi}{3}\right)\rfloor \cos 3x + \lfloor \tan^{2}x \rfloor\) as \(x \to \frac{\pi}{3}\)? (Here \textbackslash\{\}(\textbackslash\{\}lfl…\\[2pt]
\textbf{Options:}\\[-2pt]
{\fontsize{6}{7.2}\selectfont\setlength{\tabcolsep}{2pt}%
\begin{tabular}[t]{@{}p{0.04\linewidth}p{0.40\linewidth}p{0.04\linewidth}p{0.40\linewidth}@{}}
(1) & 1 & (2) & 2 \\
(3) & 3 & (4) & The limit does not exist. \\
\end{tabular}}\\[2pt]
\centering
\includegraphics[width=\linewidth, height=2.3cm, keepaspectratio]{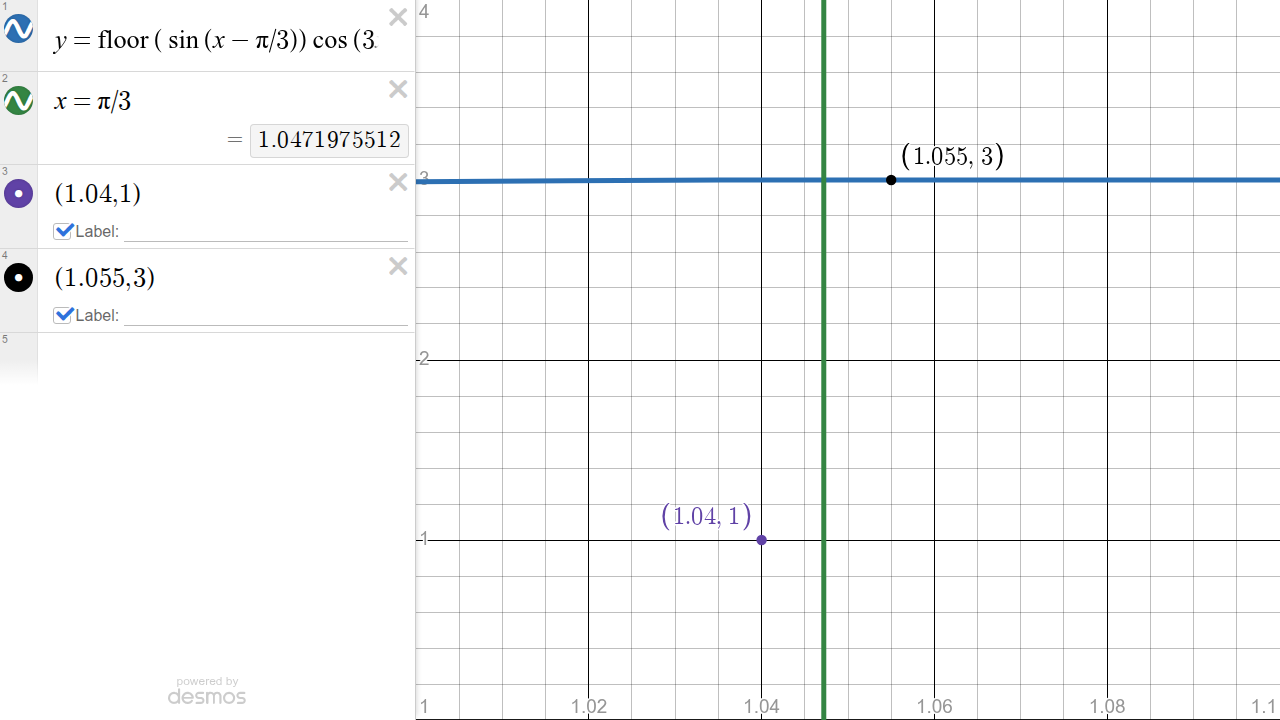}

\begin{flushleft}
\textit{GPT-5.4}: ``The Desmos screenshot shows the vertical line x=$\pi$/3 at about x=1.0472. A labeled point just left of that line is (1.04, 1), while a labeled point just right of that line…''\\[1pt]
\textbf{pred:}~4\quad\textbf{gt:}~3\quad{\color{red!70!black}$\times$}
\end{flushleft}
\end{tcolorbox}
\end{minipage}
    \caption{Catalog of hard questions, page 1 of 4: FM1 (composition-plot syntax) and FM2 (auto-label over-trust + floor-zoom). Each panel shows the question stub, a representative wrong model's self-generated Desmos screenshot, a brief verbatim excerpt from that model's final response, and the predicted vs.\ ground-truth option (\(\times\) marks the wrong selection). Panel header colour indicates the dominant failure mode (FM1\,=\,purple, FM2\,=\,orange).}
    \label{fig:catalog-1}
\end{figure}

\begin{figure}[p]
    \centering
\begin{minipage}[t]{0.485\linewidth}
\begin{tcolorbox}[
    enhanced, colback=white, colframe=red!55!black,
    boxrule=0.4pt, boxsep=2pt, left=4pt, right=4pt, top=2pt, bottom=2pt,
    title={\bfseries\scriptsize Q105 \,\textbar\, EN+FA \quad \itshape\textnormal{FM3 \textbullet{} Endpoint-direction inversion}},
    fonttitle=\scriptsize, coltitle=white,
    boxed title style={colback=red!55!black, sharp corners},
    sharp corners=south
]
\scriptsize
For the line \(y=mx+3\) to intersect the parabola \(y=-3x^{2}+2x\), what is the set of all possible values of \(m\)?\\[2pt]
\textbf{Options:}\\[-2pt]
{\setlength{\tabcolsep}{2pt}%
\begin{tabular}[t]{@{}p{0.04\linewidth}p{0.40\linewidth}p{0.04\linewidth}p{0.40\linewidth}@{}}
(1) & \(-4\le m\le 8\) & (2) & \(-8\le m\le 4\) \\
(3) & \(m\le -4 \cup m\ge 8\) & (4) & \(m\le -8 \cup m\ge 4\) \\
\end{tabular}}\\[2pt]
\centering
\includegraphics[width=\linewidth, height=2.3cm, keepaspectratio]{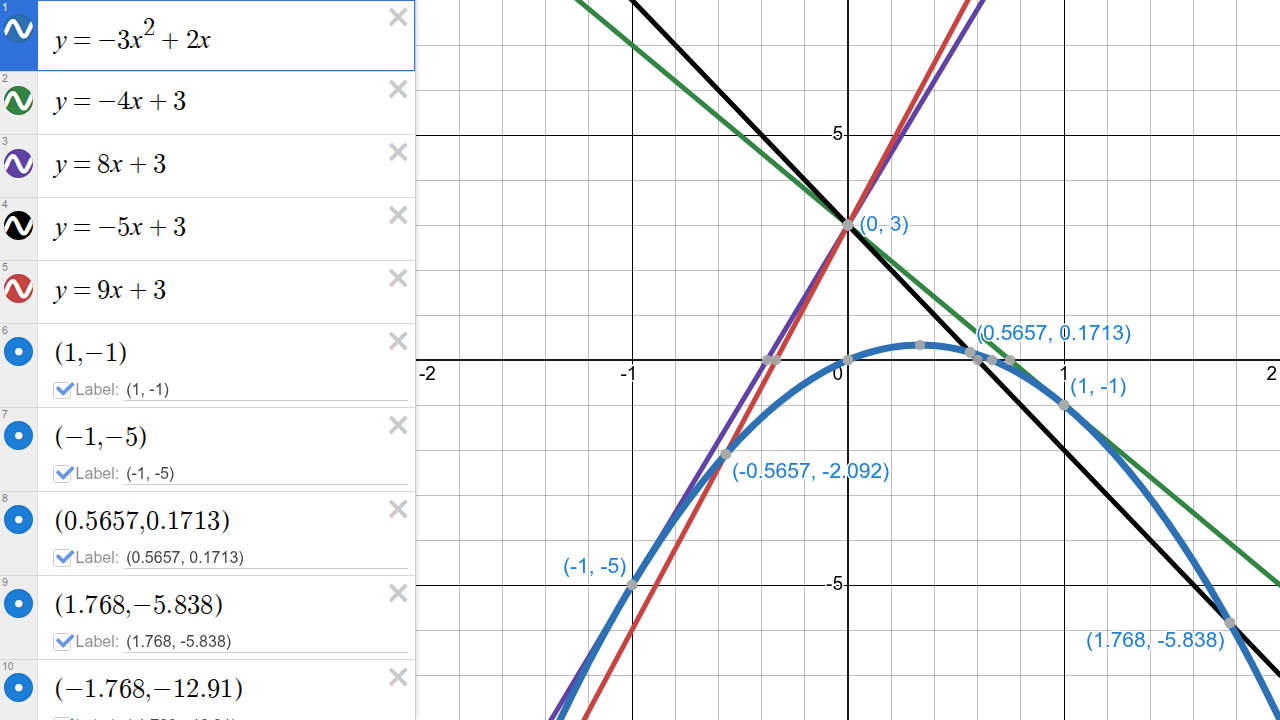}

\begin{flushleft}
\textit{Claude Sonnet 4.6}: ``So m = -4, -5, 8, 9 all give intersections.''\\[1pt]
\textbf{pred:}~1\quad\textbf{gt:}~3\quad{\color{red!70!black}$\times$}
\end{flushleft}
\end{tcolorbox}
\end{minipage}
\hfill
\begin{minipage}[t]{0.485\linewidth}
\begin{tcolorbox}[
    enhanced, colback=white, colframe=red!55!black,
    boxrule=0.4pt, boxsep=2pt, left=4pt, right=4pt, top=2pt, bottom=2pt,
    title={\bfseries\scriptsize Q192 \,\textbar\, EN \quad \itshape\textnormal{FM2 \textbullet{} Endpoint-direction inversion}},
    fonttitle=\scriptsize, coltitle=white,
    boxed title style={colback=red!55!black, sharp corners},
    sharp corners=south
]
\scriptsize
For which values of \(m\) does the line \(y=mx+4\) have no common point with the parabola \(y=-x^2+2x\)?\\[2pt]
\textbf{Options:}\\[-2pt]
{\setlength{\tabcolsep}{2pt}%
\begin{tabular}[t]{@{}p{0.04\linewidth}p{0.40\linewidth}p{0.04\linewidth}p{0.40\linewidth}@{}}
(1) & \(m<0\) & (2) & \(m>4\) \\
(3) & \(-1<m<4\) & (4) & \(-2<m<6\) \\
\end{tabular}}\\[2pt]
\centering
\includegraphics[width=\linewidth, height=2.3cm, keepaspectratio]{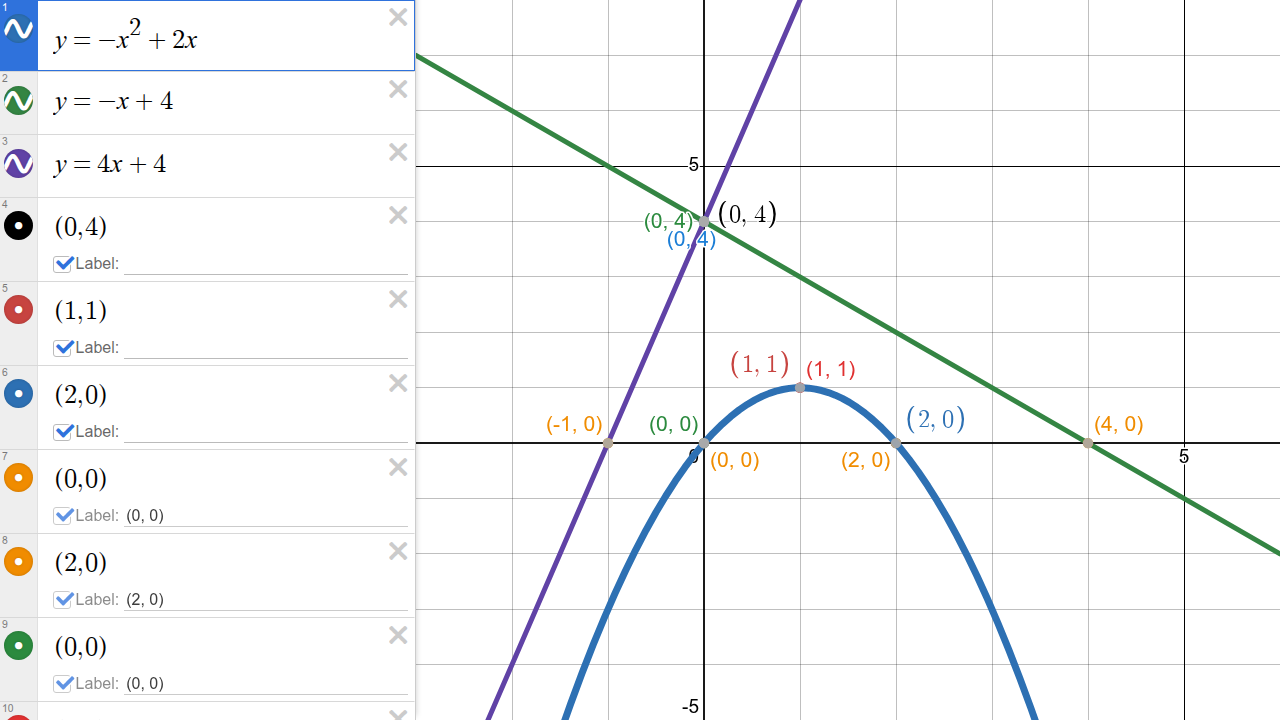}

\begin{flushleft}
\textit{GPT-5.4}: ``The screenshot shows the parabola y=-x\textasciicircum{}2+2x with vertex at the labeled point (1,1), and the two plotted lines y=-x+4 and y=4x+4 both passing through the labeled point (0,4).…''\\[1pt]
\textbf{pred:}~3\quad\textbf{gt:}~4\quad{\color{red!70!black}$\times$}
\end{flushleft}
\end{tcolorbox}
\end{minipage}
\\[6pt]
\begin{minipage}[t]{0.485\linewidth}
\begin{tcolorbox}[
    enhanced, colback=white, colframe=red!55!black,
    boxrule=0.4pt, boxsep=2pt, left=4pt, right=4pt, top=2pt, bottom=2pt,
    title={\bfseries\scriptsize Q2 \,\textbar\, EN \quad \itshape\textnormal{FM3 \textbullet{} Domain-of-inverse confusion}},
    fonttitle=\scriptsize, coltitle=white,
    boxed title style={colback=red!55!black, sharp corners},
    sharp corners=south
]
\scriptsize
The graph of the function \(y=\lvert 2x-6\rvert-\lvert x+4\rvert+x\) is decreasing on an interval. Which of the following is the formula of its inverse on that interval?\\[2pt]
\textbf{Options:}\\[-2pt]
{\setlength{\tabcolsep}{2pt}%
\begin{tabular}[t]{@{}p{0.04\linewidth}p{0.86\linewidth}@{}}
(1) & \(y=-x+6,\; x<-4\) \\
(2) & \(y=-x+5,\; x>2\) \\
(3) & \(y=-\frac{1}{2}x+1,\; -4<x<3\) \\
(4) & \(y=-\frac{1}{2}x+1,\; -4<x<10\) \\
\end{tabular}}\\[2pt]
\centering
\includegraphics[width=\linewidth, height=2.3cm, keepaspectratio]{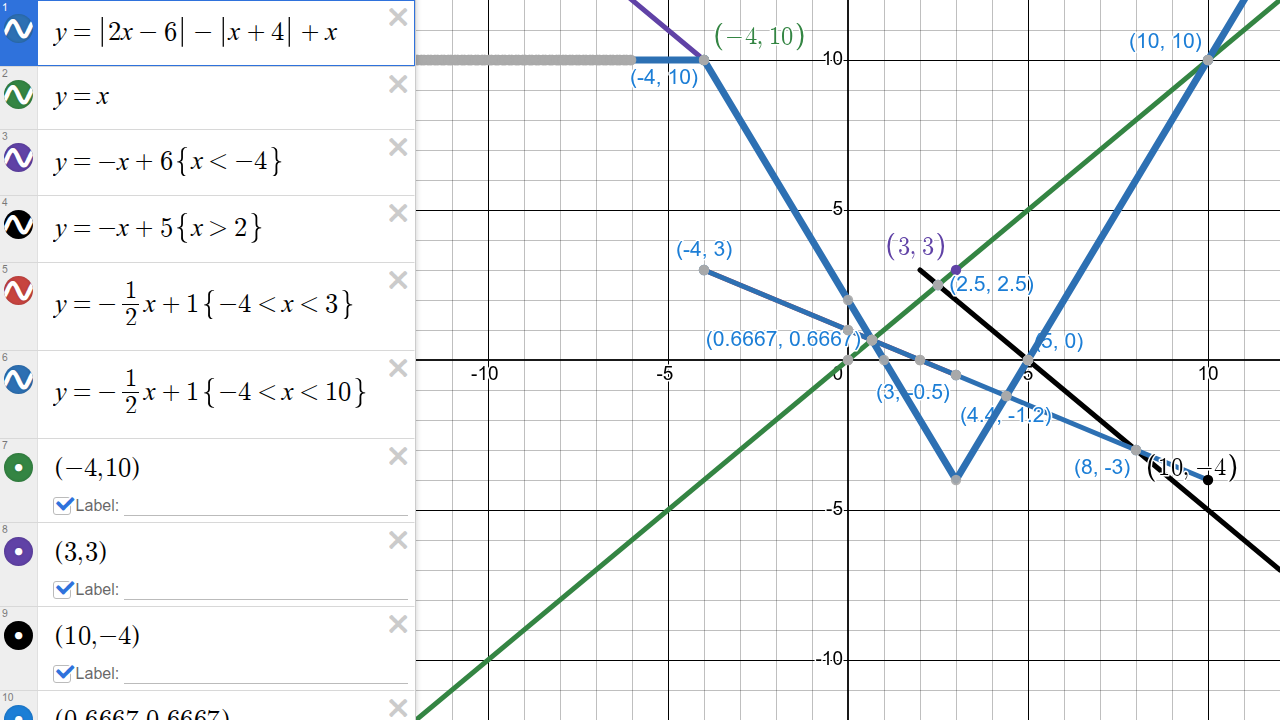}

\begin{flushleft}
\textit{GPT-5.4}: ``The screenshot shows the given graph (blue) with a decreasing segment from the labeled point (-4,10) down to the labeled point (3,-4) passing through the labeled point (5,0).…''\\[1pt]
\textbf{pred:}~2\quad\textbf{gt:}~4\quad{\color{red!70!black}$\times$}
\end{flushleft}
\end{tcolorbox}
\end{minipage}
\hfill
\begin{minipage}[t]{0.485\linewidth}
\begin{tcolorbox}[
    enhanced, colback=white, colframe=red!55!black,
    boxrule=0.4pt, boxsep=2pt, left=4pt, right=4pt, top=2pt, bottom=2pt,
    title={\bfseries\scriptsize Q43 \,\textbar\, FA \quad \itshape\textnormal{FM3 \textbullet{} Domain-of-inverse confusion}},
    fonttitle=\scriptsize, coltitle=white,
    boxed title style={colback=red!55!black, sharp corners},
    sharp corners=south
]
\scriptsize
The function $f(x)=x^{2}\sqrt{x^{2}}$ is decreasing on an interval; which of the following is the formula for the inverse function on that interval?\\[2pt]
\textbf{Options:}\\[-2pt]
{\setlength{\tabcolsep}{2pt}%
\begin{tabular}[t]{@{}p{0.04\linewidth}p{0.40\linewidth}p{0.04\linewidth}p{0.40\linewidth}@{}}
(1) & $-\sqrt{x^{3}},\ x\le 0$ & (2) & $-\sqrt[3]{x},\ x\le 0$ \\
(3) & $-\sqrt{x^{3}},\ x\ge 0$ & (4) & $-\sqrt[3]{x},\ x\ge 0$ \\
\end{tabular}}\\[2pt]
\centering
\includegraphics[width=\linewidth, height=2.3cm, keepaspectratio]{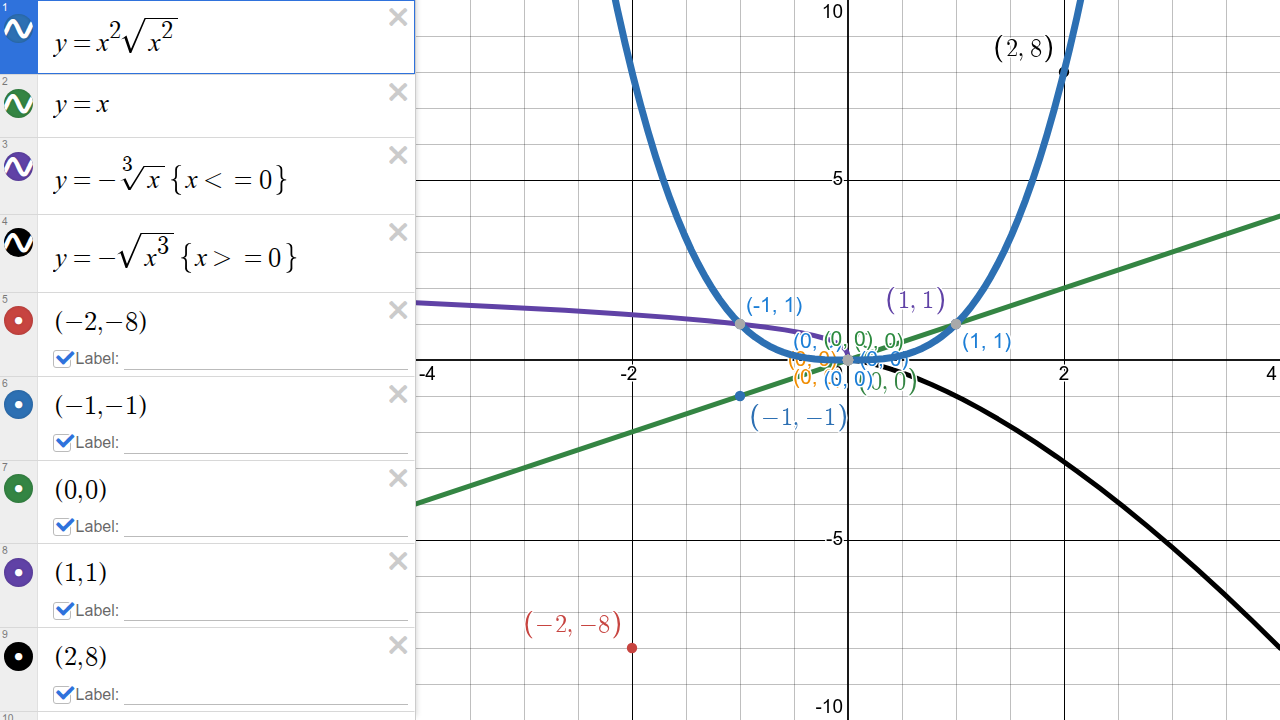}

\begin{flushleft}
\textit{GPT-5.4}: ``From the screenshot, the blue graph y=x\textasciicircum{}2\textbackslash\{\}sqrt\{x\textasciicircum{}2\} passes through the visible labeled points (-1,1), (0,0), and (2,8), and on the left side it comes down toward (0,0) as x i…''\\[1pt]
\textbf{pred:}~2\quad\textbf{gt:}~4\quad{\color{red!70!black}$\times$}
\end{flushleft}
\end{tcolorbox}
\end{minipage}
\\[6pt]
\begin{minipage}[t]{0.485\linewidth}
\begin{tcolorbox}[
    enhanced, colback=white, colframe=red!55!black,
    boxrule=0.4pt, boxsep=2pt, left=4pt, right=4pt, top=2pt, bottom=2pt,
    title={\bfseries\scriptsize Q106 \,\textbar\, EN+FA \quad \itshape\textnormal{FM3 \textbullet{} Domain-of-inverse confusion (wrong branch)}},
    fonttitle=\scriptsize, coltitle=white,
    boxed title style={colback=red!55!black, sharp corners},
    sharp corners=south
]
\scriptsize
The function \(y=x|x-2|\) is decreasing on an interval. Which of the following is the formula of its inverse on that interval?\\[2pt]
\textbf{Options:}\\[-2pt]
{\setlength{\tabcolsep}{2pt}%
\begin{tabular}[t]{@{}p{0.04\linewidth}p{0.86\linewidth}@{}}
(1) & \(1-\sqrt{1+x}\;;\; x<0\) \\
(2) & \(1-\sqrt{1-x}\;;\; x<1\) \\
(3) & \(1+\sqrt{1-x}\;;\; 0<x<1\) \\
(4) & \(1-\sqrt{1-x}\;;\; 0<x<1\) \\
\end{tabular}}\\[2pt]
\centering
\includegraphics[width=\linewidth, height=2.3cm, keepaspectratio]{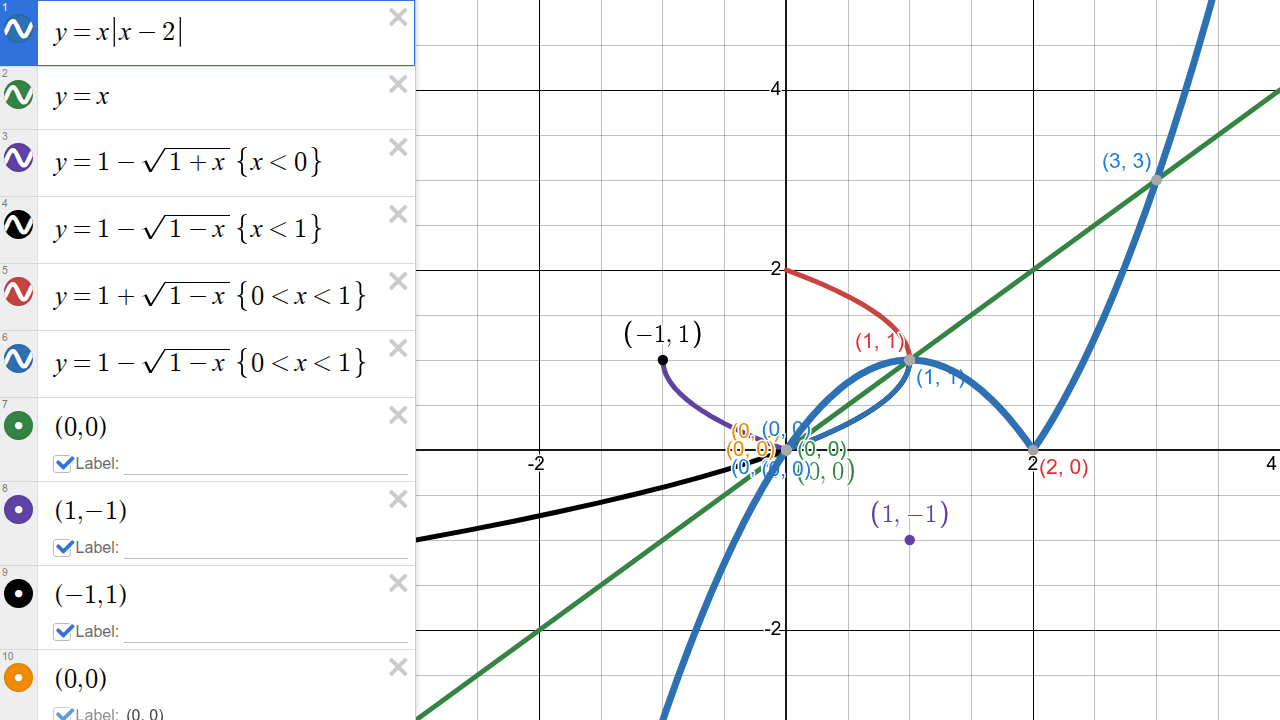}

\begin{flushleft}
\textit{GPT-5.4}: ``The screenshot shows y=x|x-2| decreasing on the blue branch from the visible point (0,0) down to about (1,-1). The inverse of that decreasing branch must be its reflection ac…''\\[1pt]
\textbf{pred:}~1\quad\textbf{gt:}~3\quad{\color{red!70!black}$\times$}
\end{flushleft}
\end{tcolorbox}
\end{minipage}

    \caption{Catalog of hard questions, page 2 of 4: FM3 endpoint-direction inversion and domain-of-inverse confusion. Each panel shows the question stub, a representative wrong model's self-generated Desmos screenshot, a brief verbatim excerpt from that model's final response, and the predicted vs.\ ground-truth option (\(\times\) marks the wrong selection). Panel header colour indicates the dominant failure mode (FM3\,=\,red).}
    \label{fig:catalog-2}
\end{figure}

\begin{figure}[p]
    \centering
\begin{minipage}[t]{0.485\linewidth}
\begin{tcolorbox}[
    enhanced, colback=white, colframe=red!55!black,
    boxrule=0.4pt, boxsep=2pt, left=4pt, right=4pt, top=2pt, bottom=2pt,
    title={\bfseries\scriptsize Q11 \,\textbar\, EN+FA \quad \itshape\textnormal{FM3 \textbullet{} Floor-function discontinuity-count}},
    fonttitle=\scriptsize, coltitle=white,
    boxed title style={colback=red!55!black, sharp corners},
    sharp corners=south
]
\scriptsize
How many points of discontinuity does the function \(f(x)=\lfloor x^{2}\rfloor\) have on the interval \([-1,2]\)? \\[2pt]
\textbf{Options:}\\[-2pt]
{\setlength{\tabcolsep}{2pt}%
\begin{tabular}[t]{@{}p{0.04\linewidth}p{0.40\linewidth}p{0.04\linewidth}p{0.40\linewidth}@{}}
(1) & 3 & (2) & 4 \\
(3) & 5 & (4) & 6 \\
\end{tabular}}\\[2pt]
\centering
\includegraphics[width=\linewidth, height=2.3cm, keepaspectratio]{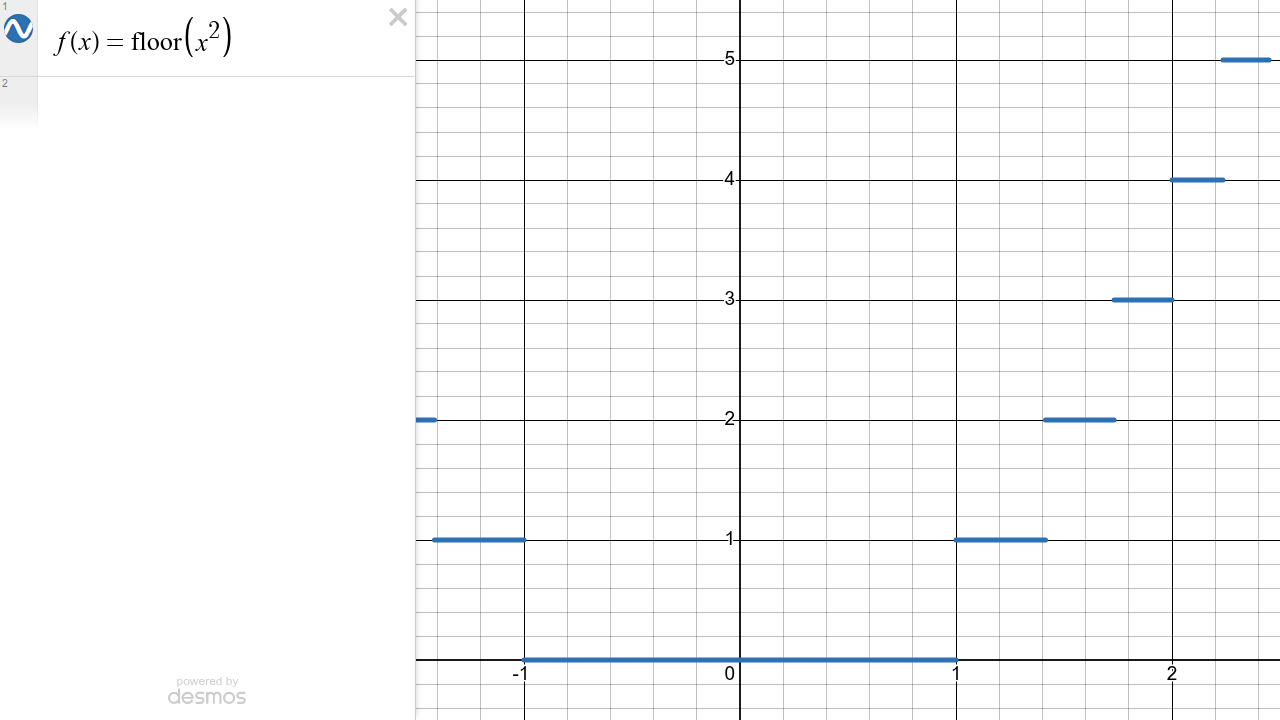}

\begin{flushleft}
\textit{Gemini 2.5 Flash}: ``The plot shows the function f(x) = floor(x\textasciicircum{}2) on the interval [-1, 2]. Discontinuities are visible as jumps in the graph. Counting the jumps within the interval: at x = -1, x…''\\[1pt]
\textbf{pred:}~4\quad\textbf{gt:}~3\quad{\color{red!70!black}$\times$}
\end{flushleft}
\end{tcolorbox}
\end{minipage}
\hfill
\begin{minipage}[t]{0.485\linewidth}
\begin{tcolorbox}[
    enhanced, colback=white, colframe=red!55!black,
    boxrule=0.4pt, boxsep=2pt, left=4pt, right=4pt, top=2pt, bottom=2pt,
    title={\bfseries\scriptsize Q63 \,\textbar\, EN \quad \itshape\textnormal{FM3 \textbullet{} Floor-function value-at-point misread}},
    fonttitle=\scriptsize, coltitle=white,
    boxed title style={colback=red!55!black, sharp corners},
    sharp corners=south
]
\scriptsize
The limit of the expression $\sin\left(\frac{x}{2}\right)\lfloor\cos\left(\frac{x}{2}\right)\rfloor-\cos x\lfloor\sin(2x)\rfloor$ as $x \to \pi$ is which of the following?\\[2pt]
\textbf{Options:}\\[-2pt]
{\fontsize{6}{7.2}\selectfont\setlength{\tabcolsep}{2pt}%
\begin{tabular}[t]{@{}p{0.04\linewidth}p{0.40\linewidth}p{0.04\linewidth}p{0.40\linewidth}@{}}
(1) & -1 & (2) & 0 \\
(3) & 1 & (4) & The limit does not exist. \\
\end{tabular}}\\[2pt]
\centering
\includegraphics[width=\linewidth, height=2.3cm, keepaspectratio]{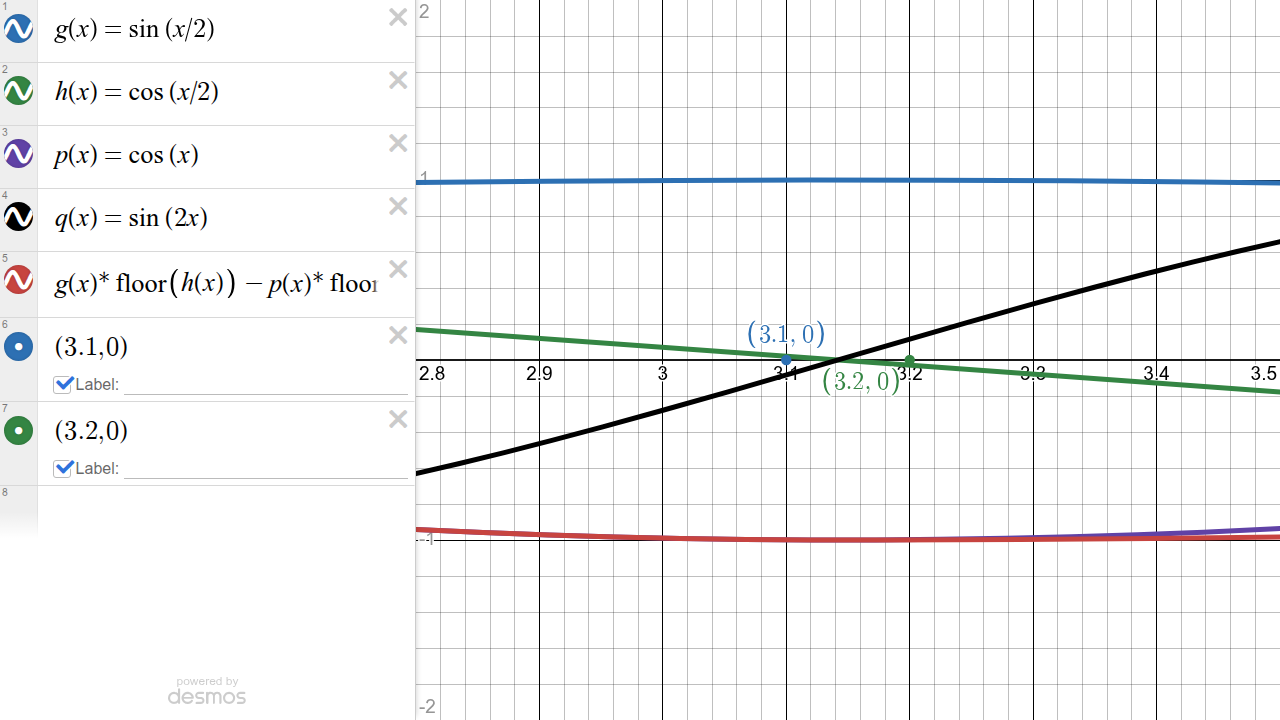}

\begin{flushleft}
\textit{Claude Sonnet 4.6}: ``I plotted g(x)=sin(x/2), h(x)=cos(x/2), p(x)=cos(x), q(x)=sin(2x), and the combined expression g(x)*floor(h(x)) - p(x)*floor(q(x)) near x=pi ($\approx$3.14159). The black cur…''\\[1pt]
\textbf{pred:}~2\quad\textbf{gt:}~1\quad{\color{red!70!black}$\times$}
\end{flushleft}
\end{tcolorbox}
\end{minipage}
\\[6pt]
\begin{minipage}[t]{0.485\linewidth}
\begin{tcolorbox}[
    enhanced, colback=white, colframe=red!55!black,
    boxrule=0.4pt, boxsep=2pt, left=4pt, right=4pt, top=2pt, bottom=2pt,
    title={\bfseries\scriptsize Q64 \,\textbar\, FA \quad \itshape\textnormal{FM3 \textbullet{} Asymptote count at floor-jump boundary}},
    fonttitle=\scriptsize, coltitle=white,
    boxed title style={colback=red!55!black, sharp corners},
    sharp corners=south
]
\scriptsize
How many asymptotes does the function $y=\frac{x}{x-\lfloor x\rfloor\ -1}$ have on the interval $(-1,3]$?\\[2pt]
\textbf{Options:}\\[-2pt]
{\setlength{\tabcolsep}{2pt}%
\begin{tabular}[t]{@{}p{0.04\linewidth}p{0.40\linewidth}p{0.04\linewidth}p{0.40\linewidth}@{}}
(1) & 2 & (2) & 3 \\
(3) & 4 & (4) & 5 \\
\end{tabular}}\\[2pt]
\centering
\includegraphics[width=\linewidth, height=2.3cm, keepaspectratio]{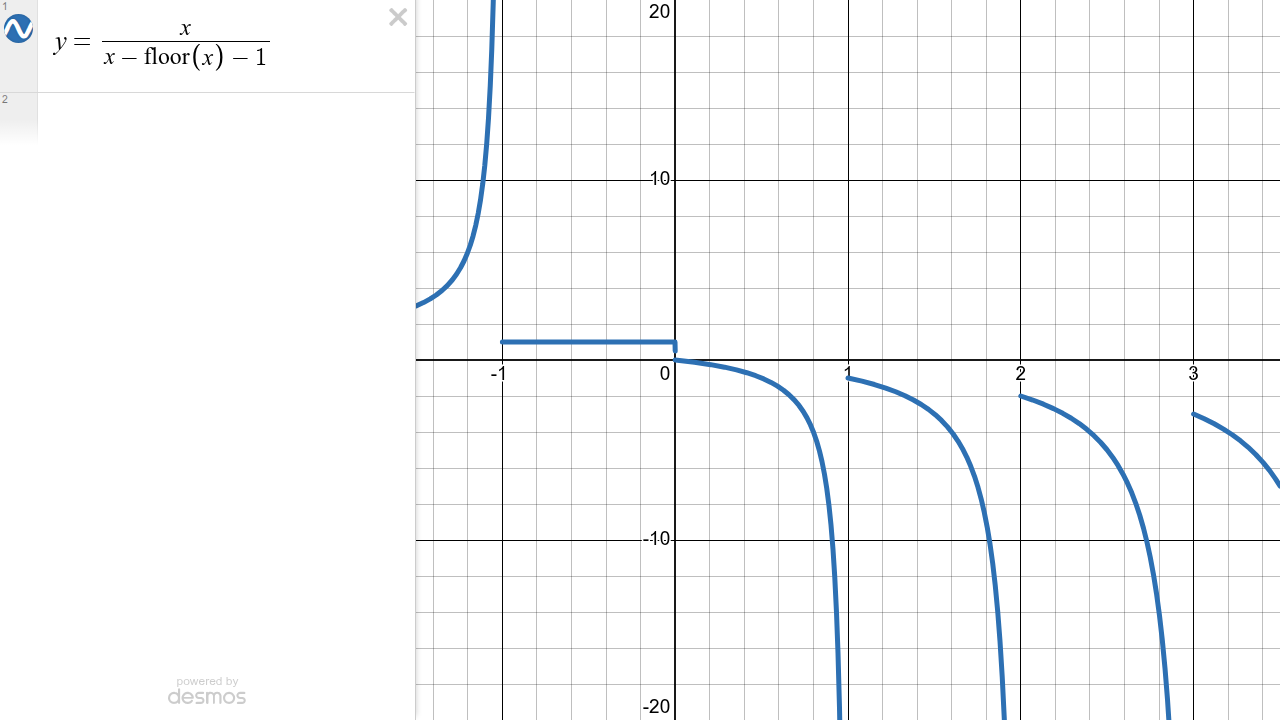}

\begin{flushleft}
\textit{Claude Sonnet 4.6}: ``From the Desmos plot of y = x/(x - floor(x) - 1) on (-1, 3], I can visually observe: (1) A vertical asymptote near x = -1 (curve shoots up to +$\infty$ as x approaches -1 fro…''\\[1pt]
\textbf{pred:}~4\quad\textbf{gt:}~2\quad{\color{red!70!black}$\times$}
\end{flushleft}
\end{tcolorbox}
\end{minipage}
\hfill
\begin{minipage}[t]{0.485\linewidth}
\begin{tcolorbox}[
    enhanced, colback=white, colframe=red!55!black,
    boxrule=0.4pt, boxsep=2pt, left=4pt, right=4pt, top=2pt, bottom=2pt,
    title={\bfseries\scriptsize Q4 \,\textbar\, EN+FA \quad \itshape\textnormal{FM2 \textbullet{} Polygon-classification under-precision}},
    fonttitle=\scriptsize, coltitle=white,
    boxed title style={colback=red!55!black, sharp corners},
    sharp corners=south
]
\scriptsize
The solutions of the trigonometric equation \(\cot x+\tan x-4\cos 2x=0\) lie on the unit circle. The points corresponding to these solutions are the vertices of whi…\\[2pt]
\textbf{Options:}\\[-2pt]
{\fontsize{6}{7.2}\selectfont\setlength{\tabcolsep}{2pt}%
\begin{tabular}[t]{@{}p{0.04\linewidth}p{0.40\linewidth}p{0.04\linewidth}p{0.40\linewidth}@{}}
(1) & Square & (2) & Rectangle \\
(3) & Triangle & (4) & Quadrilateral with non-right angles \\
\end{tabular}}\\[2pt]
\centering
\includegraphics[width=\linewidth, height=2.3cm, keepaspectratio]{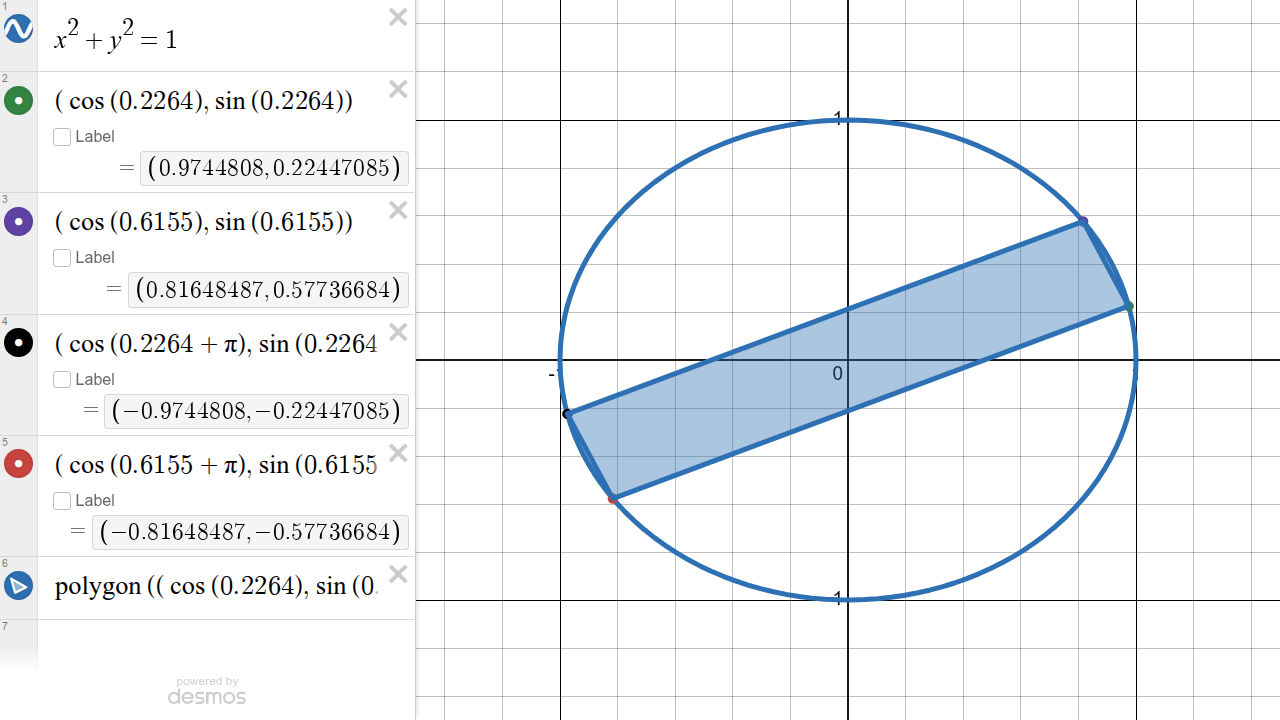}

\begin{flushleft}
\textit{Claude Sonnet 4.6}: ``From the first screenshot, the zeros of cot(x)+tan(x)-4cos(2x)=0 were found near x$\approx$0.2264 and x$\approx$0.6155 (and their $\pi$-shifted counterparts). The four soluti…''\\[1pt]
\textbf{pred:}~2\quad\textbf{gt:}~1\quad{\color{red!70!black}$\times$}
\end{flushleft}
\end{tcolorbox}
\end{minipage}
    \caption{Catalog of hard questions, page 3 of 4: FM3 floor-function / discontinuity counting and polygon classification. Each panel shows the question stub, a representative wrong model's self-generated Desmos screenshot, a brief verbatim excerpt from that model's final response, and the predicted vs.\ ground-truth option (\(\times\) marks the wrong selection). Panel header colour indicates the dominant failure mode (FM3\,=\,red).}
    \label{fig:catalog-3}
\end{figure}

\begin{figure}[p]
    \centering
\begin{minipage}[t]{0.485\linewidth}
\begin{tcolorbox}[
    enhanced, colback=white, colframe=red!55!black,
    boxrule=0.4pt, boxsep=2pt, left=4pt, right=4pt, top=2pt, bottom=2pt,
    title={\bfseries\scriptsize Q13 \,\textbar\, FA \quad \itshape\textnormal{FM3 \textbullet{} Cusp / inflection-point misidentification}},
    fonttitle=\scriptsize, coltitle=white,
    boxed title style={colback=red!55!black, sharp corners},
    sharp corners=south
]
\scriptsize
What is the x-coordinate of the inflection point of the function \(y=(5-x)\sqrt[3]{x^{2}}\)?\\[2pt]
\textbf{Options:}\\[-2pt]
{\setlength{\tabcolsep}{2pt}%
\begin{tabular}[t]{@{}p{0.04\linewidth}p{0.40\linewidth}p{0.04\linewidth}p{0.40\linewidth}@{}}
(1) & -1 & (2) & 0 \\
(3) & 1 & (4) & 2 \\
\end{tabular}}\\[2pt]
\centering
\includegraphics[width=\linewidth, height=2.3cm, keepaspectratio]{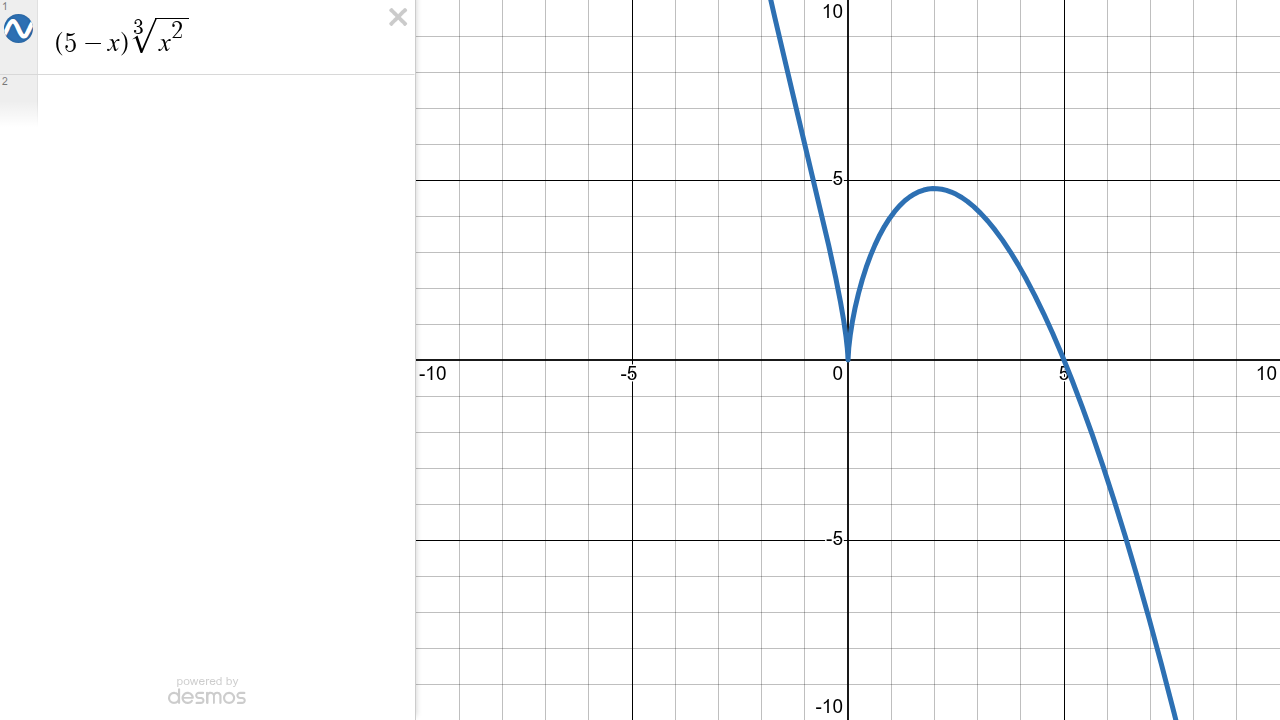}

\begin{flushleft}
\textit{GPT-5.4}: ``In the Desmos view around the origin, the curve y=(5-x)\textbackslash\{\}sqrt[3]\{x\textasciicircum{}2\} visibly bends downward on the left of x=0 and upward on the right of x=0. The change in concavity occurs…''\\[1pt]
\textbf{pred:}~N/A\quad\textbf{gt:}~1\quad{\color{red!70!black}$\times$}
\end{flushleft}
\end{tcolorbox}
\end{minipage}
\hfill
\begin{minipage}[t]{0.485\linewidth}
\begin{tcolorbox}[
    enhanced, colback=white, colframe=red!55!black,
    boxrule=0.4pt, boxsep=2pt, left=4pt, right=4pt, top=2pt, bottom=2pt,
    title={\bfseries\scriptsize Q73 \,\textbar\, EN \quad \itshape\textnormal{FM3 \textbullet{} Concavity / sign-of-derivative misread}},
    fonttitle=\scriptsize, coltitle=white,
    boxed title style={colback=red!55!black, sharp corners},
    sharp corners=south
]
\scriptsize
For the function \(f(x)=\sin^{2}x-2\sin x\) on the interval \(x\in[0,2\pi]\), in which interval is \(f\) increasing and its concavity downward?\\[2pt]
\textbf{Options:}\\[-2pt]
{\setlength{\tabcolsep}{2pt}%
\begin{tabular}[t]{@{}p{0.04\linewidth}p{0.40\linewidth}p{0.04\linewidth}p{0.40\linewidth}@{}}
(1) & \(\left(\frac{7\pi}{6},\frac{3\pi}{2}\right)\) & (2) & \(\left(\frac{3\pi}{2},\frac{11\pi}{6}\right)\) \\
(3) & \(\left(\frac{\pi}{2},\frac{5\pi}{6}\right)\) & (4) & \(\left(\frac{\pi}{2},\frac{7\pi}{6}\right)\) \\
\end{tabular}}\\[2pt]
\centering
\includegraphics[width=\linewidth, height=2.3cm, keepaspectratio]{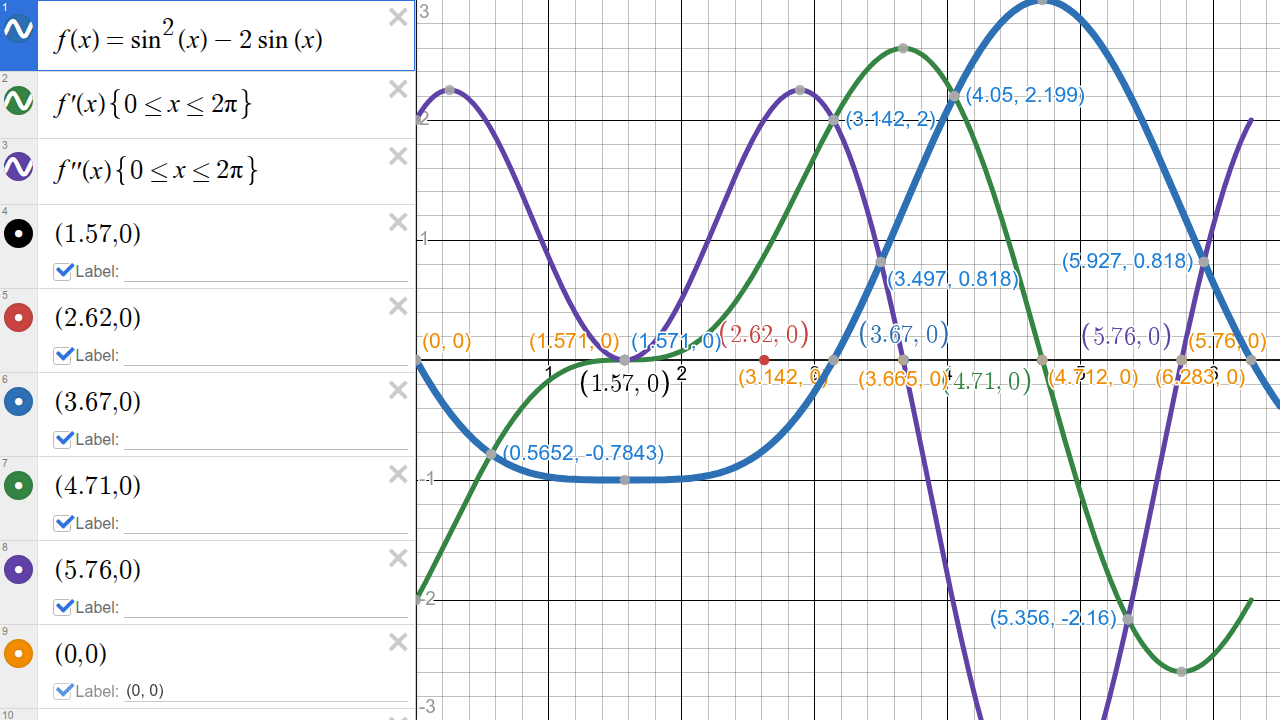}

\begin{flushleft}
\textit{GPT-5.4}: ``From the Desmos screenshot, the green graph f'(x) crosses the x-axis at labeled points near x=1.571, 4.71, and 6.283. Between x=4.71 and x=6.283, the green graph stays above…''\\[1pt]
\textbf{pred:}~2\quad\textbf{gt:}~1\quad{\color{red!70!black}$\times$}
\end{flushleft}
\end{tcolorbox}
\end{minipage}
\\[6pt]
\begin{minipage}[t]{0.485\linewidth}
\begin{tcolorbox}[
    enhanced, colback=white, colframe=red!55!black,
    boxrule=0.4pt, boxsep=2pt, left=4pt, right=4pt, top=2pt, bottom=2pt,
    title={\bfseries\scriptsize Q98 \,\textbar\, FA \quad \itshape\textnormal{FM3 \textbullet{} Strictly-decreasing interval miscount}},
    fonttitle=\scriptsize, coltitle=white,
    boxed title style={colback=red!55!black, sharp corners},
    sharp corners=south
]
\scriptsize
How many intervals are there on which the function \(f(x)=\frac{x^{4}-3}{x^{2}-2}\), with \(x\in(-2,2)\), is strictly decreasing?\\[2pt]
\textbf{Options:}\\[-2pt]
{\setlength{\tabcolsep}{2pt}%
\begin{tabular}[t]{@{}p{0.04\linewidth}p{0.40\linewidth}p{0.04\linewidth}p{0.40\linewidth}@{}}
(1) & 2 & (2) & 3 \\
(3) & 4 & (4) & 5 \\
\end{tabular}}\\[2pt]
\centering
\includegraphics[width=\linewidth, height=2.3cm, keepaspectratio]{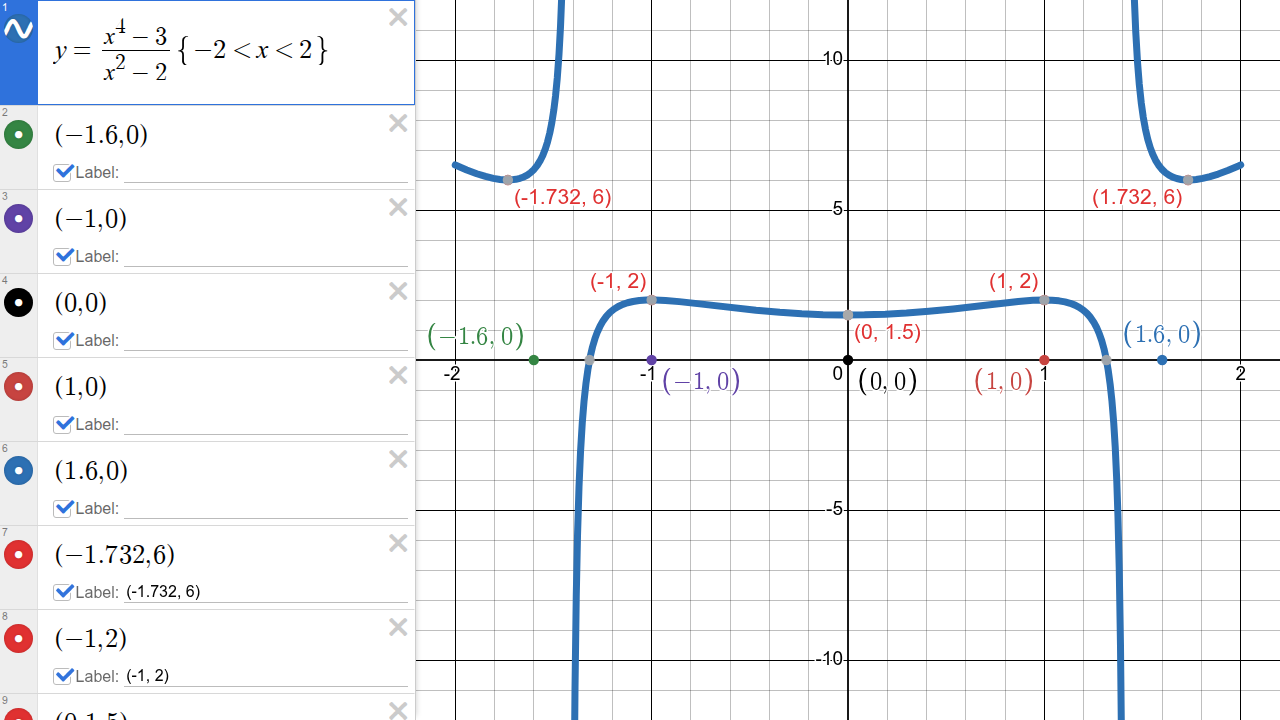}

\begin{flushleft}
\textit{GPT-5.4}: ``From the Desmos screenshot of y=(x\textasciicircum{}4-3)/(x\textasciicircum{}2-2) on -2\textless{}x\textless{}2, two vertical breaks are clearly visible near x$\approx$-1.4 and x$\approx$1.4, splitting the graph into three branc…''\\[1pt]
\textbf{pred:}~2\quad\textbf{gt:}~3\quad{\color{red!70!black}$\times$}
\end{flushleft}
\end{tcolorbox}
\end{minipage}
\hfill
\begin{minipage}[t]{0.485\linewidth}
\begin{tcolorbox}[
    enhanced, colback=white, colframe=red!55!black,
    boxrule=0.4pt, boxsep=2pt, left=4pt, right=4pt, top=2pt, bottom=2pt,
    title={\bfseries\scriptsize Q198 \,\textbar\, FA \quad \itshape\textnormal{FM3 \textbullet{} Piecewise jump-discontinuity blindness}},
    fonttitle=\scriptsize, coltitle=white,
    boxed title style={colback=red!55!black, sharp corners},
    sharp corners=south
]
\scriptsize
Determine the monotonicity of the function \(y=\begin{cases}x^2-4x, & x<0\\ -x+2, & x\ge 0\end{cases}\).\\[2pt]
\textbf{Options:}\\[-2pt]
{\fontsize{6}{7.2}\selectfont\setlength{\tabcolsep}{2pt}%
\begin{tabular}[t]{@{}p{0.04\linewidth}p{0.40\linewidth}p{0.04\linewidth}p{0.40\linewidth}@{}}
(1) & Increasing & (2) & Decreasing \\
(3) & Neither increasing nor decreasing & (4) & Both increasing and decreasing \\
\end{tabular}}\\[2pt]
\centering
\includegraphics[width=\linewidth, height=2.3cm, keepaspectratio]{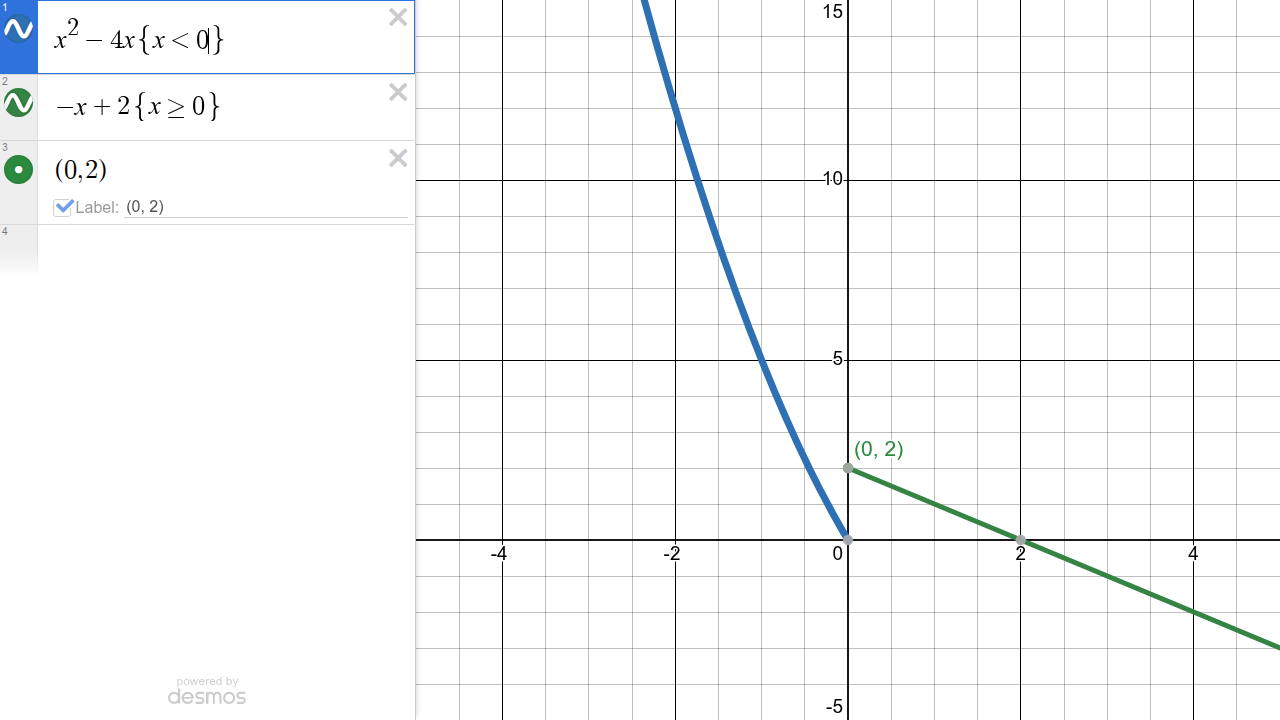}

\begin{flushleft}
\textit{Claude Sonnet 4.6}: ``This is strictly decreasing as x increases toward 0.''\\[1pt]
\textbf{pred:}~2\quad\textbf{gt:}~3\quad{\color{red!70!black}$\times$}
\end{flushleft}
\end{tcolorbox}
\end{minipage}
\\[6pt]
\begin{minipage}[t]{0.485\linewidth}
\begin{tcolorbox}[
    enhanced, colback=white, colframe=blue!55!black,
    boxrule=0.4pt, boxsep=2pt, left=4pt, right=4pt, top=2pt, bottom=2pt,
    title={\bfseries\scriptsize Q3 \,\textbar\, EN+FA \quad \itshape\textnormal{FM4 \textbullet{} Analytic-prior hallucination}},
    fonttitle=\scriptsize, coltitle=white,
    boxed title style={colback=blue!55!black, sharp corners},
    sharp corners=south
]
\scriptsize
The function \(f(x)=x^{2}+2x+1\) is defined with domain \((-1,+\infty)\). How many intersection points do the graphs of \(f\) and \(f^{-1}\) have?\\[2pt]
\textbf{Options:}\\[-2pt]
{\setlength{\tabcolsep}{2pt}%
\begin{tabular}[t]{@{}p{0.04\linewidth}p{0.40\linewidth}p{0.04\linewidth}p{0.40\linewidth}@{}}
(1) & 1 & (2) & 2 \\
(3) & 3 & (4) & No intersection \\
\end{tabular}}\\[2pt]
\centering
\includegraphics[width=\linewidth, height=2.3cm, keepaspectratio]{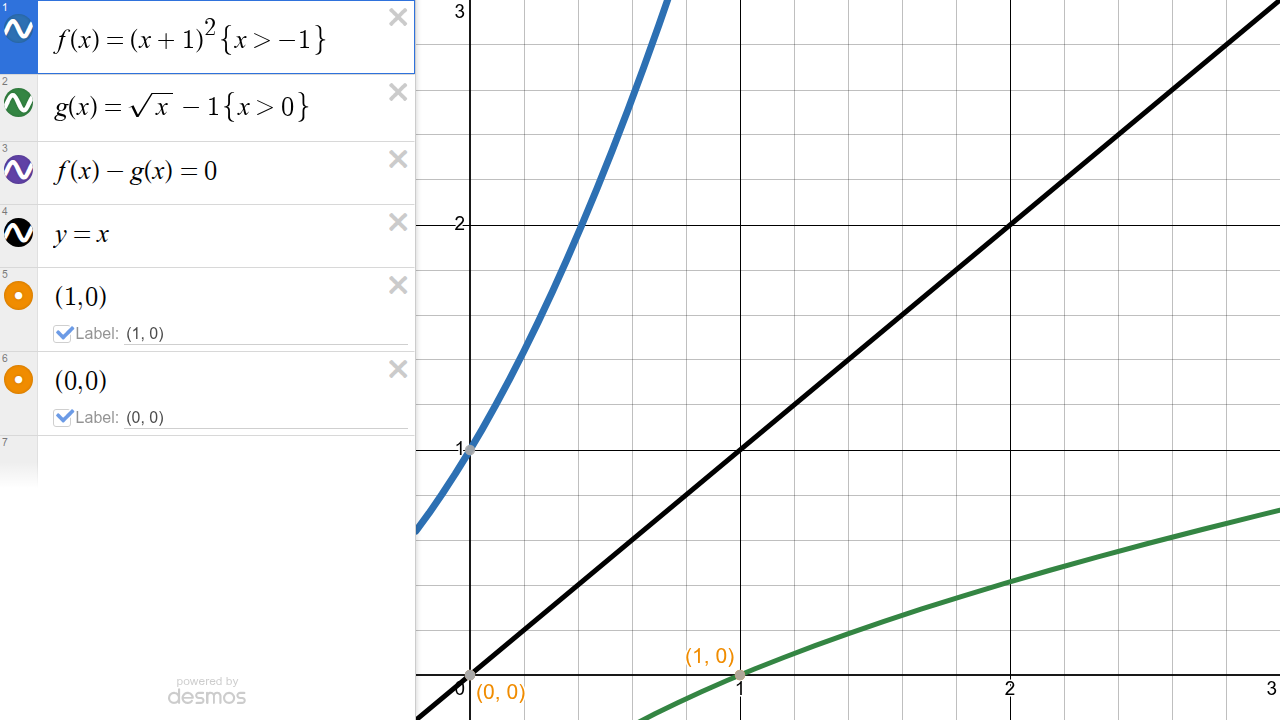}

\begin{flushleft}
\textit{Claude Sonnet 4.6}: ``I plotted f(x)=(x+1)\textasciicircum{}2 for x\textgreater{}-1 (blue), its inverse g(x)=sqrt x-1 for x\textgreater{}0 (green), the line y=x (black), and the equation f(x)-g(x)=0 to find where f and f\textasciicircum{}\{-1\} intersect. Th…''\\[1pt]
\textbf{pred:}~2\quad\textbf{gt:}~4\quad{\color{red!70!black}$\times$}
\end{flushleft}
\end{tcolorbox}
\end{minipage}
\hfill
\begin{minipage}[t]{0.485\linewidth}
\begin{tcolorbox}[
    enhanced, colback=white, colframe=blue!55!black,
    boxrule=0.4pt, boxsep=2pt, left=4pt, right=4pt, top=2pt, bottom=2pt,
    title={\bfseries\scriptsize Q78 \,\textbar\, EN \quad \itshape\textnormal{FM4 \textbullet{} Analytic-prior hallucination (sum-of-asymptotes)}},
    fonttitle=\scriptsize, coltitle=white,
    boxed title style={colback=blue!55!black, sharp corners},
    sharp corners=south
]
\scriptsize
Two functions \(f(x)=\frac{x+1}{x+\sqrt{x}}\) and \(g(x)=\frac{1-x}{x-\sqrt{x}}\) are defined. How many asymptotes does the graph of \((f+g)(x)\) have?\\[2pt]
\textbf{Options:}\\[-2pt]
{\setlength{\tabcolsep}{2pt}%
\begin{tabular}[t]{@{}p{0.04\linewidth}p{0.40\linewidth}p{0.04\linewidth}p{0.40\linewidth}@{}}
(1) & 0 & (2) & 1 \\
(3) & 2 & (4) & 3 \\
\end{tabular}}\\[2pt]
\centering
\includegraphics[width=\linewidth, height=2.3cm, keepaspectratio]{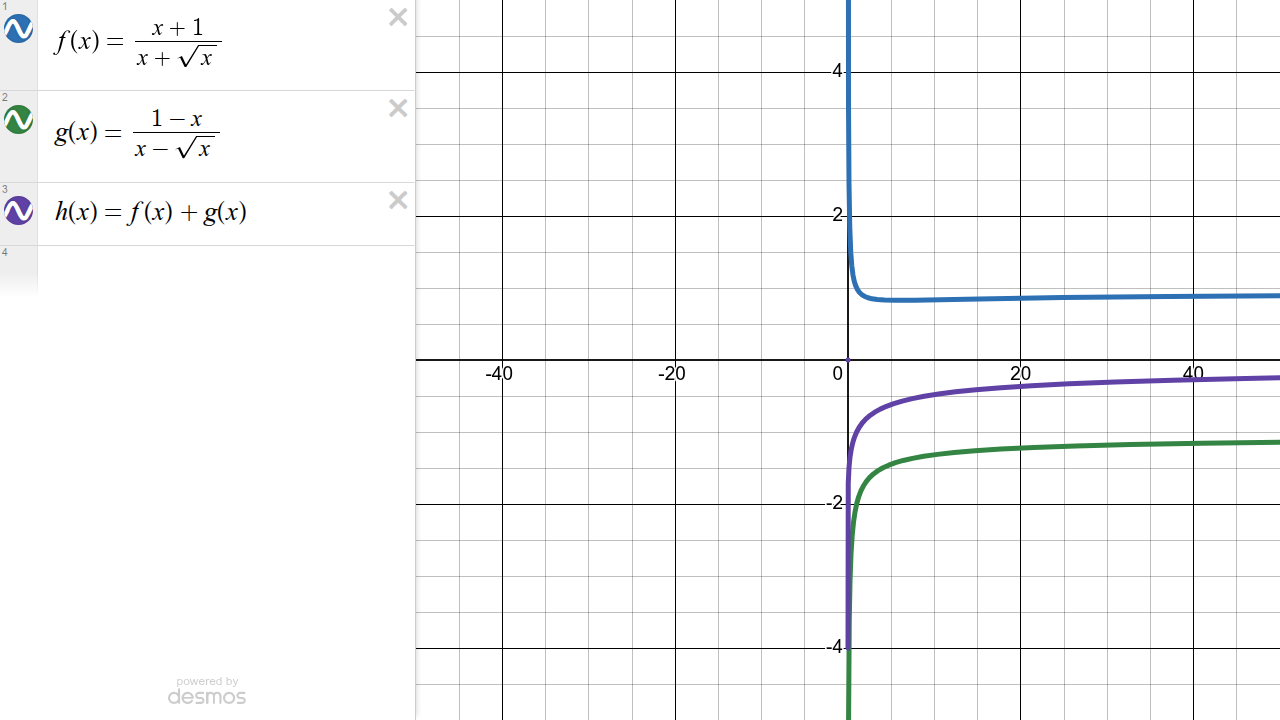}

\begin{flushleft}
\textit{Gemini 2.5 Flash}: ``The graph of h(x) = f(x) + g(x) (purple curve) shows a clear vertical asymptote at x=0, as the curve approaches positive and negative infinity as x approaches 0 from the righ…''\\[1pt]
\textbf{pred:}~3\quad\textbf{gt:}~2\quad{\color{red!70!black}$\times$}
\end{flushleft}
\end{tcolorbox}
\end{minipage}
    \caption{Catalog of hard questions, page 4 of 4: FM3 cusp / inflection, concavity, decreasing-interval, piecewise jump-discontinuity, and FM4 analytic-prior hallucination. Each panel shows the question stub, a representative wrong model's self-generated Desmos screenshot, a brief verbatim excerpt from that model's final response, and the predicted vs.\ ground-truth option (\(\times\) marks the wrong selection). Panel header colour indicates the dominant failure mode (FM3\,=\,red, FM4\,=\,blue).}
    \label{fig:catalog-4}
\end{figure}

This appendix catalogues every question on which at least 10 evaluated models
give the wrong tool-enabled answer, excluding the three questions that already
receive a full deep-dive treatment elsewhere: Question~123 (Figure~\ref{fig:appendix-case-a}),
Question~26 (Figure~\ref{fig:appendix-case-b}), and Question~199 (Figure~\ref{fig:pipeline}).
Table~\ref{tab:error-taxonomy} summarizes the failure-mode taxonomy and the
prevalence of each named sub-pattern across the catalogued questions.
Each card in Figures~\ref{fig:catalog-1}--\ref{fig:catalog-4} carries a
question identifier, the dominant failure-mode tag (FM1--FM4 from
\S\ref{sec:analysis}) with its sub-pattern label, the question stub, one
representative wrong model's self-generated Desmos screenshot, a short
verbatim excerpt from that model's final response, and its predicted option
versus the ground truth. For overlap questions hard in both languages, the
English run is shown.

The 21 questions span all four failure modes from Section~\ref{sec:analysis},
with FM3 (correct graph, incorrect interpretation) the dominant mode by a
wide margin.

\begin{table}[!t]
    \centering
    \footnotesize
    \caption{Qualitative error taxonomy for VAMPS, with prevalence on the questions where at least 10 evaluated models give the wrong tool-enabled answer. ``\#Q'' counts questions whose dominant observed failure matches the row.}
    \label{tab:error-taxonomy}
    \begin{tabularx}{\linewidth}{lXc}
        \toprule
        Failure type & Observable symptom & \#Q \\
        \midrule
        Tool-call malformation (\S\ref{sec:analysis}, FM1) & Premature handoff or invalid \texttt{desmos\_plot} JSON; final answer with no usable screenshot & model-specific \\
        Auto-label over-trust (FM2) & Trusting the count of Desmos's POI labels (zeros, intersections, extrema) instead of distinct visual features & 3 \\
        Endpoint-direction inversion (FM3) & Choosing the complement of the correct interval or region & 5 \\
        Domain-of-inverse confusion (FM3) & Swapping variables in the inverse formula but keeping the original domain instead of the range & 4 \\
        Sign / quadrant misread (FM3) & Asserting overlap between functions of opposite sign in the visible window & 2 \\
        Analytic-prior hallucination (FM4) & Invoking a canonical fact (e.g.\ ``$f$ and $f^{-1}$ always meet on $y=x$'', naive sum of asymptotes) that the plot would refute & 6 \\
        \bottomrule
    \end{tabularx}
\end{table}

\end{document}